\documentclass{article}

\PassOptionsToPackage{numbers, compress}{natbib}

\usepackage[final]{neurips_2024}
\usepackage[page,toc,titletoc,title]{appendix}

\usepackage{amsmath,amsfonts,bm}
\usepackage{hyperref}       %

\def\eqref#1{equation~\ref{#1}}

\def\1{\bm{1}}

\def\mI{{\bm{I}}}

\DeclareMathAlphabet{\mathsfit}{\encodingdefault}{\sfdefault}{m}{sl}
\SetMathAlphabet{\mathsfit}{bold}{\encodingdefault}{\sfdefault}{bx}{n}

\def\snr{\text{$\bm \nu$}}
\def\x{{\mathbf x}}
\def\z{{\mathbf z}}
\def\c{{\mathbf c}}
\def\cp{{\mathbf c_\theta}}
\def\a{{\mathbf a}_\phi}
\def\b{{\mathbf b}_\phi}
\def\d{{\mathbf d}_\phi}
\def\dif{{\text{d}}}

\def\f{{\mathbf f}}
\def\g{{\mathbf g}}
\def\r{{\mathbf r}}
\def\w{{\mathbf w}}

\def\q{{q_\phi}}
\def\p{{p_\theta}}
\def\vsnr{{{\snr}}}
\def\swrs{{S_{\text{wrs}}}}

\def\gradx{\nabla_{\x_t}}
\def\fn{f(x; \psi)}
\def\fngrad{f'(x; \psi)}
\def\fngrada{f'(x_1; \psi)}
\def\fngradb{f'(x_2; \psi)}
\def\method{{Mulan}}
\def\s{{s(i)}}
\def\t{{t(i)}}
\def\bmu{{\bm \mu}}
\def\balpha{{\bm \alpha}}
\def\alphai{{\balpha_i}}
\def\sigmai{{\bsigma_i}}
\def\nui{{\snr_i}}
\def\bSigma{{\bm \Sigma}}
\def\bsigma{{\bm \sigma}}
\def\unet{{\bm \epsilon_\theta}}
\def\score{{\mathbf s}_\theta}
\def\noise{{\bm \epsilon}}
\def\bmu{{\bm \mu}}
\def\ns{{\bm \gamma}}
\def\tneps{{\bm \hat{\epsilon}}}
\def\nsmin{{\gamma_\text{min}}}
\def\nsmax{{\gamma_\text{max}}}
\def\forcefield{{ \mathbf{f}_\theta(\x_0, \snr(\z, t))}}
\def\forcefieldnoz{{ \mathbf{f}_\theta(\x_0, \snr(t))}}

\def\lossz{{\mathcal{L}_{\text{latent}}}}
\def\lossprior{{\mathcal{L}_{\text{prior}}}}
\def\lossdiff{{\mathcal{L}_{\text{diffusion}}}}

\def\lossrecons{{\mathcal{L}_{\text{recons}}}}
\def\kl{\text{D}_{\text{KL}}}
\def\diag{{\text{diag}}}
\def\method{\textsc{MuLAN}}
\def\one{\mathbf{1_d}}

\usepackage{amsfonts}       %
\usepackage{amsmath}       %
\usepackage{amssymb}
\usepackage{nicefrac}

\newcommand{\fig}[1]{Fig.~\ref{#1}}    %
    
\newcommand{\tab}[1]{Table~\ref{#1}}
\newcommand{\Eqn}[1]{Eq.~\ref{#1}}
\renewcommand{\sec}[1]{Sec.~\ref{#1}} %
\newcommand{\supp}[1]{Suppl.~\ref{#1}}

\usepackage{xspace}
\makeatletter
\DeclareRobustCommand\onedot{\futurelet\@let@token\@onedot}
\def\@onedot{\ifx\@let@token.\else.\null\fi\xspace}

\makeatother

\newcommand{\bfI}{\mathbf{I}}

\usepackage{xcolor}
\definecolor{ourblue}{rgb}{0.368,0.507,0.71}
\definecolor{ourorange}{rgb}{0.881,0.611,0.142}
\definecolor{ourgreen}{rgb}{0.56,0.692,0.195}
\definecolor{ourred}{rgb}{0.923,0.386,0.209}
\definecolor{ourviolet}{rgb}{0.528,0.471,0.701}
\definecolor{ourbrown}{rgb}{0.772,0.432,0.102}
\definecolor{ourlightblue}{rgb}{0.364,0.619,0.782}
\definecolor{ourdarkgreen}{rgb}{0.572,0.586,0.}

\definecolor{ourcyan2}{rgb}{0.125,0.722,0.804}
\definecolor{ourred2}{rgb}{0.863,0.184,0.047}
\definecolor{ouryellow2}{cmyk}{0,0.16,1.0,0.07}
\definecolor{ourviolet2}{cmyk}{0.55,0.56,0,0.47}
\definecolor{ourorange2}{cmyk}{0,0.46,0.89,0.11}
\definecolor{url}{HTML}{d95225}
\usepackage{lipsum} 
\usepackage[utf8]{inputenc} %
\usepackage[T1]{fontenc}    %

\usepackage{xr-hyper}
\usepackage{url}            %
\usepackage{booktabs}       %
\usepackage{multirow}       
\usepackage{amsfonts}       %
\usepackage{nicefrac}       %
\usepackage{microtype}      %
\usepackage{xcolor}         %
\usepackage{graphicx}
\usepackage{amsmath}
\usepackage{amssymb}
\usepackage{float}
\usepackage{bm}
\usepackage{cases}
\usepackage[
	disable,
	textsize=tiny,
	textwidth=2.2cm
	]{todonotes}
\usepackage{blindtext}
\usepackage{textcomp}
\usepackage{mathtools}
\usepackage{xargs}
\usepackage{caption}
\usepackage{subcaption}
\usepackage{wrapfig}
\usepackage{array}

\hypersetup{
	final,
	colorlinks,
	linkcolor=ourred,
	citecolor=ourblue,
	urlcolor=url,
}

\title{Diffusion Models With Learned Adaptive Noise}

\author{
  Subham Sekhar Sahoo \\
  Cornell Tech, NYC, USA.\\
  \texttt{ssahoo@cs.cornell.edu} \\
  \And
  Aaron Gokaslan \\
  Cornell Tech, NYC, USA.\\
  \texttt{akg87@cs.cornell.edu} \\
  \And
  Chris De Sa \\
  Cornell University, Ithaca, USA.\\
  \texttt{cdesa@cs.cornell.edu} \\
  \And
  Volodymyr Kuleshov \\
  Cornell Tech, NYC, USA.\\
  \texttt{kuleshov@cornell.edu} \\
}

\begin{document}

\maketitle

\begin{abstract}
Diffusion models have gained traction as powerful algorithms for synthesizing high-quality images. Central to these algorithms is the diffusion process, a set of equations which maps data to noise 
in a way that can significantly affect performance. 
In this paper, we explore whether the diffusion
process can be learned from data.
Our work is grounded in Bayesian inference and seeks to improve log-likelihood estimation by casting the learned diffusion process as an approximate variational posterior that yields a tighter lower bound (ELBO) on the likelihood.
A widely held assumption is that the ELBO is invariant to the noise process: our work dispels this assumption and proposes multivariate learned adaptive noise (\method), a learned diffusion process that applies noise at different rates across an image. Specifically, our method relies on a multivariate noise schedule that is a function of the data to ensure that the ELBO is no longer invariant to the choice of the noise schedule as in previous works.  Empirically, \method{} sets a new state-of-the-art in density estimation on CIFAR-10 and ImageNet and reduces the number of training steps by 50\%. We provide the code\footnote{\href{https://github.com/s-sahoo/MuLAN}{https://github.com/s-sahoo/MuLAN}}, along with a blog post and video tutorial on the project page:
\looseness=-1

\vspace{0.3ex}
\centerline{\href{https://s-sahoo.com/MuLAN}{https://s-sahoo.com/MuLAN}}
\end{abstract}

\section{Introduction}

Diffusion models, inspired by the physics of heat diffusion, have gained traction as powerful tools for generative modeling, capable of synthesizing realistic, high-quality images \citep{sohldickstein2015deep, ho2020denoising, rombach2021highresolution, gokaslan2024commoncanvas}. 
Central to these algorithms is the diffusion process, a gradual mapping of clean images into white noise.
The reverse of this mapping defines the data-generating process we seek to learn---hence, its choice can significantly impact performance \citep{kingma2023understanding}. 
The conventional approach involves adopting a diffusion process derived from the laws of thermodynamics, which, albeit simple and principled, may be suboptimal due to its lack of adaptability to the dataset.

In this study, we investigate whether the notion of diffusion can be instead {\em learned from data}. Our motivating goal is to perform accurate log-likelihood estimation and probabilistic modelling, and we take an approach grounded in Bayesian inference \citep{kingma2013auto}. We view the diffusion process as an approximate variational posterior: learning this process induces a tighter lower bound (ELBO) on the marginal likelihood of the data. 
Although previous work argued that the ELBO objective of a diffusion model is invariant to the choice of diffusion process \citep{kingma2021variational,kingma2023understanding}, we show that this claim is only true for the simplest types of univariate Gaussian noise: we identify a broader class of noising processes whose optimization yields significant performance gains.

Specifically, we propose a new diffusion process, multivariate learned adaptive noise (MuLAN), which augments classical diffusion models \citep{sohldickstein2015deep,kingma2021variational} with three innovations: a per-pixel polynomial noise schedule, an adaptive input-conditional noising process, and auxiliary latent variables.
In practice, this method learns the schedule by which Gaussian noise is applied to different parts of an image, and allows  tuning this noise schedule to the each image instance. %

Our learned diffusion process yields improved log-likelihood estimates on two standard image datasets, CIFAR10 and ImageNet.
Remarkably, we achieve state-of-the-art performance with less than half of the training time of previous methods.
Our method also does not require any modifications to the underlying UNet architecture, which makes it compatible with most existing diffusion algorithms.

\paragraph{Contributions}
In summary, our paper makes the following contributions:
\begin{enumerate}
    \item We demonstrate that the ELBO of a diffusion model is not invariant to the choice of noise process for many types of noise, thus dispelling a common assumption in the field.
    \item We introduce \method, a learned noise process that adaptively adds multivariate Gaussian noise at different rates across an image in a way that is conditioned on arbitrary context (including the image itself).
    \item We empirically demonstrate that learning the diffusion process  speeds up training and matches the previous state-of-the-art models using \textbf{2x less compute}, and also achieves a new \textbf{state-of-the-art} in density estimation on CIFAR-10 and  ImageNet
\end{enumerate}

\begin{figure}
    \centering
    \includegraphics[width=\linewidth]{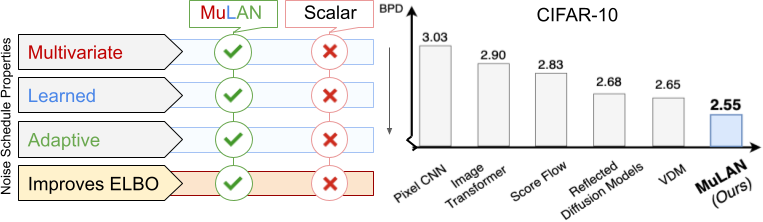}
    \caption{\textit{(Left)} Comparison of noise schedule properties: Multivariate Learned Adaptive Noise schedule (\method{}) (ours) versus a typical scalar noise schedule. Unlike scalar noise schedules, \method{}’s multivariate and input-adaptive properties improve likelihood. \textit{(Right)} Likelihood in bits-per-dimension (BPD) on CIFAR-10 without data augmentation. }
    \label{fig:enter-label}
\end{figure}

\section{Background}\label{sec:background}

A diffusion process $q$ transforms an input datapoint denoted by $\x_0$ and sampled from a distribution $q(\x_0)$ into a sequence of noisy latent variables $\x_t$ for $t \in [0, 1]$ by progressively adding Gaussian noise of increasing magnitude \citep{sohldickstein2015deep, ho2020denoising, song2020score}.
The marginal distribution of each latent is defined by $q(\x_t | \x_0) = \mathcal{N}(\x_t; \alpha_t \x_0,  \sigma_t \mI)$ where the diffusion parameters $\alpha_t, \sigma_t \in \mathbb{R}^+$ implicitly define a noise schedule as a function of $t$, 
such that $\nu(t) = \alpha^2_t / \sigma^2_t$ is a monotonically decreasing function in $t$.
Given any discretization of time into $T$ timesteps of width $1 / T$, we define $\t = i / T$ and $\s = (i - 1) / T$ and we use $\x_{0:1}$ to denote the subset of variables associated with these timesteps; the forward process $q$ can be shown to factorize into a Markov chain $q(\x_{0:1}) = q (\x_0) \prod_{i=1}^{T} q(\x_\t | \x_\s). $

The diffusion model $\p$ is defined by a neural network (with parameters $\theta$) used to denoise the forward process $q$. Given a discretization of time into $T$ steps, $p$ factorizes as $p_\theta(\x_{0:1}) = p_\theta(\x_1) \prod_{i=1}^{T} p_\theta(\x_{\s} | \x_\t).$
We treat the $\x_t$ for $t > 0$ as latent variables and fit $p_\theta$ by maximizing the evidence lower bound (ELBO) on the marginal log-likelihood given by:
\begin{align}
&\log \p (\x_0) = \text{ELBO}(p_\theta, q) + \kl[q(\x_{t(1):t(T)} | \x_0) \| p_\theta(\x_{t(1):t(T)} | \x_0)] \geq \text{ELBO}(p_\theta, q) \label{eqn:diffusion_nelbo} %
\end{align}

In most works, the noise schedule, as defined by $\nu(t)$, is either fixed or treated as a hyper-parameter~\citep{ho2020denoising, chen2023importance, hoogeboom2023simple}. \citet{chen2023importance, hoogeboom2023simple} show that the noise schedule can have a significant impact on sample quality. \citet{kingma2021variational} consider learning $\nu(t)$, but argue that the KL divergence terms in the ELBO are invariant to the choice of function $\nu$, except for the initial values $\nu(0), \nu(1)$, and they set these values to hand-specified constants in their experiments. They only consider learning $\nu$ for the purpose of minimizing the variance of the gradient of the ELBO. In this work, we show that the ELBO is not invariant to more complex forward processes.

\section{Diffusion Models With Multivariate Learned Adaptive Noise}\label{sec:method}

Here, we introduce a new diffusion process, multivariate learned adaptive noise (MuLAN), which introduces three innovations: a per-pixel polynomial noise schedule, a conditional noising process, and auxiliary-variable reverse diffusion.
We describe these below.

\subsection{Why Learned Diffusion?}

Our goal is to perform accurate density estimation and probabilistic modelling, and we take an approach grounded in Bayesian inference \citep{kingma2013auto}.
Notice that the gap between the evidence lower bound $\text{ELBO}(p,q)$ and the marginal log-likelihood (MLL) in 
\Eqn{eqn:diffusion_nelbo} 
is precisely the KL divergence $\kl[q(\x_{t(1):t(T)} | \x_0) \| p_\theta(\x_{t(1):t(T)}|\x_0)]$ between the diffusion process $q$ over the latents $\x_t$ and the true posterior of the diffusion model. 
The diffusion process plays the role of a variational posterior $q$ in $\text{ELBO}(p,q)$; optimizing $q$ thus tightens the gap $(\text{MLL}-\text{ELBO})$.

This observation suggests that the ELBO can be made tighter by choosing a diffusion processes $q$ that is closer to the true posterior $p_\theta(\x_{t(1):t(T)}|\x_0)$.
In fact, the key idea of variational inference is to optimize $\max_{q \in \mathcal{Q}} \text{ELBO}(p,q)$ over a family of approximate posteriors $\mathcal{Q}$ to induce a tighter ELBO \citep{kingma2013auto}. 
Most diffusion algorithms, however optimize $\max_{p \in \mathcal{P}} \text{ELBO}(p,q)$ within some family $\mathcal{P}$ with a fixed $q$.
Our work seeks to jointly optimize $\max_{p \in \mathcal{P}, q \in \mathcal{Q}} \text{ELBO}(p,q)$; we will show in our experiments that this improves the likelihood estimation.

The task of log-likelihood estimation is directly motivated by applied problems such as data compression~\citep{mackay2003information}. In that domain, arithmetic coding techniques can take a generative model and produce a compression algorithm that provably achieves a compression rate (in bits per dimension) that equals the model's log-likelihood~\citep{cover2005data}. 
Other applications of log-likelihood estimation include adversarial example detection~\citep{song2017pixeldefend}, semi-supervised learning~\citep{dai2017good}, and others.

Note that our primary focus is density estimation and probabilistic modeling rather than sample quality. The visual appeal of generated images (as measured by e.g., FID) correlates imperfectly with log-likelihood. We focus here on pushing the state-of-the-art in log-likelihood estimation, and while we report FID for completeness, we defer sample quality optimization to future work.

\subsection{A Forward Diffusion Process With Multivariate Adaptive Noise}\label{subsec:multivariate}

Next, our plan is to define a family of approximate posteriors $\mathcal{Q}$, as well as a family suitably matching reverse processes $\mathcal{P}$, such that the optimization problem $\max_{p \in \mathcal{P}, q \in \mathcal{Q}} \text{ELBO}(p,q)$ is tractable and does not suffer from the aforementioned invariance to the choice of $q$.
This subsection focuses on defining $\mathcal{Q}$; the next sections will show how to parameterize and train a reverse model $p \in \mathcal{P}$.

\textbf{Notation.} Given two vectors $\textbf{a}$ and $\textbf{b}$, we use the notation $\textbf{a}\textbf{b}$ to represent the Hadamard product (element-wise multiplication). Additionally, we denote element-wise division of $\textbf{a}$ by $\textbf{b}$ as $\textbf{a}$ / $\textbf{b}$. We denote the mapping \diag{}(.) that takes a vector as input and produces a diagonal matrix as output.

\subsubsection{Multivariate Gaussian Noise Schedule}

Intuitively, a multivariate noise schedule injects noise at different rates for each pixel of an input image. This enables adapting the diffusion process to spatial variations within the image. We will also see that this change is sufficient to make the ELBO no longer invariant in $q$.

Formally, we define a forward diffusion process with a multivariate noise schedule $q$ via the marginal for each latent noise variable $\x_t$ for $t \in [0,1]$, where the marginal is given by:
\begin{equation}\label{eqn:multivariate_forward}
    q(\x_t | \x_0) = \mathcal{N}(\x_t; \balpha_t \x_0, \diag(\bsigma^2_t)),
\end{equation}
where $\x_t, \x_0 \in \mathbb{R}^{d}$,  $\balpha_t, \bsigma_t \in \mathbb{R}^{d}_{+}$ and $d$ is the dimensionality of the input data. 
The $\balpha_t, \bsigma_t$ denote varying amounts of signal associated with each component (i.e., each pixel) of $\x_0$ as a function of time $\t$.
We define the multivariate signal-to-noise ratio as $\snr(t) = \balpha^2_t / \bsigma^{2}_t$ and choose $\balpha_t, \bsigma_t$ so that $\snr(t)$ decreases monotonically in $t$ along all dimensions and is differentiable in $t \in [0, 1]$.
Let $\balpha_{t|s} = \balpha_t / \balpha_s$ and $\bsigma^2_{t | s} = \bsigma_t^2 - \balpha_{t|s}^2 / \bsigma_s^2$ with all operations applied elementwise.

These marginals induce transition kernels between steps $s < t$  given by (\supp{appendix:equation:q_vdm}):
\begin{align}
q(\x_s | \x_t, \x_0) = \mathcal{N} \left (\x_s; \bmu_q = \frac{\balpha_{t | s}\bsigma^2_{s}}{\bsigma^2_{t}} \x_t + \frac{\bsigma^2_{t | s}\balpha_s}{\bsigma_t^2} \x_0, \; \bSigma_q = \diag\left (\frac{\bsigma^2_s \bsigma^2_{t | s}}{\bsigma_t^2}\right) \right).
\end{align}

In \sec{subsec:path_integral}, we argue that this class of diffusion process $\mathcal{Q}$ induces an ELBO that is not invariant to $q \in \mathcal{Q}$. The ELBO consists of a line integral along the diffusion trajectory specified by $\snr(t)$. A line integrand is almost always path-dependent, unless its integral corresponds to a conservative force field, which is rarely the case for a 
diffusion process ~\citep{spinney2012fluctuation}.  See \sec{subsec:path_integral} for details.

\subsubsection{Adaptive Noise Schedule Conditioned On Context}\label{sec:adaptive}

Next, we extend the diffusion process to support context-adaptive noise. This enables injecting noise in a way that is dependent on the features of an image.
Formally, suppose we have access to a context variable $\c \in \mathbb{R}^m$ which encapsulates high-level information regarding $\x_0$. Examples of $\c$ could be a class label, a vector of attributes (e.g., features characterizing a human face), or even the input $\x_0$ itself. 
We define the marginal of the latent $\x_t$ in the forward process as 
$q(\x_t | \x_0, \c) = \mathcal{N}(\x_t; \balpha_t(\c) \x_0,  \bsigma^2_t(\c))$; the reverse process can be similarly derived (\supp{appendix:equation:q_vdm}) as:
\begin{align}\label{eqn:q_context_conditioned}
q(\x_s | \x_t, \x_0, \c) = \mathcal{N} \left (\bmu_q = \frac{\balpha_{t | s}(\c)\bsigma^2_{s}(\c)}{\bsigma^2_{t}(\c)} \x_t + \frac{\bsigma^2_{t | s}(\c)\balpha_s(\c)}{\bsigma_t^2(\c)} \x_0, \; \bSigma_q = \diag\left(\frac{\bsigma^2_s(\c) \bsigma^2_{t | s}(\c)}{\bsigma_t^2(\c)} \right)\right),
\end{align}
where the diffusion parameters $\balpha_t$, $\bsigma_t$ are now conditioned on $\c$ via a neural network. 

Specifically, we parameterize the diffusion parameters $\balpha_t(\c), \bsigma_t(\c), \snr(t, \c)$ as $\balpha^2_t(\c) = \text{sigmoid}(- \ns_\phi(\c,t))$,
$\bsigma^2_t(\c) = \text{sigmoid}(\ns_\phi(\c, t))$, and 
$\snr(\c, t) = \exp{(- \ns_\phi(\c, t))}$.
Here, $\ns_\phi(\c, t): \mathbb{R}^m \times [0, 1] \to [\gamma_\text{min}, \gamma_\text{max}]^d$ is a neural network with the property that $\ns_\phi(\c, t)$ is monotonic in $t$. Following \citet{kingma2021variational,zheng2023improved}, we set $\gamma_\text{min}=-13.30$, $\gamma_\text{max}=5.0$.

We explore various parameterizations for $\ns_\phi(\c, t)$. These schedules are designed in a manner that guarantees $\ns_\phi(\c, 0) = \gamma_\text{min} \one$ and $\ns_\phi(\c, 1) = \gamma_\text{max} \one$, Below, we list these parameterizations. The polynomial parameterization is novel to our work and yields significant performance gains.

\textbf{Monotonic Neural Network~\citep{kingma2021variational}.} We use the monotonic neural network $\ns_{\text{vdm}}(t)$, proposed in VDM to express $\ns$ as a function of $t$ such that $\ns_{\text{vdm}}(t): [0, 1] \to [\nsmin, \nsmax]^d$. Then we use FiLM conditioning~\citep{perez2018film} in the intermediate layers of this network via a neural network that maps $\z$. The activations of the FiLM layer are constrained to be positive.

\textbf{Polynomial.} (Ours) We express $\ns_\phi(\c, t)$ as a monotonic degree 5 polynomial in $t$
. Details about the exact functional form of this polynomial and its implementation can be found in \supp{appendix:subsec:polynomial_ns}.

\subsection{Auxiliary-Variable Reverse Diffusion Processes}\label{sec:context_diffusion}

In principle, we can fit a normal diffusion model in conjunction with our proposed forward diffusion process. However, variational inference suggests that the variational and the true posterior ought to have the same dependency structure: that is the only way for the KL divergence between these two distributions to be zero.
Thus, we introduce a class of approximate reverse processes $\mathcal{P}$ that match the structure of $\mathcal{Q}$ and that are naturally suitable for joint optimization $\max_{p\in\mathcal{P},q \in \mathcal{Q}}\text{ELBO}(p,q)$.

Formally, we define a diffusion model where the reverse diffusion process is conditioned on the context $\c$. Specifically, given any discretization of $t \in [0,1]$ into $T$ time steps as in \sec{sec:background}, we introduce a context-conditional diffusion model $\p(\x_{0:1} | \c)$ that factorizes as the Markov chain
\begin{equation}
    \p(\x_{0:1} | \c) = \p(\x_1 | \c) \prod_{i=1}^{T} \p(\x_{\s} | \x_\t, \c).
    \label{eqn:conditional_diff}
\end{equation}

Given that the true reverse process is a Gaussian as specified in \Eqn{eqn:q_context_conditioned}, the ideal $p_\theta$ matches this parameterization (the proof mirrors that of regular diffusion models; \supp{appendix:multivariate_noising_schedule_context}), which yields
{\footnotesize
\begin{align}\label{eqn:p_latent}
    & \p(\x_s | \c, \x_t) = \mathcal{N}  \left( \bmu_p = \frac{\balpha_{t | s}(\c)\bsigma^2_{s}(\c)}{\bsigma^2_{t}(\c)} \x_t + \frac{\bsigma^2_{t | s}(\c)\balpha_s(\c)}{\bsigma_t^2(\c)} \x_\theta(\x_t, t), \bSigma_p = \diag \left({\bsigma^2_s(\c) \bsigma^2_{t | s}(\c)}/{\bsigma_t^2(\c)}\right) \right),
\end{align}}
where $\x_\theta(\x_t, t)$, is a neural network that approximates $\x_0$. Instead of parameterizing $\x_\theta (\x_t, t)$ directly using a neural network, we consider two other parameterizations. One is the noise parameterization~\citep{ho2020denoising} where $\unet(\x_t, \c, t)$ is the denoising model which is parameterized as $\unet(\x_t, t) = {(\x_t - \balpha_t(\c) \x_\theta(\x_t, t, \c)) }/{ \bsigma_t(\c)}$; see~\supp{appendix:subsec:noise_param} and the other is v-parameterization~\citep{salimans2022progressive} where $\mathbf{v}_\theta(\x_t, \c, t)$ is a neural network that models $\mathbf{v}_\theta(\x_t, \c, t) = (\balpha_t(\c) \x_t - \x_\theta(\x_t, \c, t)) / \bsigma_t(\c)$; see ~\supp{appendix:subsec:velocity_param}.

\subsubsection{Challenges in Conditioning on Context}\label{subsec:context_challenges}

Note that the model $\p(\x_{0:1}|\c)$ implicitly assumes the availability of $\c$ at generation time. Sometimes, this context may be available, such as when we condition on a label. We may then fit a conditional diffusion process with a standard diffusion objective $\mathbb{E}_{\x_0, c} [\text{ELBO}(\x_0, \p(\x_{0:1}|\c), \q(\x_{0:1}|\c)]$, in which both the forward and the backward processes are conditioned on $\c$ (see \sec{sec:vlb}).

When $\c$ is not known at generation time, we may fit a model $p_\theta$ that does not condition on $\c$.
Unfortunately, this also forces us to define $\p(\x_s | \x_t) = \mathcal{N} (\bmu_p(\x_t, t), \bSigma_p(\x_t, t))$ where $\bmu_p(\x_t, t), \bSigma_p(\x_t, t)$ is parameterized directly by a neural network. We can no longer use a noise parameterization $\unet(\x_t, t) = {(\x_t - \balpha_t(\c) \x_\theta(\x_t, t, \c)) }/{ \bsigma_t(\c)}$ because it requires us to compute $\balpha_t(\c)$ and $\bsigma_t(\c)$, which we do not know. Since noise parameterization plays a key role in the sample quality of diffusion models \citep{ho2020denoising}, this approach limits performance.

\subsubsection{Conditioning Noise on an Auxiliary Latent Variable}\label{subsec:aux_latent}

We propose an alternative strategy for learning conditional forward and reverse processes $p, q$ that feature the same structure and hence support efficient noise parameterization. Our approach is based on the introduction of auxiliary variables \citep{wang2023infodiffusion}, which lift the distribution $p_\theta$ into an augmented latent space.
Experiments (\supp{appendix:subsec:context_ablations}) and theory (\supp{appendix:multivariate_noising_schedule_context}) confirm that this approach performs better than parameterizing $\c$ using a neural network, $\c_\theta(\x_t, t)$.

Specifically, we introduce an auxiliary latent variable  $\z \in \mathbb{R}^m$ and define a lifted $p_\theta(\x, \z) = p_\theta(\x|\z)p_\theta(\z),$ where  $p_\theta(\x|\z)$ is the conditional diffusion model from \Eqn{eqn:conditional_diff} (with context $\c$ set to $\z$) and $p_\theta(\z)$ is a simple prior (e.g., unit Gaussian or fully factored Bernoulli).
The latents $\z$ can be interpreted as a high-level semantic representation of $\x$ that conditions both the forward and the reverse processes. Unlike $\x_{0:1}$, the $\z$ are not constrained to have a particular dimension and can be a low-dimensional vector of latent factors of variation. They can be continuous or discrete. The learning objective for the lifted $p_\theta$ is given by:
\begin{align}
    \log p_\theta(\x_0) 
    & \geq \mathbb{E}_{\q(\z|\x_0)} [ \log p_\theta(\x_0 | \z)] - \kl(\q(\z|\x_0) \| p_\theta(\z)) \\
    & \geq \mathbb{E}_{\q(\z|\x_0)} 
    \text{ELBO}(p_\theta(\x_{0:1} | \z), \q(\x_{0:1}|\z)) - \kl(\q(\z|\x_0) \| p_\theta(\z)), \label{eqn:elbo2x}
\end{align}
where $\text{ELBO}(p_\theta(\x_{0:1} | \z), \q(\x_{0:1}|\z))$ denotes the variational lower bound (VLB) of a diffusion model (defined in \Eqn{eqn:diffusion_nelbo}) with a forward process $\q(\x_{0:1}|\z)$ (defined in \Eqn{eqn:q_context_conditioned} and \sec{sec:adaptive}) and and an approximate reverse process $p_\theta(\x_{0:1} | \z)$ (defined in \Eqn{eqn:conditional_diff}), both conditioned on $\z$. The distribution $\q(\z | \x_0)$ is an approximate posterior for $\z$ parameterized by a neural network with parameters $\phi$.

Crucially, note that in the learning objective (\Eqn{eqn:elbo2x}), the context, which in this case is $\z$, is available at training time in both the forward and reverse processes. At generation time, we can still obtain a valid context vector by sampling an auxiliary latent from $p_\theta(\z)$. Thus, this approach addresses the aforementioned challenges and enables us to use the noise parameterization in \Eqn{eqn:p_latent}.

Although we apply Jensen’s inequality twice to get (\ref{eqn:elbo2x}), this also enables us to learn the noise process, which significantly offsets any potential increase in ELBO gap reduction and improves $\text{ELBO}(p_\theta(\x_{0:1} | \z), \q(\x_{0:1}|\z))$ by optimizing over a more expressive class of posteriors. This claim is empirically validated in \tab{tab:bpd}.
 
\subsection{Variational Lower Bound}\label{sec:vlb}

Next, we derive a precise formula for the learning objective (\ref{eqn:elbo2x}) of the auxiliary-variable diffusion model. 
Using the objective of a diffusion model in (\ref{eqn:diffusion_nelbo}) we can write (\ref{eqn:elbo2x}) as the sum of four terms:
\begin{align}\label{eqn:elbo}
    \log \p (\x_0) \geq \mathbb{E}_{q_\phi} [\lossrecons + \lossdiff + \lossprior + \lossz],
\end{align}

The reconstruction loss, $\lossrecons$, can be (stochastically and differentiably) estimated using standard techniques; see~\citep{kingma2013auto}, $\mathcal{L}_\text{prior} =-\kl[\q(\x_1| \x_0, \z) \| \p(\x_1)]$ is the diffusion prior term, $\mathcal{L}_\text{latent} =-\kl [\q(\z| \x_0) \| \p(\z)]$ is the latent prior term, and
$\mathcal{L}_\text{diffusion}$ is the diffusion loss term, which we examine below.
The complete derivation is given in \supp{appendix:auxiliary_latent_vlb}.

\subsubsection{Diffusion Loss}

\paragraph{Discrete-Time Diffusion.}
We start by defining $p_\theta$ in discrete time, and as in \sec{sec:background}, we let $T > 0$ be the number of total time steps and define $\t = i / T$ and $\s = (i - 1) / T$ as indexing variables over the time steps. We also use $\x_{0:1}$ to denote the subset of variables associated with these timesteps.
Starting with the expression in \Eqn{eqn:diffusion_nelbo} and following the steps in \supp{appendix:mulan}, we can write $\lossdiff$ as:
\begin{align}\label{eqn:diffusion_loss_discrete}
    \lossdiff &=
    -\sum_{i=2}^{T}\kl [ \q(\x_{\s} | \x_{\t},  \x_0, \z) \| \p(\x_{\s} | \x_{\t}, \z)] \nonumber \\
    & = \frac{1}{2} \sum_{i=2}^T
    [
        (\noise_t - \unet(\x_t, \z, \t))^\top  \diag \left( \ns(\z, \s) - \ns(\z, \t) \right)  (\noise_t - \unet(\x_t, \z, \t))
    ]
\end{align}

\paragraph{Continuous-Time Diffusion.}
We can also consider the limit of the above objective as we take an infinitesimally small partition of $t \in [0,1]$, which corresponds to the limit when $T \to \infty$. In \supp{appendix:mulan} we show that taking this limit of \Eqn{eqn:diffusion_loss_discrete} yields the continuous-time diffusion loss:
\begin{align}\label{eqn:diffusion_loss_continuous}
    \lossdiff 
    & = -\frac{1}{2} \mathbb{E}_{t \sim [0, 1]} 
    [
        (\noise_t - \unet(\x_t, \z, t))^\top   \diag \left( \nabla_t \ns(\z, t) \right)   (\noise_t - \unet(\x_t, \z, t))
    ] 
\end{align}
where $\nabla_t\ns(\z, t) \in \mathbb{R}^d$ denotes the Jacobian of $\ns(\z, t)$ with respect to the scalar $t$. We observe that the limit of $T \to \infty$ yields improved performance, matching the existing theoretical argument by \citet{kingma2021variational}.

\subsubsection{Auxiliary latent loss}
We try two different kinds of priors for $\p(\z)$: discrete ($\z \in \{0, 1\}^m$) and continuous ($\z \in \mathbb{R}^m$).
\paragraph{Continuous Auxiliary Latents.}
In the case where $\z$ is continuous, we select $\p(\z)$ as $\mathcal{N}(\mathbf{0}, \bfI_m)$. This leads to the following KL loss term:\\
$\kl(\q(\z | \x_0) \| \p(\z)) = \frac{1}{2}(\bmu^\top (\x_0) \bmu(\x_0)) + \text{tr}(\bSigma^2(\x_0) - \bfI_m) - \log|\bSigma^2(\x_0)|)$.

\paragraph{Discrete  Auxiliary Latents.}
In the case where $\z$ is discrete, we select $\p(\z)$ as a uniform distribution.
Let $\z \in \{0,1\}^m$ be a $k$-hot vector sampled from a discrete Exponential Family distribution $\p(\z; \theta)$ with logits $\theta$. \citet{niepert2021implicit} show that $\z \sim \p(\z; \theta)$ is equivalent to $\z = \arg\max_{y \in Y}\langle \theta + \epsilon_g, y \rangle$ where $\epsilon_g$ denotes the sum of gamma distribution~\supp{appendix:eqn:sog}, $Y$ denotes the set of all $k$-hot vectors of some fixed length $m$. For $k > 1$, To differentiate through the $\arg\max$ we use a relaxed estimator, Identity, as proposed by \citet{sahoo2023backpropagation}. This leads to the following KL loss term:
$\kl(\q(\z | \x_0) \| \p(\z))= - \sum_{i=1}^{m} \q(\z | \x_0)_i (\log \q(\z | \x_0)_i + \log m)$.

\subsection{The Variational Lower Bound as a Line Integral Over The Noise Schedule}\label{subsec:path_integral}
Having defined our loss, we now return to the question of whether it is invariant to the choice of diffusion process.
Notice that we may rewrite \Eqn{eqn:diffusion_loss_continuous} in the following vectorized form: 
\begin{align}\label{eqn:dot_product}
    \lossdiff & = - \frac{1}{2} \int_{0}^{1}
        (\x_0 - \x_\theta(\x_t, \z, t))^2 \cdot \nabla_t \vsnr(\z, t) \text{d}t
\end{align}
where the square is applied elementwise.
We seek to rewrite (\ref{eqn:dot_product}) as a line integral $\int_{a}^{b}\f(\r(t))\cdot \frac{\text{d}}{\text{d}t}\r(t)\text{d}t$ for some vector field $\f$ and trajectory $\r(t)$. 
Recall that $\vsnr(\z, t)$ is monotonically decreasing in each coordinate as a function of $t$; hence, it is invertible on its image, and we can write $t = \vsnr_\z^{-1}(\vsnr(\z, t))$ for some $\vsnr_z^{-1}$.
Let $\bar \x_\theta(\x_{\vsnr(\z, t)}, \z, \vsnr(\z, t)) \equiv \x_\theta(\x_{\vsnr_z^{-1}(\vsnr(\z, t))}, \z, \vsnr_\z^{-1}(\vsnr(\z, t)))$ and note that for all $t$, we can write $\x_t$ as $\x_{\snr(\z, t)}$; see \Eqn{eqn:x_snr}, and have $\bar \x_\theta(\x_{\vsnr(\z, t)}, \z, \vsnr(\z, t)) \equiv \x_\theta(\x_t, \z, t)$.
We can then write the integral in (\ref{eqn:dot_product}) as $\int_{0}^{1}
(\x_0 - \bar \x_\theta(\x_{\vsnr(\z, t)}, \z, \vsnr(\z, t)))^2 \cdot \frac{\text{d}}{\text{d}t} \vsnr(\z, t) \rangle \text{d}t$, which is a line integral with $\f(\r(t)) \equiv (\x_0 - \bar \x_\theta(\x_{\snr(\z, t)}, \z, \vsnr(\z, t)))^2$ and $\r(t) \equiv \vsnr(\z, t)$.

\paragraph{Intuitive explanation.} 
Imagine piloting a plane across a region with cyclones and strong winds, as shown in~\fig{fig:intuitive-explanation}. Plotting a direct, straight-line course through these adverse weather conditions requires more fuel and effort due to increased resistance. By navigating around the cyclones and winds, however, the plane reaches its destination with less energy, even if the route is longer.

This intuition translates into mathematical and physical terms. The plane’s trajectory is denoted by $\mathbf{r}(t) \in \mathbb{R}^n_{+}$, while the forces acting on it are represented by $\mathbf{f}(\mathbf{r}(t)) \in \mathbb{R}^n$. The work required to navigate is given by $\int_{0}^{1} \mathbf{f}(\mathbf{r}(t)) \cdot \frac{d}{dt}\mathbf{r}(t) , dt$. Here, the work depends on the trajectory because $\mathbf{f}(\mathbf{r}(t))$ is not a conservative field.

This concept also applies to the diffusion NELBO. From~\Eqn{eqn:dot_product}, it’s clear that the trajectory $\mathbf{r}(t)$ is parameterized by the noise schedule $\snr(\z, t)$, which is influenced by complex forces, $\f$ (analogous to weather patterns), represented by the dimension-wise reconstruction error of the denoising model, $(\x_0 - \x_\theta(\x_t, \z, t))^2$. Thus, the diffusion loss, $\lossdiff$, can be interpreted as the work done along the trajectory $\snr(\z, t)$ in the presence of these vector field forces $\f$. By learning the noise schedule, we can avoid “high-resistance” paths (those where the loss accumulates rapidly), thereby minimizing the overall “energy” expended, as measured by the NELBO. 
Since the diffusion process corresponds to non-conservative force fields, as noted in \citet{spinney2012fluctuation}, different noise schedules should yield different NELBOs—a result supported by our empirical findings. In \supp{appendix:recover_vdm_loss_from_path_integral}, we show that variational diffusion models are limited to linear trajectories $\vsnr(t)$, rendering their objective invariant to the noise schedule. In contrast, our approach learns a multivariate $\vsnr$, enabling paths that achieve a better ELBO.

\section{Experiments}

This section reports experiments on the CIFAR-10~\citep{krizhevsky2009learning} and  ImageNet-32~\citep{van2016pixel} datasets. We don't employ data augmentation and we use the same architecture and settings as in the VDM model \citep{kingma2021variational}.
The encoder, $\q(\z | \x)$, is modeled using a sequence of 4 ResNet blocks which is much smaller than the denoising network that uses 32 such blocks (i.e., we increase parameter count by only about 10\%); the noise schedule $\ns_\phi$ is modeled using a two-layer MLP.
In all our experiments, we use discrete auxiliary latents with $m=50$ and $k=15$. A detailed description can be found in \supp{section:experiment_details}.

\subsection{Training Speed} 
In these experiments, we replace VDM's noise process with \method{}.
On CIFAR-10, \method{} \textbf{attains VDM's likelihood score of 2.65 in just 2M steps, compared to VDM's 10M steps} \ref{tab:vdm_and_mulan}). When trained on 4 V100 GPUs, VDM achieves a training rate of 2.6 steps/second, while \method{} trains slightly slower at 2.24 steps/second due to the inclusion of an additional encoder network. However, despite this slower training pace, VDM requires 30 days to reach a BPD of 2.65, whereas Mulan achieves the same BPD within a significantly shorter timeframe of 10 days. On ImageNet-32, VDM integrated with \method{} reaches a likelihood of 3.71 in half the time, \textbf{achieving this score in 1M steps versus the 2M steps required by VDM}. 

\subsection{Likelihood Estimation}

In \tab{tab:bpd}, we also compare \method{} with other recent methods on CIFAR-10 and ImageNet-32. 
\method{} was trained using v-parameterization for $8$M steps on CIFAR-10 and $2$M steps on Imagenet-32.
During inference, we extract the underlying probability flow ODE and use it to estimate the log-likelihood; see~\supp{appendix:subsec:mulan_ode}. Our algorithm \textbf{establishes a new state-of-the-art in density estimation} on both ImageNet-32 and CIFAR-10.
In Table \ref{tab:dequantization}, we also compute variational lower bounds (VLBs) of $\leq$2.59 and $\leq$3.71 on CIFAR-10 and ImageNet, respectively. Each bound improves over published results (\tab{tab:bpd}); our true NLLs (via flow ODEs) are even lower.

\begin{table}[]
    \centering
    \caption{Likelihood in bits per dimension (BPD) based on the Variational Lower Bound (VLB) estimate (\supp{appendix:subsec:vlb}), sample quality (FID scores) and number of function evaluations (NFE) on CIFAR-10, for vanilla VDM and VDM when endowed with \method{}. FID and NFE were computed for 10k samples generated using an adaptive-step ODE solver. Both methods use noise parameterization~(\supp{appendix:subsec:noise_param}).}
    \begin{tabular}{lcccc|cccc}
        \toprule
        \multirow{2}{*}{Model}
        & \multicolumn{4}{c}{CIFAR-10} & \multicolumn{4}{c}{ImageNet} \\
        &  Steps & VLB ($\downarrow$) & FID ($\downarrow$) & NFE ($\downarrow$) &  Steps & VLB ($\downarrow$) & FID ($\downarrow$) & NFE ($\downarrow$)\\
        \midrule 
        VDM \cite{kingma2021variational}             & 10M & 2.65 & 23.91 & 56 & 2M & 3.72 & 14.26 & \textbf{56}\\
        \hspace{0.1cm} + \method{} & \textbf{2M} & 2.65 & 18.54 & 55 & \textbf{1M} & 3.72 & 15.00 & 62\\
        \hspace{0.1cm} + \method{} & 10M & \textbf{2.60} & \textbf{17.62} & \textbf{50} & 2M & \textbf{3.71} & \textbf{13.19} & 62\\
        \bottomrule
    \end{tabular}
    \label{tab:vdm_and_mulan}
\end{table}

\begin{table*}[t]
    \centering
    \caption{Likelihood in bits per dimension (BPD) on the test set of CIFAR-10 and ImageNet. Results with “/” means they are not reported in the original papers. Model types are autoregressive (AR), normalizing flows (Flow), diffusion models (Diff). We only compare with results achieved {without data augmentation}.
    }
    \begin{tabular}{lccc}
         \toprule
         Model & Type & CIFAR-10 ($\downarrow$)&  ImageNet ($\downarrow$) \\
         \midrule
         PixelCNN~\citep{van2016conditional} & AR & 3.03& 3.83\\
         Image Transformer~\citep{parmar2018image} & AR & 2.90 & 3.77\\
         DDPM~\citep{ho2020denoising} & Diff & $\leq3.69$ & / \\
         ScoreFlow~\citep{song2021maximum} & Diff & 2.83 & 3.76 \\
         VDM~\citep{kingma2021variational}& Diff & $\leq2.65$ & $\leq3.72$ \\
         Flow Matching~\citep{lipman2022flow} & Flow & 2.99 & / \\
         Reflected Diffusion Models~\citep{lou2023reflected} & Diff & 2.68 & 3.74 \\
         \hline
         \method{} (\textbf{Ours})  &  Diff & \textbf{2.55} $\pm 10^{-3}$ & \textbf{3.67} $\pm 10^{-3}$\\
         \bottomrule
    \end{tabular}
    \label{tab:bpd}
\end{table*}

\subsection{Alternative Learned Diffusion Methods}
\begin{wraptable}{r}{0.3\textwidth}
    \vspace{-1.4em}
    \centering
    \caption{Likelihood in bits per dimension (bpd) on CIFAR-10 for learned diffusion methods.
    }
    \begin{tabular}{lc}
        \toprule
        Model & NLL ($\downarrow$)  \\
        \midrule 
        DNF~\citep{zhang2021diffusion} & 3.04 \\
        NDM~\citep{bartosh2023neural} & $\leq2.70$ \\
        DiffEnc~\citep{nielsen2023diffenc} & $\leq2.62$ \\
        \hline
        \method{}  & \textbf{2.55} \\
        \bottomrule
    \end{tabular}
    \label{tab:mulan_vs_learned_forward}
\end{wraptable}
Concurrent work that seeks to improve log-likelihood estimation by learning the forward diffusion process includes Neural Diffusion Models (NDMs)~\citep{bartosh2023neural} and DiffEnc~\citep{nielsen2023diffenc}. In NDMs, the noise schedule is fixed, but the mean of each marginal $q(\x_t|x_0)$ is learned, while DiffEnc adds a correction term to $q$. Diffusion normalizing flows (DNFs) represent an earlier effort where $q$ is a normalizing flow trained by backpropagating through sampling. In~\tab{tab:mulan_vs_learned_forward}, we compare against NDMs, DiffEnc, and DNFs on the CIFAR-10 dataset, using the authors' published results; note that their published ImageNet numbers are either not available or are reported on a different dataset version that is not comparable. Our approach to learned diffusion outperforms previous and concurrent work.

\subsection{Ablation Analysis And Additional Experiments}
\begin{figure}
    \centering
    \begin{subfigure}[t]{.47\linewidth}
    \includegraphics[width=\linewidth]{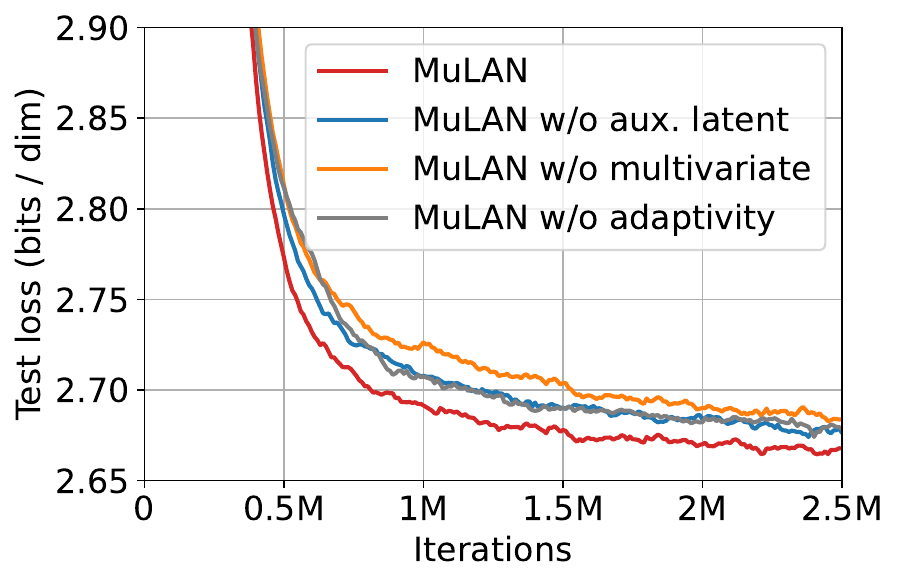}
    \caption{In \method{} w/o aux.~latent, the noise isn't conditioned on a latent. \method{} w/o multivariate uses a scalar noise schedule. \method{} w/o adaptivity has a linear schedule and no auxiliary latents.
    }
    \label{fig:mulan_ablations_left}
    \end{subfigure}\hfill
    \begin{subfigure}[t]{.47\linewidth}
    \includegraphics[width=\linewidth]{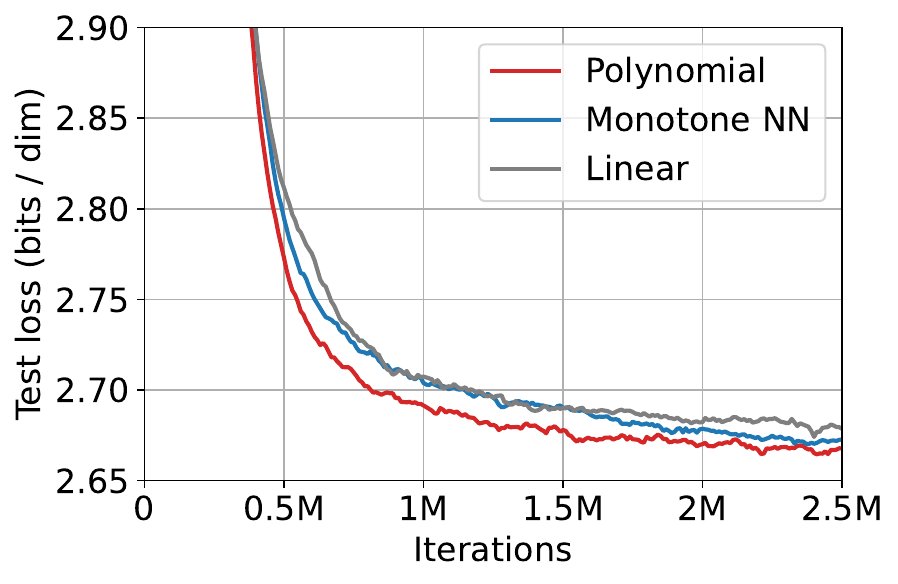}
    \caption{\method{} with different noise schedule parameterizations: polynomial, monotonic neural network, and linear. Our proposed polynomial parameterization performs the best.}
    \label{fig:intuitive-updates-b}
    \end{subfigure}
    \caption{Ablating components of \method{} on CIFAR-10 over 2.5M steps with batch size of 64.}
    \label{fig:mulan_ablations_right}
		\vspace{-0.5em}
\end{figure}

Due to the expensive cost of training, we only performed ablation studies on CIFAR-10 with a reduced batch size of $64$ and trained the model for $2.5$M training steps.
In~\fig{fig:mulan_ablations_left} we ablate each component of \method{}: when we remove the conditioning on an auxiliary latent space from \method{} so that we have a multivariate noise schedule that is solely conditioned on time $t$, our performance becomes comparable to that of VDM, on which our model is based. 
Changing to a scalar noise schedule based on latent variable $\z$ initially underperforms compared to VDM. This drop aligns with our likelihood formula (\Eqn{eqn:p_latent}) which includes $\kl(\q(\z|\x_0) | p_\theta(\z))$, an extra term not in VDM. The input-conditioned scalar schedule doesn't offer any advantage over the scalar schedule used in VDM. This is due to the reasons outlined in ~\sec{subsec:path_integral}. %

\paragraph{Perceptual Quality}

While perceptual quality is not the focus of this work, we report FID numbers for the VDM model and MuLAN (Table \ref{tab:vdm_and_mulan}).
We use RK45 ODE solver 
to generate samples by solving the reverse time Flow ODE~(\Eqn{eqn:ode_reverse_simple}). 
We observe that MuLAN does not degrade FIDs, while improving log-likelihood estimation. Note that MuLAN does not incorporate many tricks that improve FID such as exponential moving averages, truncations, specialized learning schedules, etc.; our FID numbers can be improved in future work using these techniques.

\paragraph{Loss curves for different noise schedules.}
We investigate different parameterizations of the noise schedule in~\fig{fig:intuitive-updates-b}. Among polynomial, linear, and monotonic neural network, we find that the polynomial parameterization yields the best performance. The polynomial noise schedule is a novel component introduced in our work. The reason why a polynomial function works better than a linear or a monotonic neural network as proposed by VDM is rooted in Occam’s razor. In \supp{appendix:subsec:polynomial_ns}, we show that a degree 5 polynomial is the simplest polynomial that satisfies several desirable properties, including monotonicity and having a derivative that equals zero exactly twice. 
More expressive models (e.g., monotonic 3-layer MLPs) are more difficult to optimize. %

\paragraph{Examining the noise schedule.}
\begin{wrapfigure}{r}{0.48\textwidth}
    \centering
    \vspace{-1.5em}
    \includegraphics[width=0.48\textwidth]{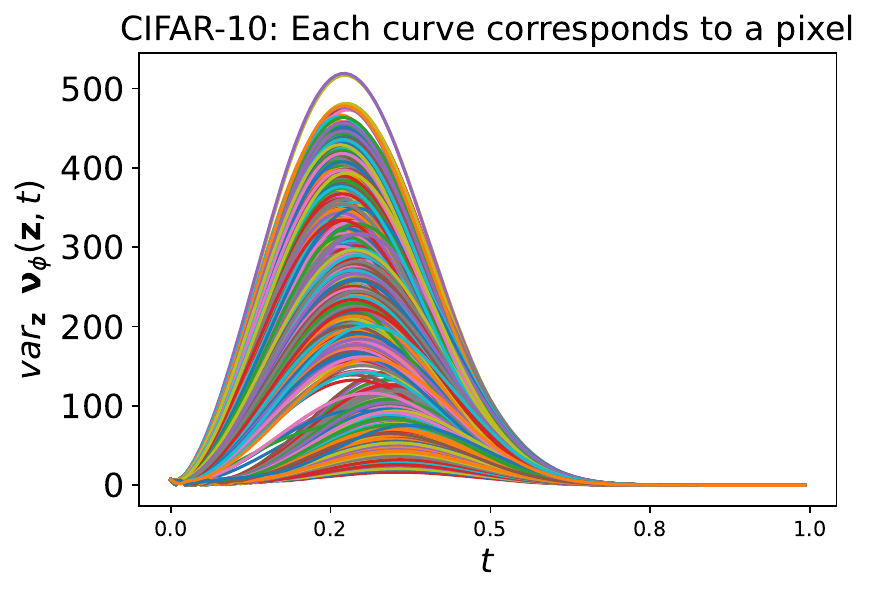}
    \caption{Noise schedule visualizations for \method{} on CIFAR-10. In this figure, we plot the variance of $\snr_\phi(\z, t)$ across different $\z \sim \p(\z)$ where each curve represents the SNR corresponding to an input dimension.}
    \label{fig:mulan_ablations}
    \vspace{-1em}
\end{wrapfigure}
Since the noise schedule, $\ns_\phi(\z, t)$ is multivariate, we expect to learn different noise schedules for different input dimensions and different inputs $\z \sim \p(\z)$. In~\fig{fig:mulan_ablations}, we take our best trained model on CIFAR-10 and visualize the variance of the noise schedule at each point in time for different pixels, where the variance is taken on 128 samples $\z \sim p_\theta(\z)$.

We note an increased variation in the early portions of the noise schedule. However, on an absolute scale, the variance of this noise is smaller than we expected. We also tried to visualize noise schedules across different dataset images and across different areas of the same image; refer to ~\fig{fig:noise_schedules_all}. We also generated synthetic datasets in which each datapoint contained only high frequencies or only low frequencies, and with random masking applied to parts of the data points; see~\supp{appendix:datasets}. Surprisingly, none of these experiments revealed human-interpretable patterns in the learned schedule, although we did observe clear differences in likelihood estimation. We hypothesize that other architectures and other forms of conditioning may reveal interpretable patterns of variation; however, we leave this exploration to future work.

\paragraph{Replacing the noise schedules in a trained denoising model.}\label{subsec:unet_with_diff_ns}

We also confirm experimentally our claim that the learning objective is not invariant to the multivariate noise schedule.
We replace the noise schedule in the trained denoising model with two alternatives: \method{}
with scalar noise schedule, and a linear noise schedule: $\ns_\phi(\z, t) = \nsmin + t(\nsmax - \nsmin)\one$; see ~\citet{kingma2021variational}. For both the noise schedules the likelihood reduces to the same value as that of the VDM: $2.65$.

\section{Related Work}

Diffusion models have emerged in recent years as powerful tools for modeling complex distributions ~\citep{sohldickstein2015deep, ho2020denoising}, extending flow-based methods \citep{song2020score,kingma2018glow,si2022autoregressive,si2023semi} 
 The noise schedule, which determines the amount and type of noise added at each step, plays a critical role in diffusion models. 
\citet{chen2023importance} empirically demonstrate that different noise schedules can significantly impact the generated image quality using various handcrafted noise schedules. \citet{kingma2021variational} showed that the likelihood of a diffusion model remains invariant to the noise schedule with a scalar noise schedule. In this work we show that the ELBO is no longer invariant to multivariate noise schedules.

Recent works explored multivariate noise schedules (including blurring, masking, etc.) \citep{hoogeboom2022blurring, rissanen2022generative, pearl2023svnr, du2023flexible}, yet none have delved into learning the noise schedule conditioned on the input data itself. Likewise, conditional noise processes are typically not learned \citep{lee2021priorgrad, popov2021grad, yang2024cross} and their conditioner (e.g., a prompt) is always available. Auxiliary variable models \citep{yang2023lossy, wang2023infodiffusion} add semantic latents in $p$, but not in $q$, and they don't condition or learn $q$. In contrast, we learn multivariate noise conditioned on latent context.

Diffusion normalizing flows (DNFs)~\citep{zhang2021diffusion} learn a $q$ parameterized by a normalizing flow; however, such $q$ do not admit tractable marginals and require sampling full data-to-noise trajectories from $q$, which is expensive. Concurrent work on neural diffusion models (NDMs) and DiffEnc admits tractable marginals $q$ with learned means and univariate schedules; this yields more expressive $q$ than ours but requires computing losses in a modified space that precludes using a noise parameterization and certain sampling strategies. Empirically, MuLAN performs better with fewer parameters (\supp{appendix:previous}).

Optimal transport techniques seek to learn a noise process that minimizes the transport cost from data to noise, which in practice produces smoother diffusion trajectories that facilitate sampling. Schrondinger bridges~\citep{shi2024diffusion, de2021diffusion,wang2021deep,peluchetti2023diffusion} learn expressive $q$ do not admit analytical marginals, require computing full data-to-noise trajectories and involve iterative optimization (e.g., sinkhorn), which can be slow. Rectification \cite{lee2023minimizing} seeks diffusion paths that are close to linear; this improves sampling, while our method chooses paths that improve log-likelihood.
See \supp{appendix:previous} for more detailed comparisons.

\section{Conclusion}
We introduced \method{}, a context-adaptive noise process that applies Gaussian noise at varying rates across input data. Our theory challenges the prevailing notion that the likelihood of diffusion models is independent of the noise schedule: this independence only holds true for univariate schedules.
Our evaluation of \method{} spans multiple image datasets, where it outperforms state-of-the-art generative diffusion models. We hope our work will motivate further research into the design of noise schedules, not only for improving likelihood estimation but also to improve image quality generation \citep{parmar2018image,song2020score}. A stronger fit to the data distribution also holds promise for improving downstream applications of generative modeling, e.g., compression or decision-making \citep{DBLP:conf/icml/NguyenG22,deshpande2022deep,deshpande2023calibrated,rastogi2023semiparametric}.

\section*{Acknowledgments and Disclosure of Funding}
This work was partially funded by Volodymyr Kuleshov's the National Science Foundation under awards DGE-1922551,
CAREER awards 2046760 and 2145577, and the National Institute of Health under award MIRA R35GM151243, and by Christopher De Sa's NSF RI-CAREER award 2046760.

\bibliographystyle{neurips_2024}
\bibliography{refs}

\begin{thebibliography}{65}
\providecommand{\natexlab}[1]{#1}
\providecommand{\url}[1]{\texttt{#1}}
\expandafter\ifx\csname urlstyle\endcsname\relax
  \providecommand{\doi}[1]{doi: #1}\else
  \providecommand{\doi}{doi: \begingroup \urlstyle{rm}\Url}\fi

\bibitem[Bartosh et~al.(2023)Bartosh, Vetrov, and Naesseth]{bartosh2023neural}
Grigory Bartosh, Dmitry Vetrov, and Christian~A Naesseth.
\newblock Neural diffusion models.
\newblock \emph{arXiv preprint arXiv:2310.08337}, 2023.

\bibitem[Chen et~al.(2018)Chen, Rubanova, Bettencourt, and Duvenaud]{chen2018neural}
Ricky~TQ Chen, Yulia Rubanova, Jesse Bettencourt, and David~K Duvenaud.
\newblock Neural ordinary differential equations.
\newblock \emph{Advances in neural information processing systems}, 31, 2018.

\bibitem[Chen(2023)]{chen2023importance}
Ting Chen.
\newblock On the importance of noise scheduling for diffusion models.
\newblock \emph{arXiv preprint arXiv:2301.10972}, 2023.

\bibitem[Cover \& Thomas(2005)Cover and Thomas]{cover2005data}
Thomas~M Cover and Joy~A Thomas.
\newblock Data compression.
\newblock \emph{Elements of Information Theory}, pp.\  103--158, 2005.

\bibitem[Dai et~al.(2017)Dai, Yang, Yang, Cohen, and Salakhutdinov]{dai2017good}
Zihang Dai, Zhilin Yang, Fan Yang, William~W Cohen, and Russ~R Salakhutdinov.
\newblock Good semi-supervised learning that requires a bad gan.
\newblock \emph{Advances in neural information processing systems}, 30, 2017.

\bibitem[De~Bortoli et~al.(2021)De~Bortoli, Thornton, Heng, and Doucet]{de2021diffusion}
Valentin De~Bortoli, James Thornton, Jeremy Heng, and Arnaud Doucet.
\newblock Diffusion schr{\"o}dinger bridge with applications to score-based generative modeling.
\newblock \emph{Advances in Neural Information Processing Systems}, 34:\penalty0 17695--17709, 2021.

\bibitem[Deng et~al.(2009)Deng, Dong, Socher, Li, Li, and Fei-Fei]{5206848}
Jia Deng, Wei Dong, Richard Socher, Li-Jia Li, Kai Li, and Li~Fei-Fei.
\newblock Imagenet: A large-scale hierarchical image database.
\newblock In \emph{2009 IEEE Conference on Computer Vision and Pattern Recognition}, pp.\  248--255, 2009.
\newblock \doi{10.1109/CVPR.2009.5206848}.

\bibitem[Deshpande \& Kuleshov(2023)Deshpande and Kuleshov]{deshpande2023calibrated}
Shachi Deshpande and Volodymyr Kuleshov.
\newblock Calibrated uncertainty estimation improves bayesian optimization, 2023.

\bibitem[Deshpande et~al.(2022)Deshpande, Wang, Sreenivas, Li, and Kuleshov]{deshpande2022deep}
Shachi Deshpande, Kaiwen Wang, Dhruv Sreenivas, Zheng Li, and Volodymyr Kuleshov.
\newblock Deep multi-modal structural equations for causal effect estimation with unstructured proxies.
\newblock \emph{Advances in Neural Information Processing Systems}, 35:\penalty0 10931--10944, 2022.

\bibitem[Dhariwal \& Nichol(2021)Dhariwal and Nichol]{dhariwal2021diffusion}
Prafulla Dhariwal and Alexander Nichol.
\newblock Diffusion models beat gans on image synthesis.
\newblock \emph{Advances in Neural Information Processing Systems}, 34:\penalty0 8780--8794, 2021.

\bibitem[Dormand \& Prince(1980)Dormand and Prince]{DORMAND198019}
J.R. Dormand and P.J. Prince.
\newblock A family of embedded runge-kutta formulae.
\newblock \emph{Journal of Computational and Applied Mathematics}, 6\penalty0 (1):\penalty0 19--26, 1980.
\newblock ISSN 0377-0427.
\newblock \doi{https://doi.org/10.1016/0771-050X(80)90013-3}.
\newblock URL \url{https://www.sciencedirect.com/science/article/pii/0771050X80900133}.

\bibitem[Du et~al.(2023)Du, Zhang, Yang, and Du]{du2023flexible}
Weitao Du, He~Zhang, Tao Yang, and Yuanqi Du.
\newblock A flexible diffusion model.
\newblock In \emph{International Conference on Machine Learning}, pp.\  8678--8696. PMLR, 2023.

\bibitem[Efraimidis \& Spirakis(2006)Efraimidis and Spirakis]{EFRAIMIDIS2006181}
Pavlos~S. Efraimidis and Paul~G. Spirakis.
\newblock Weighted random sampling with a reservoir.
\newblock \emph{Information Processing Letters}, 97\penalty0 (5):\penalty0 181--185, 2006.
\newblock ISSN 0020-0190.
\newblock \doi{https://doi.org/10.1016/j.ipl.2005.11.003}.
\newblock URL \url{https://www.sciencedirect.com/science/article/pii/S002001900500298X}.

\bibitem[Gokaslan et~al.(2024)Gokaslan, Cooper, Collins, Seguin, Jacobson, Patel, Frankle, Stephenson, and Kuleshov]{gokaslan2024commoncanvas}
Aaron Gokaslan, A~Feder Cooper, Jasmine Collins, Landan Seguin, Austin Jacobson, Mihir Patel, Jonathan Frankle, Cory Stephenson, and Volodymyr Kuleshov.
\newblock Commoncanvas: Open diffusion models trained on creative-commons images.
\newblock In \emph{Proceedings of the IEEE/CVF Conference on Computer Vision and Pattern Recognition}, pp.\  8250--8260, 2024.

\bibitem[Grathwohl et~al.(2018)Grathwohl, Chen, Bettencourt, Sutskever, and Duvenaud]{grathwohl2018ffjord}
Will Grathwohl, Ricky~TQ Chen, Jesse Bettencourt, Ilya Sutskever, and David Duvenaud.
\newblock Ffjord: Free-form continuous dynamics for scalable reversible generative models.
\newblock \emph{arXiv preprint arXiv:1810.01367}, 2018.

\bibitem[Ho et~al.(2020)Ho, Jain, and Abbeel]{ho2020denoising}
Jonathan Ho, Ajay Jain, and Pieter Abbeel.
\newblock Denoising diffusion probabilistic models.
\newblock \emph{Advances in Neural Information Processing Systems}, 33:\penalty0 6840--6851, 2020.

\bibitem[Hoogeboom \& Salimans(2022)Hoogeboom and Salimans]{hoogeboom2022blurring}
Emiel Hoogeboom and Tim Salimans.
\newblock Blurring diffusion models.
\newblock \emph{arXiv preprint arXiv:2209.05557}, 2022.

\bibitem[Hoogeboom et~al.(2023)Hoogeboom, Heek, and Salimans]{hoogeboom2023simple}
Emiel Hoogeboom, Jonathan Heek, and Tim Salimans.
\newblock simple diffusion: End-to-end diffusion for high resolution images.
\newblock \emph{arXiv preprint arXiv:2301.11093}, 2023.

\bibitem[Hutchinson(1989)]{hutchinson1989stochastic}
Michael~F Hutchinson.
\newblock A stochastic estimator of the trace of the influence matrix for laplacian smoothing splines.
\newblock \emph{Communications in Statistics-Simulation and Computation}, 18\penalty0 (3):\penalty0 1059--1076, 1989.

\bibitem[Kingma et~al.(2021)Kingma, Salimans, Poole, and Ho]{kingma2021variational}
Diederik Kingma, Tim Salimans, Ben Poole, and Jonathan Ho.
\newblock Variational diffusion models.
\newblock \emph{Advances in neural information processing systems}, 34:\penalty0 21696--21707, 2021.

\bibitem[Kingma \& Ba(2014)Kingma and Ba]{kingma2014adam}
Diederik~P Kingma and Jimmy Ba.
\newblock Adam: A method for stochastic optimization.
\newblock \emph{arXiv preprint arXiv:1412.6980}, 2014.

\bibitem[Kingma \& Gao(2023)Kingma and Gao]{kingma2023understanding}
Diederik~P Kingma and Ruiqi Gao.
\newblock Understanding the diffusion objective as a weighted integral of elbos.
\newblock \emph{arXiv preprint arXiv:2303.00848}, 2023.

\bibitem[Kingma \& Welling(2013)Kingma and Welling]{kingma2013auto}
Diederik~P Kingma and Max Welling.
\newblock Auto-encoding variational bayes.
\newblock \emph{arXiv preprint arXiv:1312.6114}, 2013.

\bibitem[Kingma \& Dhariwal(2018)Kingma and Dhariwal]{kingma2018glow}
Durk~P Kingma and Prafulla Dhariwal.
\newblock Glow: Generative flow with invertible 1x1 convolutions.
\newblock \emph{Advances in neural information processing systems}, 31, 2018.

\bibitem[Krizhevsky et~al.(2009)Krizhevsky, Hinton, et~al.]{krizhevsky2009learning}
Alex Krizhevsky, Geoffrey Hinton, et~al.
\newblock Learning multiple layers of features from tiny images.
\newblock 2009.

\bibitem[Lee et~al.(2021)Lee, Kim, Shin, Tan, Liu, Meng, Qin, Chen, Yoon, and Liu]{lee2021priorgrad}
Sang-gil Lee, Heeseung Kim, Chaehun Shin, Xu~Tan, Chang Liu, Qi~Meng, Tao Qin, Wei Chen, Sungroh Yoon, and Tie-Yan Liu.
\newblock Priorgrad: Improving conditional denoising diffusion models with data-dependent adaptive prior.
\newblock \emph{arXiv preprint arXiv:2106.06406}, 2021.

\bibitem[Lee et~al.(2023)Lee, Kim, and Ye]{lee2023minimizing}
Sangyun Lee, Beomsu Kim, and Jong~Chul Ye.
\newblock Minimizing trajectory curvature of ode-based generative models.
\newblock In \emph{International Conference on Machine Learning}, pp.\  18957--18973. PMLR, 2023.

\bibitem[Lipman et~al.(2022)Lipman, Chen, Ben-Hamu, Nickel, and Le]{lipman2022flow}
Yaron Lipman, Ricky~TQ Chen, Heli Ben-Hamu, Maximilian Nickel, and Matt Le.
\newblock Flow matching for generative modeling.
\newblock \emph{arXiv preprint arXiv:2210.02747}, 2022.

\bibitem[Loshchilov \& Hutter(2017)Loshchilov and Hutter]{loshchilov2017decoupled}
Ilya Loshchilov and Frank Hutter.
\newblock Decoupled weight decay regularization.
\newblock \emph{arXiv preprint arXiv:1711.05101}, 2017.

\bibitem[Lou \& Ermon(2023)Lou and Ermon]{lou2023reflected}
Aaron Lou and Stefano Ermon.
\newblock Reflected diffusion models.
\newblock In \emph{International Conference on Machine Learning}, pp.\  22675--22701. PMLR, 2023.

\bibitem[MacKay(2003)]{mackay2003information}
David~JC MacKay.
\newblock \emph{Information theory, inference and learning algorithms}.
\newblock Cambridge university press, 2003.

\bibitem[Nguyen \& Grover(2022)Nguyen and Grover]{DBLP:conf/icml/NguyenG22}
Tung Nguyen and Aditya Grover.
\newblock Transformer neural processes: Uncertainty-aware meta learning via sequence modeling.
\newblock In Kamalika Chaudhuri, Stefanie Jegelka, Le~Song, Csaba Szepesv{\'{a}}ri, Gang Niu, and Sivan Sabato (eds.), \emph{International Conference on Machine Learning, {ICML} 2022, 17-23 July 2022, Baltimore, Maryland, {USA}}, volume 162 of \emph{Proceedings of Machine Learning Research}, pp.\  16569--16594. {PMLR}, 2022.
\newblock URL \url{https://proceedings.mlr.press/v162/nguyen22b.html}.

\bibitem[Nielsen et~al.(2023)Nielsen, Christensen, Dittadi, and Winther]{nielsen2023diffenc}
Beatrix~MG Nielsen, Anders Christensen, Andrea Dittadi, and Ole Winther.
\newblock Diffenc: Variational diffusion with a learned encoder.
\newblock \emph{arXiv preprint arXiv:2310.19789}, 2023.

\bibitem[Niepert et~al.(2021)Niepert, Minervini, and Franceschi]{niepert2021implicit}
Mathias Niepert, Pasquale Minervini, and Luca Franceschi.
\newblock Implicit {MLE:} backpropagating through discrete exponential family distributions.
\newblock In Marc'Aurelio Ranzato, Alina Beygelzimer, Yann~N. Dauphin, Percy Liang, and Jennifer~Wortman Vaughan (eds.), \emph{Advances in Neural Information Processing Systems 34: Annual Conference on Neural Information Processing Systems 2021, NeurIPS 2021, December 6-14, 2021, virtual}, pp.\  14567--14579, 2021.
\newblock URL \url{https://proceedings.neurips.cc/paper/2021/hash/7a430339c10c642c4b2251756fd1b484-Abstract.html}.

\bibitem[Parmar et~al.(2018)Parmar, Vaswani, Uszkoreit, Kaiser, Shazeer, Ku, and Tran]{parmar2018image}
Niki Parmar, Ashish Vaswani, Jakob Uszkoreit, Lukasz Kaiser, Noam Shazeer, Alexander Ku, and Dustin Tran.
\newblock Image transformer.
\newblock In \emph{International conference on machine learning}, pp.\  4055--4064. PMLR, 2018.

\bibitem[Pearl et~al.(2023)Pearl, Brodsky, Berman, Zomet, Acha, Cohen-Or, and Lischinski]{pearl2023svnr}
Naama Pearl, Yaron Brodsky, Dana Berman, Assaf Zomet, Alex~Rav Acha, Daniel Cohen-Or, and Dani Lischinski.
\newblock Svnr: Spatially-variant noise removal with denoising diffusion.
\newblock \emph{arXiv preprint arXiv:2306.16052}, 2023.

\bibitem[Peluchetti(2023)]{peluchetti2023diffusion}
Stefano Peluchetti.
\newblock Diffusion bridge mixture transports, schr{\"o}dinger bridge problems and generative modeling.
\newblock \emph{Journal of Machine Learning Research}, 24\penalty0 (374):\penalty0 1--51, 2023.

\bibitem[Perez et~al.(2018)Perez, Strub, De~Vries, Dumoulin, and Courville]{perez2018film}
Ethan Perez, Florian Strub, Harm De~Vries, Vincent Dumoulin, and Aaron Courville.
\newblock Film: Visual reasoning with a general conditioning layer.
\newblock In \emph{Proceedings of the AAAI conference on artificial intelligence}, volume~32, 2018.

\bibitem[Popov et~al.(2021)Popov, Vovk, Gogoryan, Sadekova, and Kudinov]{popov2021grad}
Vadim Popov, Ivan Vovk, Vladimir Gogoryan, Tasnima Sadekova, and Mikhail Kudinov.
\newblock Grad-tts: A diffusion probabilistic model for text-to-speech.
\newblock In \emph{International Conference on Machine Learning}, pp.\  8599--8608. PMLR, 2021.

\bibitem[Preechakul et~al.(2022)Preechakul, Chatthee, Wizadwongsa, and Suwajanakorn]{preechakul2021diffusion}
Konpat Preechakul, Nattanat Chatthee, Suttisak Wizadwongsa, and Supasorn Suwajanakorn.
\newblock Diffusion autoencoders: Toward a meaningful and decodable representation.
\newblock In \emph{IEEE Conference on Computer Vision and Pattern Recognition (CVPR)}, 2022.

\bibitem[Rastogi et~al.(2023)Rastogi, Schiff, Hacohen, Li, Lee, Deng, Sabuncu, and Kuleshov]{rastogi2023semiparametric}
Richa Rastogi, Yair Schiff, Alon Hacohen, Zhaozhi Li, Ian Lee, Yuntian Deng, Mert~R. Sabuncu, and Volodymyr Kuleshov.
\newblock Semi-parametric inducing point networks and neural processes.
\newblock In \emph{The Eleventh International Conference on Learning Representations}, 2023.
\newblock URL \url{https://openreview.net/forum?id=FE99-fDrWd5}.

\bibitem[Rissanen et~al.(2022)Rissanen, Heinonen, and Solin]{rissanen2022generative}
Severi Rissanen, Markus Heinonen, and Arno Solin.
\newblock Generative modelling with inverse heat dissipation.
\newblock \emph{arXiv preprint arXiv:2206.13397}, 2022.

\bibitem[Rombach et~al.(2021)Rombach, Blattmann, Lorenz, Esser, and Ommer]{rombach2021highresolution}
Robin Rombach, Andreas Blattmann, Dominik Lorenz, Patrick Esser, and Björn Ommer.
\newblock High-resolution image synthesis with latent diffusion models, 2021.

\bibitem[Sahoo et~al.(2023)Sahoo, Paulus, Vlastelica, Musil, Kuleshov, and Martius]{sahoo2023backpropagation}
Subham~Sekhar Sahoo, Anselm Paulus, Marin Vlastelica, V{\'\i}t Musil, Volodymyr Kuleshov, and Georg Martius.
\newblock Backpropagation through combinatorial algorithms: Identity with projection works.
\newblock In \emph{The Eleventh International Conference on Learning Representations}, 2023.
\newblock URL \url{https://openreview.net/forum?id=JZMR727O29}.

\bibitem[Salimans \& Ho(2022)Salimans and Ho]{salimans2022progressive}
Tim Salimans and Jonathan Ho.
\newblock Progressive distillation for fast sampling of diffusion models.
\newblock \emph{arXiv preprint arXiv:2202.00512}, 2022.

\bibitem[Salimans et~al.(2017)Salimans, Karpathy, Chen, and Kingma]{salimans2017pixelcnn++}
Tim Salimans, Andrej Karpathy, Xi~Chen, and Diederik~P Kingma.
\newblock Pixelcnn++: Improving the pixelcnn with discretized logistic mixture likelihood and other modifications.
\newblock \emph{arXiv preprint arXiv:1701.05517}, 2017.

\bibitem[Shi et~al.(2024)Shi, De~Bortoli, Campbell, and Doucet]{shi2024diffusion}
Yuyang Shi, Valentin De~Bortoli, Andrew Campbell, and Arnaud Doucet.
\newblock Diffusion schr{\"o}dinger bridge matching.
\newblock \emph{Advances in Neural Information Processing Systems}, 36, 2024.

\bibitem[Si et~al.(2022)Si, Bishop, and Kuleshov]{si2022autoregressive}
Phillip Si, Allan Bishop, and Volodymyr Kuleshov.
\newblock Autoregressive quantile flows for predictive uncertainty estimation.
\newblock In \emph{International Conference on Learning Representations}, 2022.

\bibitem[Si et~al.(2023)Si, Chen, Sahoo, Schiff, and Kuleshov]{si2023semi}
Phillip Si, Zeyi Chen, Subham~Sekhar Sahoo, Yair Schiff, and Volodymyr Kuleshov.
\newblock Semi-autoregressive energy flows: exploring likelihood-free training of normalizing flows.
\newblock In \emph{International Conference on Machine Learning}, pp.\  31732--31753. PMLR, 2023.

\bibitem[Skilling(1989)]{Skilling1989TheEO}
John Skilling.
\newblock The eigenvalues of mega-dimensional matrices.
\newblock 1989.
\newblock URL \url{https://api.semanticscholar.org/CorpusID:117844915}.

\bibitem[Sohl-Dickstein et~al.(2015)Sohl-Dickstein, Weiss, Maheswaranathan, and Ganguli]{sohldickstein2015deep}
Jascha Sohl-Dickstein, Eric~A. Weiss, Niru Maheswaranathan, and Surya Ganguli.
\newblock Deep unsupervised learning using nonequilibrium thermodynamics, 2015.

\bibitem[Song et~al.(2017)Song, Kim, Nowozin, Ermon, and Kushman]{song2017pixeldefend}
Yang Song, Taesup Kim, Sebastian Nowozin, Stefano Ermon, and Nate Kushman.
\newblock Pixeldefend: Leveraging generative models to understand and defend against adversarial examples.
\newblock \emph{arXiv preprint arXiv:1710.10766}, 2017.

\bibitem[Song et~al.(2020)Song, Sohl-Dickstein, Kingma, Kumar, Ermon, and Poole]{song2020score}
Yang Song, Jascha Sohl-Dickstein, Diederik~P Kingma, Abhishek Kumar, Stefano Ermon, and Ben Poole.
\newblock Score-based generative modeling through stochastic differential equations.
\newblock \emph{arXiv preprint arXiv:2011.13456}, 2020.

\bibitem[Song et~al.(2021)Song, Durkan, Murray, and Ermon]{song2021maximum}
Yang Song, Conor Durkan, Iain Murray, and Stefano Ermon.
\newblock Maximum likelihood training of score-based diffusion models.
\newblock \emph{Advances in neural information processing systems}, 34:\penalty0 1415--1428, 2021.

\bibitem[Spinney \& Ford(2012)Spinney and Ford]{spinney2012fluctuation}
Richard~E. Spinney and Ian~J. Ford.
\newblock Fluctuation relations: a pedagogical overview, 2012.

\bibitem[Uria et~al.(2013)Uria, Murray, and Larochelle]{uria2013rnade}
Benigno Uria, Iain Murray, and Hugo Larochelle.
\newblock Rnade: The real-valued neural autoregressive density-estimator.
\newblock \emph{Advances in Neural Information Processing Systems}, 26, 2013.

\bibitem[Van~den Oord et~al.(2016)Van~den Oord, Kalchbrenner, Espeholt, Vinyals, Graves, et~al.]{van2016conditional}
Aaron Van~den Oord, Nal Kalchbrenner, Lasse Espeholt, Oriol Vinyals, Alex Graves, et~al.
\newblock Conditional image generation with pixelcnn decoders.
\newblock \emph{Advances in neural information processing systems}, 29, 2016.

\bibitem[Van Den~Oord et~al.(2016)Van Den~Oord, Kalchbrenner, and Kavukcuoglu]{van2016pixel}
A{\"a}ron Van Den~Oord, Nal Kalchbrenner, and Koray Kavukcuoglu.
\newblock Pixel recurrent neural networks.
\newblock In \emph{International conference on machine learning}, pp.\  1747--1756. PMLR, 2016.

\bibitem[Wang et~al.(2021)Wang, Jiao, Xu, Wang, and Yang]{wang2021deep}
Gefei Wang, Yuling Jiao, Qian Xu, Yang Wang, and Can Yang.
\newblock Deep generative learning via schr{\"o}dinger bridge.
\newblock In \emph{International conference on machine learning}, pp.\  10794--10804. PMLR, 2021.

\bibitem[Wang et~al.(2023)Wang, Schiff, Gokaslan, Pan, Wang, De~Sa, and Kuleshov]{wang2023infodiffusion}
Yingheng Wang, Yair Schiff, Aaron Gokaslan, Weishen Pan, Fei Wang, Christopher De~Sa, and Volodymyr Kuleshov.
\newblock Infodiffusion: Representation learning using information maximizing diffusion models.
\newblock In \emph{International Conference on Machine Learning}, pp.\  xxxx--xxxx. PMLR, 2023.

\bibitem[Xie \& Ermon(2019)Xie and Ermon]{xie2019reparameterizable}
Sang~Michael Xie and Stefano Ermon.
\newblock Reparameterizable subset sampling via continuous relaxations.
\newblock \emph{arXiv preprint arXiv:1901.10517}, 2019.

\bibitem[Yang et~al.(2024)Yang, Zhang, Yu, Liu, Xu, Ermon, and Cui]{yang2024cross}
Ling Yang, Zhilong Zhang, Zhaochen Yu, Jingwei Liu, Minkai Xu, Stefano Ermon, and Bin Cui.
\newblock Cross-modal contextualized diffusion models for text-guided visual generation and editing.
\newblock \emph{arXiv preprint arXiv:2402.16627}, 2024.

\bibitem[Yang \& Mandt(2023)Yang and Mandt]{yang2023lossy}
Ruihan Yang and Stephan Mandt.
\newblock Lossy image compression with conditional diffusion models, 2023.

\bibitem[Zhang \& Chen(2021)Zhang and Chen]{zhang2021diffusion}
Qinsheng Zhang and Yongxin Chen.
\newblock Diffusion normalizing flow.
\newblock \emph{Advances in Neural Information Processing Systems}, 34:\penalty0 16280--16291, 2021.

\bibitem[Zheng et~al.(2023)Zheng, Lu, Chen, and Zhu]{zheng2023improved}
Kaiwen Zheng, Cheng Lu, Jianfei Chen, and Jun Zhu.
\newblock Improved techniques for maximum likelihood estimation for diffusion odes.
\newblock \emph{arXiv preprint arXiv:2305.03935}, 2023.

\end{thebibliography}
\newpage

\setcounter{tocdepth}{2}
\tableofcontents
\allowdisplaybreaks
\begin{appendices}

\section{Comparing to Previous Work}
\label{appendix:previous}
\method{} is the first method to introduce a learned adaptive noise process. A widely held assumption is that the ELBO objective of a diffusion model is invariant to the noise process~\citep{kingma2021variational}. We dispel this assumption: we show that when input-conditioned noise is combined with (a) multivariate noise, (b) a novel polynomial parameterization, and (c) auxiliary variables, a learned noise process yields an improved variational posterior and a tighter ELBO. This approach sets a new state-of-the-art in density estimation. While (a), (c) were proposed in other contexts, we leverage them as subcomponents of a novel algorithm. We elaborate further on this below.

\subsection{Diffusion Models with Custom Noise}

The noise schedule, which determines the amount and type of noise added at each step, plays a critical role in diffusion models. 
\citet{chen2023importance} empirically demonstrate that different noise schedules can significantly impact the generated image quality using various handcrafted noise schedules. \citet{kingma2021variational} showed that the likelihood of a diffusion model remains invariant to the noise schedule with a scalar noise schedule. In this work we show that the ELBO is no longer invariant to multivariate noise schedules.

Recent works, including~\citet{hoogeboom2022blurring, rissanen2022generative, pearl2023svnr}, have explored per-pixel noise schedules (including blurring and other types of noising), yet none have delved into learning or conditioning the noise schedule on the input data itself. The shared components among these models are summarized and compared in \tab{tab:mulan_vs_others}. 

\subsection{Advanced Diffusion Models}

\citet{yang2024cross} proposes noise processes that are conditioned on an external context (e.g., text). We also propose context-conditioned noise processes; however, their setting is that of conditional generation, where the context is always available at training and inference time, and the context represents external data. Our paper instead looks at unconditional generation, and we condition the noising process on the image itself that we want to generate (via latent variable) and learn how to apply noise across an image as a function of the image.

\citet{lee2021priorgrad, popov2021grad} proposed using a data-dependent prior: however, they do not learn q and their noise process is not adaptive to the input $\x_0$. Thus they propose a fairly different set of methods from what we introduce.

\citet{yang2023lossy, wang2023infodiffusion} have explored diffusion models with an auxiliary latent space, where the denoising network is conditioned on a latent distribution.
Our paper also incorporate auxiliary latents, but unlike previous works, we add them to both $p$ and $q$ and we also also focus on learning the process $q$ (as opposed to doing representation learning using the auxiliary learned space). Lastly, our algorithm relies on many other components including a custom noise schedule, multivariate noise, etc. The shared components among these models are summarized and compared in \tab{tab:mulan_vs_others}.

\begin{table}[H]
    \centering
    \caption{\method{} is a noise schedule that can be integrated into any diffusion model such as VDM~\citep{kingma2021variational} or InfoDiffusion~\citep{wang2023infodiffusion}. The shared components between \method{} and these models are summarized and compared in this table.}\label{tab:mulan_vs_others}
    \begin{tabular}{lccccc}
    \toprule
        \multirow{3}{*}{Method}
        & \multirow{3}{*}{\parbox{1cm}{learned \\noise}}
        & \multirow{3}{*}{\parbox{1.5cm}{multivariate\\ noise}}
        & \multirow{3}{*}{\parbox{2.5cm}{input conditioned \\ noise}}
        & \multirow{3}{*}{\parbox{1.5cm}{auxiliary \\ latents}}
        & \multirow{3}{*}{\parbox{1.5cm}{noise \\schedule}} \\
        & & & & & \\
        & & & & & \\
        \midrule
        \multirow{3}{*}{VDM~\citep{kingma2021variational}}
            & \multirow{3}{*}{\textbf{Yes}}
            & \multirow{3}{*}{No}
            & \multirow{3}{*}{No}
            & \multirow{3}{*}{No}
            & \multirow{3}{*}{\parbox{1.5cm}{Monotonic \\neural \\network}} \\
        & & & & & \\
        & & & & & \\
        \hline
        \multirow{4}{*}{\parbox{2cm}{Blurring \\Diffusion \\Model~\citep{hoogeboom2022blurring}}}
            & \multirow{4}{*}{No}
            & \multirow{4}{*}{\textbf{Yes}}
            & \multirow{4}{*}{No}
            & \multirow{4}{*}{No}
            & \multirow{4}{*}{\parbox{1.5cm}{Frequency\\ scaling}} \\
            & & & & & \\
            & & & & & \\
            & & & & & \\
        \hline
        \multirow{3}{*}{\parbox{2cm}{Inverse Heat \\Dissipation~\citep{rissanen2022generative}}}
            & \multirow{3}{*}{No}
            & \multirow{3}{*}{\textbf{Yes}}
            & \multirow{3}{*}{No}
            & \multirow{3}{*}{No}
            & \multirow{3}{*}{Exponential} \\
            & & & & & \\
            & & & & & \\
        \hline
        \multirow{2}{*}{SVNR~\citep{pearl2023svnr}}
            & \multirow{2}{*}{No}
            & \multirow{2}{*}{\textbf{Yes}}
            & \multirow{2}{*}{No}
            & \multirow{2}{*}{No}
            & \multirow{2}{*}{\parbox{1.5cm}{Linear}} \\
            & & & & & \\
        \hline
        \multirow{3}{*}{\parbox{2cm}{InfoDiffusion~\citep{wang2023infodiffusion}}}
            & \multirow{3}{*}{No}
            & \multirow{3}{*}{No}
            & \multirow{3}{*}{No}
            & \multirow{3}{*}{\parbox{2cm}{In denoising\\process}}
            & \multirow{3}{*}{\parbox{1.5cm}{Cosine}} \\
            & & & & & \\
            & & & & & \\
        \hline
        \multirow{3}{*}{\parbox{2cm}{Lossy\\Compression~\citep{yang2023lossy}}}
            & \multirow{3}{*}{No}
            & \multirow{3}{*}{No}
            & \multirow{3}{*}{No}
            & \multirow{3}{*}{\parbox{2cm}{In denoising\\process}}
            & \multirow{3}{*}{\parbox{1.5cm}{Linear /\\ Cosine}} \\
            & & & & & \\
            & & & & & \\
        \hline
        \multirow{4}{*}{\method{} (\textbf{Ours})}
            & \multirow{4}{*}{\textbf{Yes}}
            & \multirow{4}{*}{\textbf{Yes}}
            & \multirow{4}{*}{\textbf{Yes}}
            & \multirow{4}{*}{\parbox{1.5cm}{
                In \textbf{noising} \\ and \\\textbf{denoising} \\ process}}
            & \multirow{4}{*}{Polynomial} \\
            & & & & & \\
            & & & & & \\
            & & & & & \\
    \bottomrule
    \end{tabular}
\end{table}

\subsection{Learned Diffusion}

Diffusion Normalizing Flow (DNF) uses the following forward process:
\begin{align}
    \text{d}\x_t = \mathbf{f}_\theta(\x_t, t)\text{d}t + g(t)\text{d}\mathbf{w},    
\end{align}
where the drift term $\mathbf{f}_\theta: \mathbb{R}^d \times \mathbb{R} \to \mathbb{R}^d$ is parameterized by a neural network with parameters $\theta$ and the diffusion term $g(t) \in \mathbb{R} ^+$ is a scalar constant and $\mathbf{w}$ is the standard Brownian motion. However, in MuLAN, the forward process is given by
\begin{equation}
    \text{d}\x_t = \mathbf{f}_\theta(\mathbf{z}, t)\odot\x_t\text{d}t + \mathbf{g}_\theta(\mathbf{z}, t)\odot \text{d} \mathbf{w} ; \z \sim q_\phi(\mathbf{z} | \x_0),
\end{equation}
where $\mathbf{z} \in \{0, 1\}^m$ is the auxiliary latent vector, $\mathbf{f}_\theta: \mathbb{R}^m \times \mathbb{R} \to \mathbb{R}^d$ and $\mathbf{g}_\theta: \mathbb{R}^m \times \mathbb{R} \to \mathbb{R}^d$ are parameterized by a neural network. Notice that the drift term in DNF, $\mathbf{f}_\theta(\x, t)$, is a non-linear function in $\x_0$, and the same holds for MuLAN since in the drift term, $\mathbf{f}_\theta(\mathbf{z}, t)\odot\x$, $\mathbf{z}$ and $\x$ depend on $\x_0$. Additionally, the diffusion coefficient, $\mathbf{g}_\theta(\mathbf{z}, t)$, is multivariate and conditioned on $\x_0$ via $\mathbf{z}$.
The two parameterizations are different: on one hand, DNF admits more general classes of neural networks because it does not require marginals to be tractable. On the other hand MuLAN admits a more flexible noise model $\mathbf{g}_\theta(\mathbf{z}, t)$ and admits more efficient training (see the summarized \tab{app:tab:mulan_vs_dnf} below).

MuLAN has the advantage that it is simulation free; which means that given a data $\x_0$, the noisy sample $\x_t$ can be computed in closed form; however, in Diffusion Normalizing Flow, to compute $\x_t$, one needs to simulate the forward SDE which is resource intensive and limits its scalability to larger denoising models. While MuLAN optimizes the ELBO, DNF optimizes an approximation for the ELBO. In particular, the DNF training objective does not involve a term that accounts for the entropy of the encoder. Thus, the objective is closer to that of a normal auto-encoder in that regard.

\begin{table}[H]
    \centering
    \renewcommand{\arraystretch}{1.5} %
    \caption{The key differences between \method{} and DNF is listed below.}
    \begin{tabular}{c|c|c|p{3cm}}
        \textbf{Aspect} & \textbf{Property} & \textbf{Diffusion Normalizing Flow} & \textbf{\method{}} \\ \hline
        \multirow{3}{*}{Drift Term}
            & Multivariate & Yes & Yes \\ \cline{2-4}
            & Adaptive & Yes & Yes \\ \cline{2-4}
            & Learnable & Yes & Yes \\ \hline
        \multirow{3}{*}{Diffusion Term}
            & Multivariate & No & \textbf{Yes} \\ \cline{2-4}
            & Adaptive & No & \textbf{Yes} \\ \cline{2-4}
            & Learnable & No & \textbf{Yes} \\ \hline
        Simulation Free Training & ~ & No & \textbf{Yes} \\ \hline
        Exact ELBO Optimization & ~ & No & \textbf{Yes} \\
        \hline
        NLL ($\downarrow$) & CIFAR-10 & 3.04 & \textbf{2.55}
    \end{tabular}
    \label{app:tab:mulan_vs_dnf}
\end{table}

Other concurrent work seeks to improve log-likelihood estimation by learning the forward diffusion process in a simulation-free setting. In neural diffusion models (NDMs), the noise schedule is fixed, but the mean of each marginal $q(\x_t|\x_0)$ is learned. This requires
denoising $\x$ in a transformed space, which prevents using noise parameterization, a design choice that is important for performance. Their denoising family also induces a parameterization that limits the kinds of samplers that can be sued.
Lastly, NDMs use a model that is 2x larger than a regular diffusion model, while ours only adds 10\% more parameters.

The DiffEnc framework adds an extra learned correction term to $q$ to adjust the mean of each marginal $q(\x_t|\x_0)$. This noise choice also requires using certain parameterizations for $\x$ that are not compatible with noise parameterization; while their approach supports v-parameterization, it also requires training a mean parameterization network. Similarly to NDMs, the noise schedule remains fixed, while the mean of each marginal is adjusted by the network. Our approach towards learning the noise schedule yields better empirical performance and is, in our opinion, simpler; it can also be combined with this prior work on learning the marginals' means.

\subsection{Optimal Transport}

In techniques based on optimal transport, the goal is to learn a noise process that minimizes the transport cost from data to noise, which in practice produces smoother diffusion trajectories that facilitate sampling.

Minimizing Trajectory Curvature (MTC) of ODE-based generative models \citet{lee2023minimizing}, the primary goal is to design the forward diffusion process that is optimal for fast sampling; however, MuLAN strives to learn a forward process that optimizes for log-likelihood. In the former, the marginals $\x_t$ in the forward process are given as
\begin{align} \x_t = (1 - t)\x_0 + t\mathbf{z}; \mathbf{z} \sim q_\phi(\mathbf{z} | \x_0) \end{align} where $\x_t, \x_0, \mathbb{z} \in \mathbb{R}^d$. However for MuLAN the marginals are $\x_t = \balpha_\phi(\mathbf{z}, t)\odot\x_0 + \sqrt{1 - \balpha_\phi^2(\mathbf{z}, t)} \odot \epsilon$ ; $\epsilon \sim \mathcal{N}(0, \mathbf{I}_d)$ ; $\mathbf{z} \sim q_\phi(\mathbf{z} | \x_0)$ where $\balpha_\phi(\mathbf{z}, t): \mathbb{R}^d \times \mathbb{R} \to \mathbb{R}^d_{\geq 0}$ , $\mathbf{z} \in \{0, 1\}^m$ , $\epsilon \in \mathbb{R}^d$
Notice that in the MTC formula, the coefficient of $\x_0$ , the time integral of the drift term, is a scalar and linear function of, and is independent of the input $\x_0$. In MuLAN, that term is a multivariate non-linear function in $t$, and conditioned on $\x_0$ via the auxiliary latent variable $\mathbf{z}$. This implies that the forward diffusion process in MuLAN is more expressive than MTC. The simplistic forward process in MTC enables faster sampling whereas the richer / more expressive forward process in MuLAN leads to improved likelihood estimates. \tab{app:tab:mulan_vs_mtc} summarizes the key differences.

\begin{table}[H]
    \centering
    \caption{Comparison between Minimizing Trajectory Curvature and MuLAN methods.}
    \renewcommand{\arraystretch}{1.5} %
    \begin{tabular}{c|c|c|p{3cm}}
        \textbf{Aspect} & \textbf{Property} & \textbf{Minimizing Trajectory Curvature} & \textbf{MuLAN} \\ \hline
        Goal & ~ & Design faster sampler & Improve log-likelihood \\ \hline
        \multirow{4}{*}{Drift Term} & Learnable & No & \textbf{Yes} \\ \cline{2-4}
        & Linearity & Linear in time $t$, linear in $\x_0$ & \textbf{Non-linear} in time $t$, \textbf{Non-linear} in $\z$ (and hence $\x_0$) \\ \cline{2-4}
        & Dimensionality & Scalar & \textbf{Multivariate} \\ \cline{2-4}
        & Adaptive & No & \textbf{Yes} \\ \hline
        \multirow{4}{*}{Diffusion Term} & Linearity & Linear in time $t$ & \textbf{Non-linear} in time $t$ \\ \cline{2-4}
        & Dimensionality & Multivariate & Multivariate \\ \cline{2-4}
        & Learnable & Yes & Yes \\ \cline{2-4}
        & Adaptive & Yes & Yes \\ %
    \end{tabular}
    \label{app:tab:mulan_vs_mtc}
    
\end{table}

An alternative approach to learning a forward process that performs optimal transport is via the theory of Schrodinger bridges~\citep{shi2024diffusion, de2021diffusion,wang2021deep,peluchetti2023diffusion} .
Similarly to the DNF framework, these methods do not admit analytical marginals and therefore involve computing full trajectories from noisy and clean data. Additionally, they are typically trained using an iterative procedure that generalizes the sinkhorn algorithm and involves iteratively training $q$ and $p$. As such, these types of methods are typically more expensive to train and competitive results on standard benchmarks (e.g., CIFAR10, ImageNet) are not yet available to our knowledge.

\section{Standard Diffusion Models}
We have a Gaussian diffusion process that begins with the data $\x_0$, and defines a sequence of increasingly noisy versions of $\x_0$ which we call the latent variables $\x_t$, where $t$ runs from $t=0$ (least noisy) to $t=1$ (most noisy). 
Given, $T$, we discretize time uniformly into $T$ timesteps each with a width $1 / T$. We define $\t = i / T$ and $\s = (i - 1) / T$. 

\subsection{Forward Process}
\begin{equation}
    q(\x_t | \x_s) = \mathcal{N}(\alpha_{t | s} \x_s,  \sigma^2_{t|s} \bfI_n)
\end{equation}
where
\begin{align}
    \alpha_{t|s} &= \frac{\alpha_t}{\alpha_s}\\
    \sigma^2_{t | s} &= \sigma_t^2 - \frac{\alpha_{t|s}^2}{\sigma_s^2} %
\end{align}

\subsection{Reverse Process}
\citet{kingma2021variational} show that the distribution $q(\x_s | \x_t, \x_0)$ is also gaussian,

\begin{align}\label{appendix:equation:q_vdm}
    & q(\x_s | \x_t, \x_0) = \mathcal{N} \left ( \bmu_q = \frac{\alpha_{t | s}\sigma^2_{s}}{\sigma^2_{t}} \x_t + \frac{\sigma^2_{t | s}\alpha_s}{\sigma_t^2} \x_0, \; \bSigma_q = \frac{\sigma^2_s \sigma^2_{t | s}}{\sigma_t^2} \bfI_n\right) 
\end{align}

Since during the reverse process, we don't have access to $\x_0$, we approximate it using a neural network $\x_\theta(\x_t, t)$ with parameters $\theta$. Thus,
\begin{equation}\label{appendix:equation:p_vdm}
    \p(\x_s | \x_t) = \mathcal{N} \left ( \bmu_p = \frac{\alpha_{t | s}\sigma^2_{s}}{\sigma^2_{t}} \x_t + \frac{\sigma^2_{t | s}\alpha_s}{\sigma_t^2} \x_\theta(\x_t, t), \; \bSigma_p = \frac{\sigma^2_s \sigma^2_{t | s}}{\sigma_t^2} \bfI_n\right) 
\end{equation}

\subsection{Variational Lower Bound}\label{appendix:standard_vlb}This corruption process $q$ is the following markov-chain as $q(\x_{0:1}) = q (\x_0) \left(\prod_{i=1}^{T} q(\x_\t | \x_\s) \right)$. In the Reverse Process, or the Denoising Process, $\p$, a neural network (with parameters $\theta$) is used to denoise the noising process $q$. The Reverse Process factorizes as: $p_\theta(\x_{0:1}) = p_\theta(\x_1) \prod_{i=1}^{T} p_\theta(\x_{\s} | \x_\t)$. Let $\x_\theta(\x_t, t)$ be the reconstructed input by a neural network from $\x_t$. Similar to \citet{sohldickstein2015deep, kingma2021variational} we decompose the negative lower bound (VLB) as:
\begin{align}
    -\log \p (\x_0)
        & \leq  \underbrace{\mathbb{E}_\q \left [ - \log \frac{\p (\x_{t(0):t(T)})}{q(\x_{t(1):t(T)} | \x_0)} \right]}_{\text{ELBO}(p_\theta(\x_{0:1}), q(\x_{0:1})) \text{defined in \Eqn{eqn:diffusion_nelbo}}} \nonumber \\
        & = \mathbb{E}_{\x_{t(1)} \sim q(\x_{t(1)} | \x_0)} [- \log \p (\x_{0} | \x_{t(1)})] \nonumber \\
        & \hspace{4mm} + 
            \sum_{i=2}^{T} \mathbb{E}_{\x_{t(i)} \sim q(\x_{t(i)} | \x_0)} \kl [ \p (\x_{\s} | \x_{\t}) \|  q(\x_{\s} | \x_\t, \x_0)] \nonumber \\
        & \hspace{4mm} + \kl [\p(\x_1) \| \q(\x_1| \x_0)] \nonumber \\
        & = \underbrace{
            \mathbb{E}_{\x_{t(1)} \sim q(\x_{t(1)} | \x_0)} [- \log \p (\x_{0} | \x_{t(1)})]}_{\lossrecons} \nonumber \\
        & \hspace{4mm} + \underbrace{
            \frac{T}{2} \mathbb{E}_{
                \epsilon \sim \mathcal{N}(0, \bfI_n), i \sim U\{2, T\}}
                    \kl [ \p (\x_{\s} | \x_{\t}) \|  q(\x_{\s} | \x_\t,\x_0)]}_{\lossdiff} \nonumber \\
        & \hspace{4mm} + \underbrace{\kl [\p(\x_1) \| q(\x_1| \x_0)]}_{\lossprior}
\end{align}
The prior loss, $\lossprior$, and reconstruction loss, $\lossrecons$, can be (stochastically and differentiably) estimated using standard techniques; see \citet{kingma2013auto}. The diffusion loss, $\lossdiff$, varies with the formulation of the noise schedule. We provide an exact formulation for it in the subsequent sections.

\subsection{Diffusion Loss}
For brevity, we use the notation $s$ for $s(i)$ and $t$ for $t(i)$. From \Eqn{eqn:backward_exact_multivariate} and \Eqn{eqn:backward_approx_multivariate} we get the following expression for $q(\x_s | \x_t, \x_0)$:
\begin{align}
    & \kl (q(\x_s | \x_t, \x_0) \| p_\theta(\x_s | \x_t)) \nonumber \\
    & = \frac{1}{2}\left(
        (\bmu_q - \bmu_p)^\top 
        \bSigma^{- 1}_\theta
        (\bmu_q - \bmu_p)
        + \text{tr} \left(\bSigma_q \bSigma_p^{-1} - \bfI_n\right)
        - \log \frac{|\bSigma_q|}{|\bSigma_p|}\right) \nonumber \\
    & = \frac{1}{2} (\bmu_q - \bmu_p)^\top  \Sigma^{- 1}_\theta (\bmu_q - \bmu_p) \nonumber \\
    & \text{\footnotesize Substituting $\bmu_q, \bSigma_q, \bmu_p, \bSigma_p$ from \eqref{appendix:equation:p_vdm} and \eqref{appendix:equation:q_vdm}; for the exact derivation see \citet{kingma2021variational}} \nonumber  \\
    & = \frac{1}{2} \left(\nu(s) - \nu(t) \right) \|(\x_0 - \x_\theta(\x_t, t))\|_2^2
\end{align}

Thus $\lossdiff$ is given by

\begin{align}
    & \lossdiff \nonumber \\
    &= \lim_{T \to \infty}\frac{T}{2} \mathbb{E}_{
                \epsilon \sim \mathcal{N}(0, \bfI_n), i \sim U\{2, T\}}
                    \kl [ \p (\x_{\s} | \x_{\t}) \|  \q(\x_{\s} | \x_\t,\x_0)] \nonumber \\
    & = \lim_{T \to \infty} \frac{1}{2} \sum_{i=2}^T \mathbb{E}_{
        \epsilon \sim \mathcal{N}(0, \bfI_n)}
           \left(\nu(s) - \nu(t) \right) \|\x_0 - \x_\theta(\x_t, t)\|_2^2 \nonumber \\
    & = \frac{1}{2} \mathbb{E}_{\epsilon \sim \mathcal{N}(0, \bfI_n)} \left [
        \lim_{T \to \infty} \sum_{i=2}^T 
            \left(\nu(s) - \nu(t) \right) \|\x_0 - \x_\theta(\x_t, t)\|_2^2\right] \nonumber \\
    & = \frac{1}{2} \mathbb{E}_{\epsilon \sim \mathcal{N}(0, \bfI_n)} \left [
        \lim_{T \to \infty} \sum_{i=2}^T 
            T\left(\nu(s) - \nu(t) \right) \|\x_0 - \x_\theta(\x_t, t)\|_2^2 \frac{1}{T}\right] \nonumber \\
    & \text{Substituting $\lim_{T \to \infty}T (\nu(s) - \nu(t)) = \frac{\text{d}}{\text{d}t}\nu(t) \equiv \nu'(t)$; see \citet{kingma2021variational}} \nonumber \\
    & = \frac{1}{2} \mathbb{E}_{\epsilon \sim \mathcal{N}(0, \bfI_n)} \left [
        \int_{0}^1 
            \nu'(t) \|\x_0 - \x_\theta(\x_t, t)\|_2^2\right] \text{d}t \\
    & \text{\footnotesize In practice instead of computing the integral is computed by MC sampling.} \nonumber \\
    & = - \frac{1}{2} \mathbb{E}_{
        \epsilon \sim \mathcal{N}(0, \bfI_n), t \sim U[0, 1]}
           \left[ \nu'(t) \|\x_0 - \x_\theta(\x_t, t)\|_2^2 \right]
\end{align}

\section{Multivariate noise schedule}\label{appendix:multivariate_ns}
For a multivariate noise schedule we have  $\balpha_t, \bsigma_t \in \mathbb{R}_{+}^{d}$ where $t \in [0, 1]$. $\balpha_t, \bsigma_t$ are vectors. The timesteps $s, t$ satisfy $0 \leq s < t \leq 1$. Furthermore, we use the following notations where arithmetic division represents element wise division between 2 vectors:

\begin{align}
    \balpha_{t|s} &= \frac{\balpha_t}{\balpha_s} \label{eqn:notations_multivariate_alpha}\\
    \bsigma^2_{t | s} &= \bsigma_t^2 - \frac{\balpha_{t|s}^2}{\bsigma_s^2} \label{eqn:notations_multivariate_sigma}
\end{align}

\subsection{Forward Process}\label{appendix:multivariate_ns_forward}

\begin{equation}
    q(\x_t | \x_s) = \mathcal{N} \left ( \balpha_{t|s} \x_s, \bsigma^2_{t|s} \right )
\end{equation}

\paragraph{Change of variables.}
We can write $\x_t$ explicitly in terms of the signal-to-noise ratio, $\snr(t)$, and input $\x_0$ in the following manner:

\begin{align}\label{eqn:change_of_variables}
    & \snr_t = \frac{\balpha^2_t}{\bsigma^2_t} \nonumber \\
    & \text{\footnotesize We know $\alpha^2_t = 1 - \sigma^2_t$ for Variance Preserving process; see \sec{sec:background}.} \nonumber \\
    & \implies  \frac{1 - \bsigma^2_t}{\bsigma^2_t} = \snr_t \nonumber \\
    & \implies  \bsigma^2_t = \frac{1}{1 + \snr_t} \;\;\; \text{and}\;\;\;  \balpha^2_t = \frac{\snr_t}{1 + \snr_t}
\end{align}
\begin{align}
    & \nu_t = \frac{\alpha^2_t}{\sigma^2_t} \nonumber \\
    & \text{\footnotesize We know $\alpha^2_t = 1 - \sigma^2_t$ for Variance Preserving process; see \sec{sec:background}.} \nonumber \\
    & \implies  \frac{1 - \sigma^2_t}{\sigma^2_t} = \nu_t \nonumber \\
    & \implies  \sigma^2_t = \frac{1}{1 + \nu_t} \;\;\; \text{and}\;\;\;  \alpha^2_t = \frac{\nu_t}{1 + \nu_t}
\end{align}
Thus, we write $\x_t$ in terms of the signal-to-noise ratio in the following manner:
\begin{align}\label{eqn:x_snr}
    \x_{\snr(t)} & = \balpha_{t} \x_0 +  \bsigma_{t} \noise_t; \; \noise_t \sim \mathcal{N}(0, \bfI_n) & \nonumber\\
    & = \frac{\sqrt{\snr(t)}}{\sqrt{1 + \snr(t)}}\x_0 + \frac{1}{\sqrt{1 + \snr(t)}}\noise_t & \text{\footnotesize Using \Eqn{eqn:change_of_variables}}
\end{align}

\subsection{Reverse Process}
The distribution of $\x_t$ given $\x_s$ is given by:
\begin{align}\label{eqn:backward_exact_multivariate}
    & q(\x_s | \x_t, \x_0) = \mathcal{N} \left ( \bmu_q = \frac{\balpha_{t | s}\bsigma^2_{s}}{\bsigma^2_{t}} \x_t + \frac{\bsigma^2_{t | s}\balpha_s}{\bsigma_t^2} \x_0, \; \bSigma_q = \diag \left(\frac{\bsigma^2_s \bsigma^2_{t | s}}{\bsigma_t^2} \right) \right) 
\end{align}

Let $\x_\theta(\x_t, t)$ be the neural network approximation for $\x_0$. Then we get the following reverse process:
\begin{align}\label{eqn:backward_approx_multivariate}
    & p_\theta(\x_s | \x_t) = \mathcal{N} \left ( \bmu_p = \frac{\balpha_{t | s}\bsigma^2_{s}}{\bsigma^2_{t}} \x_t + \frac{\bsigma^2_{t | s}\balpha_s}{\bsigma_t^2} \x_\theta(\x_t, t),
    \; \bSigma_p = \diag \left( \frac{\bsigma^2_s \bsigma^2_{t | s}}{\bsigma_t^2} \right) \right)
\end{align}

\subsection{Diffusion Loss}
For brevity we use the notation $s$ for $s(i)$ and $t$ for $t(i)$. From \Eqn{eqn:backward_exact_multivariate} and \Eqn{eqn:backward_approx_multivariate} we get the following expression for $q(\x_s | \x_t, \x_0)$:
\begin{align}\label{eqn:diffusion_loss_multivariate}
    & \kl (q(\x_s | \x_t, \x_0) \| p_\theta(\x_s | \x_t)) \nonumber \\
    & = \frac{1}{2}\left(
        (\bmu_q - \bmu_p)^\top 
        \bSigma^{- 1}_\theta
        (\bmu_q - \bmu_p)
        + \text{tr} \left(\bSigma_q \bSigma_p^{-1} - \bfI_n\right)
        - \log \frac{|\bSigma_q|}{|\bSigma_p|}\right) \nonumber \\
    & = \frac{1}{2} (\bmu_q - \bmu_p)^\top  \Sigma^{- 1}_\theta (\bmu_q - \bmu_p) \nonumber \\
    & \text{\footnotesize Substituting $\bmu_q, \bmu_p, \bSigma_p$ from \eqref{eqn:backward_approx_multivariate} and \eqref{eqn:backward_exact_multivariate}.} \nonumber \\
    & = \frac{1}{2} \left (\frac{\bsigma^2_{t | s}\balpha_s}{\bsigma_t^2} \x_0
                     - \frac{\bsigma^2_{t | s}\balpha_s}{\bsigma_t^2} \x_\theta(\x_t, t) \right)^\top 
                     \diag \left( \frac{\bsigma^2_s \bsigma^2_{t | s}}{\bsigma_t^2} \right) ^ {-1}
                     \left (\frac{\bsigma^2_{t | s}\balpha_s}{\bsigma_t^2} \x_0
                     - \frac{\bsigma^2_{t | s}\balpha_s}{\bsigma_t^2} \x_\theta(\x_t, t) \right) \nonumber \\
    & = \frac{1}{2} (\x_0 - \x_\theta(\x_t, t))^\top  \diag \left(
        \frac{\bsigma^2_{t | s}\balpha_s}{\bsigma_t^2} \right)^\top 
        \diag  \left( \frac{\bsigma^2_s \bsigma^2_{t | s}}{\bsigma_t^2} \right) ^ {-1}
         \diag \left(\frac{\bsigma^2_{t | s}\balpha_s}{\bsigma_t^2} \right)  (\x_0 - \x_\theta(\x_t, t)) \nonumber \\
    & = \frac{1}{2} (\x_0 - \x_\theta(\x_t, t))^\top  \diag \left(
        \frac{\bsigma^2_{t | s}\balpha_s}{\bsigma_t^2} \odot \frac{\bsigma_t^2}{\bsigma^2_s \bsigma^2_{t | s}} \odot \frac{\bsigma^2_{t | s}\balpha_s}{\bsigma_t^2} \right)  (\x_0 - \x_\theta(\x_t, t))\nonumber \\
    & = \frac{1}{2} (\x_0 - \x_\theta(\x_t, t))^\top 
        \diag \left(\frac{\bsigma^2_{t | s}\balpha^2_s}{\bsigma_t^2\bsigma_s^2} \right)
        (\x_0 - \x_\theta(\x_t, t)) \nonumber \\
    & \text{\footnotesize Simplifying the expression using \Eqn{eqn:notations_multivariate_alpha} and \Eqn{eqn:notations_multivariate_sigma} we get,} \nonumber \\
    & = \frac{1}{2} (\x_0 - \x_\theta(\x_t, t))^\top 
        \diag \left(\frac{\balpha^2_s}{\bsigma_s^2} - \frac{\balpha^2_t}{\bsigma_t^2} \right)
        (\x_0 - \x_\theta(\x_t, t)) \nonumber \\
    & \text{\footnotesize Using the relation $\snr(t) = \balpha^2_t / \bsigma^2_t$ we get,} \nonumber \\
    & = \frac{1}{2} (\x_0 - \x_\theta(\x_t, t))^\top 
        \diag \left(\snr(s) - \snr(t) \right)
        (\x_0 - \x_\theta(\x_t, t))
\end{align}

Like \citet{kingma2021variational} we train the model in the continuous domain with $T \to \infty$.
\begin{align}
    & \lossdiff \nonumber \\
    & = \lim_{T \to \infty} \frac{1}{2} \sum_{i=2}^T \mathbb{E}_{
        \epsilon \sim \mathcal{N}(0, \bfI_n)}
        \kl (q(\x_\s | \x_\t, \x_0) \| p_\theta(\x_\s | \x_\t)) \nonumber \\
    & = \lim_{T \to \infty} \frac{1}{2} \sum_{i=2}^T \mathbb{E}_{
        \epsilon \sim \mathcal{N}(0, \bfI_n)}
            (\x_0 - \x_\theta(\x_\t, \t))^\top 
            \diag \left(\snr_\s - \snr_\t \right)
            (\x_0 - \x_\theta(\x_\t, t)) \nonumber \\
    & = \frac{1}{2} \mathbb{E}_{\epsilon \sim \mathcal{N}(0, \bfI_n)} \left [
        \lim_{T \to \infty} \sum_{i=2}^T 
            (\x_0 - \x_\theta(\x_\t, \t))^\top 
            \diag \left(\snr_\s - \snr_\t \right)
            (\x_0 - \x_\theta(\x_\t, t))\right] \nonumber \\
    & = \frac{1}{2} \mathbb{E}_{\epsilon \sim \mathcal{N}(0, \bfI_n)} \left [
        \lim_{T \to \infty} \sum_{i=2}^T 
            T (\x_0 - \x_\theta(\x_\t, \t))^\top 
            \diag \left(\snr_\s - \snr_\t \right)
            (\x_0 - \x_\theta(\x_\t, t)) \frac{1}{T}\right] \nonumber \\
    & \text{Let \footnotesize $\lim_{T \to \infty} T (\snr_\s - \snr_\t) = \frac{\text{d}}{\text{d}t}\snr(t)$ denote the scalar derivative of the vector $\snr(t)$ w.r.t $t$} \nonumber \\
    & = \frac{1}{2} \mathbb{E}_{\epsilon \sim \mathcal{N}(0, \bfI_n)} \left [
        \int_{0}^1 
            (\x_0 - \x_\theta(\x_t, t))^\top 
            \diag \left(\frac{\text{d}}{\text{d}t}\snr(t) \right)
            (\x_0 - \x_\theta(\x_t, t)) \text{d}t \right] \label{eqn:multivariate_diffusion_loss_integral} \\
    & \text{\footnotesize In practice instead of computing the integral is computed by MC sampling.} \nonumber \\
    & = - \frac{1}{2} \mathbb{E}_{
        \epsilon \sim \mathcal{N}(0, \bfI_n), t \sim U[0, 1]} \left[
            (\x_0 - \x_\theta(\x_t, t))^\top 
            \diag \left(\frac{\text{d}}{\text{d}t}\snr(t) \right)
            (\x_0 - \x_\theta(\x_t, t)) \right]
\end{align}

\subsection{Vectorized Representation of the diffusion loss}
Let  $\vsnr(t)$ be the vectorized representation of the diagonal entries of the matrix $\snr(t)$.
We can rewrite the integral in $\ref{eqn:multivariate_diffusion_loss_integral}$ in the following vectorized form where $\odot$ denotes element wise multiplication and $\langle, \rangle$ denotes dot product between 2 vectors.
\begin{align}
    & \lossdiff \nonumber \\
    & = - \frac{1}{2} \int_{0}^1 
        (\x_0 - \x_\theta(\x_t, t))^\top 
        \diag \left(\frac{\text{d}}{\text{d}t}\snr(t) \right)
        (\x_0 - \x_\theta(\x_t, t)) \text{d}t \nonumber \\
     & = - \frac{1}{2} \int_{0}^{1}
        \left \langle (\x_0 - \x_\theta(\x_t, t)) \odot (\x_0 - \x_\theta(\x_t, t)), \frac{\text{d}}{\text{d}t} \vsnr(t) \right \rangle \text{d}t \nonumber \\
    & \text{\footnotesize Using change of variables as mentioned in \sec{subsec:multivariate} we have}\nonumber \\
    & = - \frac{1}{2} \int_{0}^{1}
        \left \langle (\x_0 - \Tilde{\x}_\theta(\x_{\snr(t)}, \snr(t))) \odot (\x_0 - \Tilde{\x}_\theta(\x_{\snr(t)}, \snr(t))), \frac{\text{d}}{\text{d}t} \vsnr(t) \right \rangle \text{d}t \nonumber \\
    & \text{\footnotesize Let $ \forcefieldnoz= (\x_0 - \Tilde{\x}_\theta(\x_{\snr(t)}, \snr(t))) \odot (\x_0 - \Tilde{\x}_\theta(\x_{\snr(t)}, \snr(t)))$ } \nonumber \\
    & = \int_{0}^1 \left \langle  \forcefieldnoz, \frac{\text{d}}{\text{d}t} \vsnr(t) \right \rangle \text{d}t
\end{align}

Thus $\lossdiff$ can be interpreted as the amount of work done along the trajectory $\vsnr(0) \xrightarrow{} \vsnr(1)$ in the presence of a vector field $ \mathbf{f}_\theta(\x_0, \snr(\z, t))$. From the perspective of thermodynamics, this is precisely equal to the amount of heat lost into the environment during the process of transition between 2 equilibria via the noise schedule specified by $\snr(t)$.

\subsection{Log likelihood and Noise Schedules: A Thermodynamics perspective}
A diffusion model characterizes a quasi-static process that occurs between two equilibrium distributions: $q(\x_0) \xrightarrow[]{} q(\x_1)$, via a stochastic trajectory \citep{sohldickstein2015deep}. According to \citet{spinney2012fluctuation}, it is demonstrated that the diffusion schedule or the noising process plays a pivotal role in determining the "measure of irreversibility" for this stochastic trajectory which is expressed as $\log \frac{P_F(\x_{0:1})}{P_B(\x_{1:0})}$. $P_F(\x_{0:1})$ represents the probability of observing the forward path $\x_{0:1}$ and $P_B(\x_{1:0})$ represents the probability of observing the reverse path $\x_{1:0}$. It's worth noting that $\log \frac{P_F(\x_{0:1})}{P_B(\x_{1:0})}$ corresponds precisely to the ELBO \Eqn{eqn:diffusion_nelbo} that we optimize when training a diffusion model. Consequently, thermodynamics asserts that the noise schedule indeed has an impact on the log-likelihood of the diffusion model which contradicts \citet{kingma2021variational}.

\section{Multivariate noise schedule conditioned on context}\label{appendix:multivariate_noising_schedule_context}
Let's say we have a context variable $\c \in \mathbb{R}^m$ that captures high level information about $\x_0$. $\balpha_t(\c), \bsigma_t(\c) \in \mathbb{R}_{+}^{d}$ 
are vectors. The timesteps $s, t$ satisfy $0 \leq s < t \leq 1$. Furthermore, we use the following notations:
\begin{align}\label{eqn:notations}
    \balpha_{t|s}(\c) &= \frac{\balpha_t(\c)}{ \balpha_s(\c)} \\
    \bsigma^2_{t | s}(\c) &= \bsigma_t^2(\c) - \frac{\balpha_{t|s}^2(\c)}{\bsigma_s^{2}(\c)}
\end{align}

The forward process for such a method is given as:
\begin{equation}
    \q(\x_t | \x_s, \c) = \mathcal{N} \left ( \balpha_{t|s}(\c) \x_s, \bsigma^2_{t|s}(\c) \right )
\end{equation}
The distribution of $\x_t$ given $\x_s$ is given by (the derivation is similar to \citet{hoogeboom2022blurring}):
\begin{align}\label{eqn:p_theta_available_context}
    & \q(\x_s | \x_t, \x_0, \c) \nonumber \\
    & = \mathcal{N} \left ( \bmu_q = \frac{\balpha_{t | s}(\c)\bsigma^2_{s}(\c)}{\bsigma^2_{t}(\c)} \x_t + \frac{\bsigma^2_{t | s}(\c)\balpha_s(\c)}{\bsigma_t^2(\c)} \x_0, \; \bSigma_q = \diag \left (\frac{\bsigma^2_s(\c) \bsigma^2_{t | s}(\c)}{\bsigma_t^2(\c)}\right) \right) 
\end{align}

\subsection{context \textbf{is} available during the inference time.}

Even though $\c$ represents the input $\x_0$, it could be available during during inference. For example $\c$ could be class labels \citep{dhariwal2021diffusion} or prexisting embeddings from an auto-encoder \citep{preechakul2021diffusion}.

\subsubsection{Reverse Process: Approximate}
Let $\x_\theta(\x_t, \c, t)$ be an approximation for $\x_0$. Then we get the following reverse process (for brevity we write $\x_\theta(\x_t, \c, t)$ as $\x_\theta$):
\begin{align}
    \p(\x_s | \x_t, \c) = \mathcal{N} \left ( \bmu_p = \frac{\balpha_{t | s}(\c)\bsigma^2_{s}(\c)}{\bsigma^2_{t}(\c)} \x_t + \frac{\bsigma^2_{t | s}(\c)\balpha_s(\c)}{\bsigma_t^2(\c)} \x_\theta,
    \; \bSigma_p = \diag \left(\frac{\bsigma^2_s(\c) \bsigma^2_{t | s}(\c)}{\bsigma_t^2(\c)}\right) \right)
\end{align}

\subsubsection{Diffusion Loss}

Similar to the derivation of multi-variate $\lossdiff$ in \Eqn{eqn:diffusion_loss_multivariate} we can derive $\lossdiff$ for this case too:
\begin{align}
     \lossdiff
    = - \frac{1}{2} \mathbb{E}_{
        \epsilon \sim \mathcal{N}(0, \bfI_n), t \sim U[0, 1]} \left[
            (\x_0 - \x_\theta(\x_t, \c, t))^\top 
            \diag \left(\frac{\text{d}}{\text{d}t}\snr(t) \right)
            (\x_0 - \x_\theta(\x_t, \c, t)) \right]
\end{align}

\subsubsection{Limitations of this method}

This approach is very limited where the diffusion process is only conditioned on class labels. Using pre-existing embeddings like Diff-AE~\citep{preechakul2021diffusion} is also not possible in general and is only limited to tasks such as attribute manipulation in datasets.

\subsection{context \textbf{isn't} available during the inference time.}
If the context, $\c$ is an explicit function of the input $\x_0$ things become challenging because $\x_0$ isn't available during the inference stage. For this reason, \Eqn{eqn:p_theta_available_context} can't be used to parameterize $\bmu_p, \bSigma_p$ in $p_\theta(\x_s | \x_t)$. Let $p_\theta(\x_s | \x_t) = \mathcal{N} (\bmu_p(\x_t, t), \bSigma_p(\x_t, t))$ where $\bmu_p, \bSigma_p$ are parameterized directly by a neural network. Using \Eqn{eqn:q_context_conditioned} we get the following diffusion loss:
\begin{align}\label{eqn:diffusion_loss_unavailable_context}
    \lossdiff & = T \; \mathbb{E}_{i \sim U[0, T]} \kl \left (q(\x_\s | \x_\t, \x_0) \| p_\theta(\x_\s | \x_\t) \right) \nonumber \\
    & = \mathbb{E}_{\q}
            \left (\underbrace{\frac{T}{2} 
                    (\bmu_q - \bmu_p)^\top 
                    \bSigma^{- 1}_\theta
                    (\bmu_q - \bmu_p)}_{\text{term 1}}
        + \underbrace{\frac{T}{2} \left(
            \text{tr} \left(\bSigma_q \bSigma_p^{-1} - \bfI_n\right)
            - \log \frac{|\bSigma_q|}{|\bSigma_p|} \right)}_{\text{term 2}}
            \right)
\end{align}

\subsubsection{Reverse Process: Approximate}
Due to the challenges associated with parameterizing $\bmu_p, \bSigma_p$ directly using a neural network we parameterize $\c$ using a neural network that approximates $\c$ in the reverse process.
Let $\x_\theta(\x_t, t)$ be an approximation for $\x_0$. Then we get the following reverse Rrocess (for brevity we write $\x_\theta(\x_t, t)$ as $\x_\theta$, and $\cp$ denotes an approximation to $\c$ in the reverse process.):

\begin{align}\label{eqn:p_theta_unavailable_context}
    & p_\theta(\x_s | \x_t) \nonumber \\
    & = \mathcal{N} \left ( \bmu_p = \frac{\balpha_{t | s}(\cp)\bsigma^2_{s}(\cp)}{\bsigma^2_{t}(\cp)} \x_t + \frac{\bsigma^2_{t | s}(\cp)\balpha_s(\cp)}{\bsigma_t^2(\cp)} \x_\theta,
    \; \bSigma_p = \diag \left(\frac{\bsigma^2_s(\cp) \bsigma^2_{t | s}(\cp)}{\bsigma_t^2(\cp)} \right) \right)
\end{align}

Consider the limiting case where $T \to \infty$. Let's analyze the 2 terms in \Eqn{eqn:diffusion_loss_unavailable_context} separately.

Using \Eqn{eqn:q_context_conditioned} and \Eqn{eqn:p_latent}, \textbf{term 1} in \Eqn{eqn:diffusion_loss_unavailable_context} simplifies in the following manner:
\begin{align}\label{eqn:term1}
    & \lim_{T \to \infty} \frac{T}{2}
        (\bmu_q - \bmu_p)^\top 
        \bSigma^{- 1}_\theta
        (\bmu_q - \bmu_p) \nonumber \\
    & \lim_{T \to \infty} \frac{T}{2}
        \sum_{i=1}^d
        \frac{((\bmu_q)_i - (\bmu_p)_i)^2}{(\bSigma_\theta)_i}\\
    & \text{\footnotesize Substituting 1 / T as $\delta$} \nonumber \\
    & \lim_{\delta \to 0^+} \sum_{i=1}^d \frac{1}{\delta \sigmai^2(\x_\theta, t - \delta) \left(1 - \frac{\nui(\x_\theta, t)}{\nui(\x_\theta, t - \delta)}\right)} \times \nonumber \\
    & \hspace{1.5cm} \Biggl[
        \frac{\alphai(\x, t - \delta)}{\alphai(\x, t)} \frac{\nui(\x, t)}{\nui(\x, t - \delta)}\z_t + \alphai(\x, t - \delta) \left(1 - \frac{\nui(\x, t)}{\nui(\x, t - \delta)} \right) x_i
         \nonumber \\
    & \hspace{1.5cm} - \frac{\alphai(\x_\theta, t - \delta)}{\alphai(\x_\theta, t)} \frac{\nui(\x_\theta, t)}{\nui(\x_\theta, t - \delta)}\z_t + \alphai(\x_\theta, t - \delta) \left(1 - \frac{\nui(\x_\theta, t)}{\nui(\x_\theta, t - \delta)}\right)(x_\theta)_i   \Biggl]^2
\end{align}

Consider the scalar case:
substituting $\delta = 1 / T$,

\begin{align}
    & \lim_{\delta \to 0} \frac{1}{\delta \sigma^2(\x_\theta, t - \delta) \left(1 - \frac{\nu(\x_\theta, t)}{\nu(\x_\theta, t - \delta)}\right)} \times \nonumber \\
    & \Biggl[
        \frac{\alpha(\x, t - \delta)}{\alpha(\x, t)} \frac{\nu(\x, t)}{\nu(\x, t - \delta)}\z_t + \alpha(\x, t - \delta) \left(1 - \frac{\nu(\x, t)}{\nu(\x, t - \delta)} \right) \x
         \nonumber \\
    & - \frac{\alpha(\x_\theta, t - \delta)}{\alpha(\x_\theta, t)} \frac{\nu(\x_\theta, t)}{\nu(\x_\theta, t - \delta)}\z_t + \alpha(\x_\theta, t - \delta) \left(1 - \frac{\nu(\x_\theta, t)}{\nu(\x_\theta, t - \delta)}\right)\x_\theta   \Biggl]^2
\end{align}

Notice that this equation is in indeterminate for when we substitute $\delta = 0$. One can apply L'Hospital rule twice or break it down into 3 terms below. For this reason let's write it as

\begin{align}
    \text{expression 1:}
    & \lim_{\delta \to 0} \frac{1}{\delta} \times \Biggl[
        \frac{\alpha(\x, t - \delta)}{\alpha(\x, t)} \frac{\nu(\x, t)}{\nu(\x, t - \delta)}\z_t + \alpha(\x, t - \delta) \left(1 - \frac{\nu(\x, t)}{\nu(\x, t - \delta)} \right) \x
         \nonumber \\
    & - \frac{\alpha(\x_\theta, t - \delta)}{\alpha(\x_\theta, t)} \frac{\nu(\x_\theta, t)}{\nu(\x_\theta, t - \delta)}\z_t + \alpha(\x_\theta, t - \delta) \left(1 - \frac{\nu(\x_\theta, t)}{\nu(\x_\theta, t - \delta)}\right)\x_\theta   \Biggl] \\
    \text{expression 2:}
    & \lim_{\delta \to 0} \frac{1}{\left(1 - \frac{\nu(\x_\theta, t)}{\nu(\x_\theta, t - \delta)}\right)} \times \Biggl[
        \frac{\alpha(\x, t - \delta)}{\alpha(\x, t)} \frac{\nu(\x, t)}{\nu(\x, t - \delta)}\z_t + \alpha(\x, t - \delta) \left(1 - \frac{\nu(\x, t)}{\nu(\x, t - \delta)} \right) \x
         \nonumber \\
    & - \frac{\alpha(\x_\theta, t - \delta)}{\alpha(\x_\theta, t)} \frac{\nu(\x_\theta, t)}{\nu(\x_\theta, t - \delta)}\z_t + \alpha(\x_\theta, t - \delta) \left(1 - \frac{\nu(\x_\theta, t)}{\nu(\x_\theta, t - \delta)}\right)\x_\theta   \Biggl]^2
\end{align}

Applying L'Hospital rule in expression 1 we get,
\begin{align}
    \frac{d}{d \delta} \left(\frac{\alpha(\x, t - \delta)}{\alpha(\x, t)} \frac{\nu(\x, t)}{\nu(\x, t - \delta)} \right)\Bigg|_{\delta = 0}
    & = \frac{\nu(\x, t)}{\alpha(\x, t)}\frac{-\nu(\x, t) \alpha'(\x, t) + \alpha(\x, t)\nu'(\x, t)}{\nu^2(\x, t)} \nonumber \\
    & = \frac{- \alpha'(\x, t)}{\alpha(\x, t)} + \frac{\nu'(\x, t)}{\nu(\x, t)} \\
    \frac{d}{d \delta} \alpha(\x, t -\delta) \left(1 - \frac{\nu(\x, t)}{\nu(\x, t - \delta)} \right)\Bigg|_{\delta = 0} & = - \alpha(\x, t) \frac{\nu'(\x, t)}{\nu(\x, t)}
\end{align}

\begin{align}
    \Biggl[ \left( \frac{- \alpha'(\x, t)}{\alpha(\x, t)} + \frac{\nu'(\x, t)}{\nu(\x, t)} +
    \frac{\alpha'(\x_\theta, t)}{\alpha(\x_\theta, t)} - \frac{\nu'(\x_\theta, t)}{\nu(\x_\theta, t)}\right)\z_t \\
    \hspace{2cm} - \alpha(\x, t) \frac{\nu'(\x, t)}{\nu(\x, t)}\x
    + \alpha(\x_\theta, t) \frac{\nu'(\x_\theta, t)}{\nu(\x_\theta, t)}\x_\theta \Biggl]^2 \times \frac{\nu(\x, t)}{\nu'(\x, t)}
\end{align}

Thus the final result:

\begin{align}\label{eqn:term1_per_dim}
    & \sum_{i=1}^d \Biggl[ \left( \frac{- \alphai'(\x, t)}{\alphai(\x, t)} + \frac{\nui'(\x, t)}{\nui(\x, t)} +
    \frac{\alphai'(\x_\theta, t)}{\alphai(\x_\theta, t)} - \frac{\nui'(\x_\theta, t)}{\nui(\x_\theta, t)}\right)\z_t \nonumber \\
    & \hspace{2cm} - \alphai(\x, t) \frac{\nui'(\x, t)}{\nui(\x, t)}\x
    + \alphai(\x_\theta, t) \frac{\nui'(\x_\theta, t)}{\nui(\x_\theta, t)}\x_\theta \Biggl]^2 \times \frac{\nui(\x, t)}{\nui'(\x, t)} \nonumber \\
    & = \Lambda^\top \text{diag} \left(\frac{\snr(\x, t)}{\snr'(\x, t)} \right) \Lambda \nonumber \\
    & \text{\tiny where $\Lambda = \Biggl[ \left( \frac{- \balpha'(\x, t)}{\balpha(\x, t)} + \frac{\snr'(\x, t)}{\snr(\x, t)} + \frac{\balpha'(\x_\theta, t)}{\balpha(\x_\theta, t)} - \frac{\snr'(\x_\theta, t)}{\snr(\x_\theta, t)}\right)\z_t - \balpha(\x, t) \frac{\snr'(\x, t)}{\snr(\x, t)}\x
    + \balpha(\x_\theta, t) \frac{\snr'(\x_\theta, t)}{\snr(\x_\theta, t)}\x_\theta \Biggl]$}
\end{align}

For the second term we have the following:

\begin{align}\label{eqn:term2}
    & \lim_{T \to \infty} \frac{T}{2} \left(
            \text{tr} \left(\bSigma_q \bSigma_p^{-1} - \bfI_n\right)
            - \log \frac{|\bSigma_q|}{|\bSigma_p|} \right) \nonumber \\
    & = \lim_{T \to \infty} \frac{T}{2} \Biggl[
            \text{tr} \left(\diag \left(\bsigma^2(\c, s)\left(1 - \frac{\snr(\c, t)}{\snr(\c, s)}\right)\right) \bigg /  \diag \left(\bsigma^2(\cp, s) \left(1 - \frac{\snr(\cp, t)}{\snr(\cp, s)}\right) \right) - \bfI_n\right) \nonumber \\
    & \hspace{2cm} - \log \frac{\Bigg|\diag \left(\bsigma^2(\c, s)(1 - \frac{\snr(\c, t)}{\snr(\c, s)})\right)\Bigg|}{\Bigg|\diag \left(\bsigma^2(\cp, s)(1 - \frac{\snr(\cp, t)}{\snr(\cp, s)}) \right)\Bigg|} \Biggl]  \nonumber \\
    & = \lim_{T \to \infty} \frac{T}{2} \sum_{i=1}^d\left(
            \frac{ \sigmai^2(\c, s)\left(1 - \frac{\nui(\c, t)}{\nui(\c, s)}\right)}{  \sigmai^2(\cp, s) \left(1 - \frac{\nui(\cp, t)}{\nui(\cp, s)}\right)} - 1
        - \log \frac{ \sigmai^2(\c, s)\left(1 - \frac{\nui(\c, t)}{\nui(\c, s)}\right)}{  \sigmai^2(\cp, s) \left(1 - \frac{\nui(\cp, t)}{\nui(\cp, s)}\right)} \right)  \\
\end{align}

Let $p_i = \frac{ \sigmai^2(\c, s)\left(1 - \frac{\nui(\c, t)}{\nui(\c, s)}\right)}{  \sigmai^2(\cp, s) \left(1 - \frac{\nui(\cp, t)}{\nui(\cp, s)}\right)}$

The sequence $\lim_{T \to \infty} \frac{T}{2} \sum_{i=1}^d (p_i - 1 - \log p_i)$ converges iff $\lim_{T \to \infty}\sum_{i=1}^d (p_i - 1 - \log p_i)= 0$. Notice that the function $f(x) = x - 1 - \log x \geq 0\;\; \forall x \in \mathbb{R}$ and the equality holds for $x = 1$. Thus, the condition $\lim_{T \to \infty} \frac{T}{2} \sum_{i=1}^d (p_i - 1 - \log p_i)$ holds iff $\lim_{T \to \infty} p_i = 0 \;\; \forall i \in \{1, \dots, d\}$. Thus,

\begin{align}
    & \lim_{T \to \infty} p_i = 1 \nonumber \\
    & \implies \lim_{T \to \infty} \left(
        \frac{ \sigmai^2(\c, s)\left(1 - \frac{\nui(\c, t)}{\nui(\c, s)}\right)}{  \sigmai^2(\cp, s) \left(1 - \frac{\nui(\cp, t)}{\nui(\cp, s)}\right)} \right) = 1 \nonumber \\
    & \text{\footnotesize Substituting 1/T as $\delta$,} \nonumber \\
    & \implies \lim_{\delta \to 0^+} \left(
        \frac{ \sigmai^2(\c, t - \delta)\left(1 - \frac{\nui(\c, t)}{\nui(\c, t - \delta)}\right)}{  \sigmai^2(\cp, t - \delta) \left(1 - \frac{\nui(\cp, t)}{\nui(\cp, t - \delta)}\right)} \right) = 1 \nonumber \\
    & \implies \frac{\sigmai^2(\c, t)}{\sigmai^2(\cp, t)} \lim_{\delta \to 0^+} \left(
        \frac{ 1 - \frac{\nui(\c, t)}{\nui(\c, t - \delta)})}{1 - \frac{\nui(\cp, t)}{\nui(\cp, t - \delta)}} \right) = 1 \nonumber \\
    & \text{\footnotesize Applying L'Hospital rule,} \nonumber \\
    & \implies \frac{\sigmai^2(\c, t)}{\sigmai^2(\cp, t)} \left(
        \frac{ \frac{- \nui'(\c, t)}{\nui(\c, t)})}{\frac{- \nui'(\cp, t)}{\nui(\cp, t)}} \right) = 1 \nonumber \\
    & \implies \frac{\sigmai^2(\c, t)}{\sigmai^2(\cp, t)} \left(
        \frac{ \nui'(\c, t)\nui(\cp, t)}{{\nui(\c, t) \nui'(\cp, t)})} \right) = 1
\end{align}

In the vector form the above equation can be written as,
\begin{equation}\label{eqn:term2_convergence}
    \frac{
        \bsigma^2_t(\c) \snr_t(\cp) \nabla_t\snr(\c, t)}{
        \bsigma^2_t(\cp) \snr_t(\c)\nabla_t\snr(\cp, t)} = \one
\end{equation}

\Eqn{eqn:term2_convergence} holds if:
\begin{itemize}
    \item $x_\theta = x_0$ i.e. the unet can perfectly map $\x_t$ to $\x_0$  $\forall t \in [0, 1]$ which is unrealistic.
    \item Clever parameterizations for $\bsigma, \balpha, \snr$ that ensure \Eqn{eqn:term2_convergence} holds.
\end{itemize}
Because of aforementioned challenges we evaluate this method with finite $T=1000$. We demonstrate the performance of the model empirically in~\fig{fig:context_plots}.

\subsubsection{Recovering VDM}
If we substitute $\snr_t(\c), \snr_t(\cp)$ with $\snr(t)$ (since the SNR isn't conditioned on the context $\c$), $\bsigma_t(\cp), \bsigma_t(\c)$ with $\sigma_t$ and $\balpha_t(\cp), \balpha_t(\c)$ with $\balpha_t$,
\Eqn{eqn:term1} reduces to the intermediate loss in VDM i.e. $\frac{1}{2} (\x_\theta - \x_0)^\top  \nabla_t \snr(t) \; (\x_\theta - \x_0 )$ and
\Eqn{eqn:term2} reduces to 0.

\subsection{Experimental results}\label{appendix:subsec:context_ablations}
\begin{figure}
    \centering
    \includegraphics[width=0.8\linewidth]{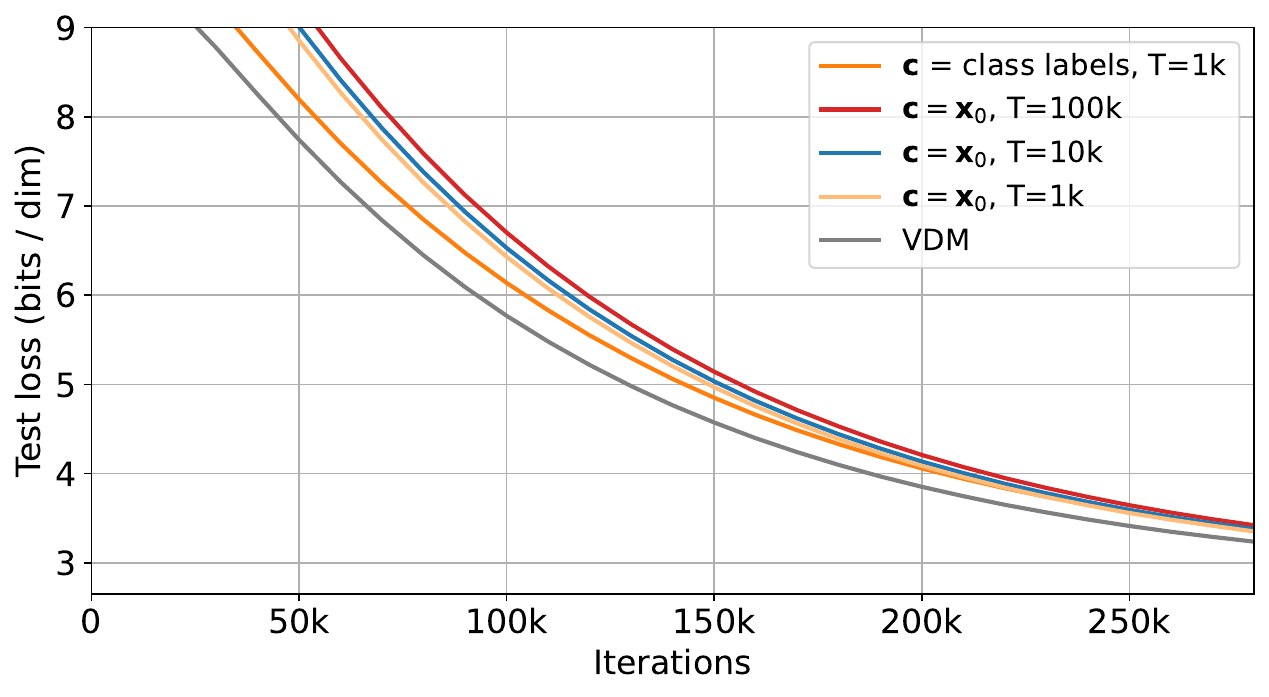}
    \caption{For $\c=$ ``class labels'' or $\c=\x_0$ the likelihood estimates are worse than VDM. For $\c=\x_0$, we see that the VLB degrades with increasing T, but for VDM and \method{}, it improves with increasing T. This empirical observation is consistent with our mathematical insights earlier. As these models consistently exhibit inferior performance w.r.t VDM, in line with our initial conjectures, we refrain from training them beyond 300k iterations due to the substantial computational cost involved.}
    \label{fig:context_plots}
\end{figure}

In \fig{fig:context_plots} we demonstrate that the multivariate diffusion processes where $\c=$ ``class labels'' or $\c=\x_0$ perform worse than VDM. Since a continuous time formulation i.e. $T \to \infty$ for the case when $\c=\x_0$ isn't possible (unlike \method{} or VDM) we evaluate these models in the discrete time setting where we use $T=1000$. Furthermore we also ablate $T={10k, 100k}$ for $\c=\x_0$ to show that the VLB degrades with increasing T whereas for VDM and \method{} it improves for increasing T; see \citet{kingma2021variational}. This empirical observation is consistent with our mathematical insights earlier. As these models consistently exhibit inferior performance w.r.t VDM, in line with our initial conjectures, we refrain from training them beyond 300k iterations due to the substantial computational cost involved.

\section{\method{}: MUltivariate Latent Auxiliary variable Noise Schedule}\label{appendix:mulan}

\subsection{Parameterization in the reverse process}
\subsubsection{Noise parameterization}\label{appendix:subsec:noise_param}
Since the forward pass is given by $\x_t = \balpha_t(\z) \x_0 + \bsigma_t(\z) \noise_t$, we can write the noise $\noise_t$ in terms of $\x_0, \x_t$ in the following manner:
\begin{equation}\label{eqn:input_to_noise}
    \noise_t = \frac{\x_t - \balpha_t(\z) \x_0}{\bsigma_t(\z)}
\end{equation}
Following \citet{dhariwal2021diffusion, kingma2021variational}, instead of parameterizing $\x_\theta(\x_t, \z, t)$ using a neural network, we use~\Eqn{eqn:input_to_noise} to parameterize the denoising model in terms of a noise prediction model $\unet(\x_t, \z, t)$,
\begin{equation}\label{eqn:noise_param_p}
    \unet(\x_t, \z, t) = \frac{\x_t - \balpha_t(\z) \x_\theta(\x_t, \z, t)}{\bsigma_t(\z)}
\end{equation}

\subsubsection{Velocity parameterization}\label{appendix:subsec:velocity_param}
Following~\citet{salimans2022progressive, zheng2023improved}, we explore another parameterization of the denoising network which is given by
\begin{equation}\label{eqn:velocity_param_p}
    \mathbf{v}_\theta(\x_t, \z, t) = \frac{\balpha_t(\z) \x_t - \x_\theta(\x_t, \z, t)}{\bsigma_t(\z)}
\end{equation}

In practice, v-parameterization leads to a better performance than noise parameterization; as illustrated in~\tab{tab:parameterization_comparision}.

\begin{table}
    \centering
    \caption{Likelihood in bits per dimension (BPD) (mean and 95\% confidence interval), on the test set of CIFAR-10 computed using VLB estimate.}
    \begin{tabular}{c|c|c}
         parameterization & Num training steps & CIFAR-10 ($\downarrow$) \\
         \hline
         Noise parameterization & $10$M & $2.60 \pm 10^{-3}$ \\
         v-parameterization & $8$M & $2.59 \pm 10^{-3}$ \\
    \end{tabular}
    \label{tab:parameterization_comparision}
\end{table}

\subsection{Polynomial Noise Schedule}\label{appendix:subsec:polynomial_ns}
Let $\fn$ be a scalar-valued polynomial of degree $n$ with coefficients $\psi \in \mathbb{R}^{n + 1}$ expressed as:
$$\fn = \psi_n x^n + \psi_{n-1} x^{n-1} + \dots + \psi_1 x + \psi_0,$$
and denote its derivative with respect to $x$ as $\frac{d}{dx}\fn$, represented by $\fngrad$.
Now we'd like to find least $n$ such that $\fn$ satisfies the following properties:
\begin{enumerate}
    \item $\fn$ is monotonically increasing, i.e. $\fngrad \geq 0 \; \forall x \in \mathbb{R}, \psi \in \mathbb{R}^{n+1}$.
    \item $\fngrada = 0, \fngradb = 0$ $\exists x_1, x_2 \in \mathbb{C}, x_1 \neq x_2, \forall \psi \in \mathbb{R}^{n+1}$.
\end{enumerate}
For the \textbf{first} condition to hold, we can design $\fngrad$ such that it's a perfect square with real / imaginary roots. That way $\fngrad \geq 0 \; \forall x \in \mathbb{R}, \psi \in \mathbb{R}^{n+1}$ . This means that $\fngrad$ is an even degree polynomial, i.e. the degree of $\fngrad$ can take the following values: $2, 4, \dots$. Also, note that at least half of the roots of $\fngrad$ are repeated since $\fngrad$ can be expressed as a perfect square, i.e., if $\fngrad$ has a degree 2 then it has exactly 1 unique root (real / imaginary),  if $\fngrad$ has a degree 4 then it has at most 2 unique roots (real / imaginary), and so on.

For the \textbf{second} condition to hold, $\fngrad$ needs to have at least 2 unique roots $\exists \psi \in \mathbb{R}^{n+1}$. For this reason $\fngrad$ is a polynomial of degree 4. Thus, $\fngrad$ can be written as $\fngrad = (\psi_3x^2 + \psi_2x + \psi_1)^2$. This ensures that $\exists \psi \in \mathbb{R}^5$ s.t. $\fngrad = 0$ twice in $x \in \mathbb{R}$, and $\fngrad \geq 0 \; \forall \psi \in \mathbb{R}^5$.

Thus, $\fn$ takes the following functional form:
\begin{align}
    \fn & = \int (\psi_3x^2 + \psi_2x + \psi_1)^2 dx \nonumber \\
         & = \frac{\psi_3^2}{5} x^5 + \frac{\psi_3 \psi_2}{2}x^4 + \frac{\psi_2^2 + 2\psi_3\psi_1}{3} x^3 + \psi_2 \psi_1 x^2 + \psi_1^2x + \text{constant.}
\end{align}

For the above-mentioned reasons we express $\ns_\phi(\c ,t): \mathbb{R}^m \times [0, 1] \to \mathbb{R}^d$ as a degree 5 polynomial in $t$. We define neural networks $\a(\c): \mathbb{R}^m \to \mathbb{R}^d$, $\b(\c): \mathbb{R}^m \to \mathbb{R}^d$, and $\d(\c): \mathbb{R}^m \to \mathbb{R}^d$ with parameters $\phi$. Let $f_\phi: \mathbb{R}^m \times [0, 1] \to \mathbb{R}^d$ be defined as:
$$f_\phi(\c, t) = \frac{\a^2(\c)}{5} t^5 + \frac{\a(\c)\b(\c)}{2}t^4 + \frac{\b^2(\c) + 2\a(\c)\d(\c)}{3} t^3 + \b(\c)\d(\c) t^2 + \d^2(\c)t$$ 
where the multiplication and division operations are  elementwise.
The the noise schedule, $\ns(\c, t)$, is given as follows:
\begin{equation}\label{eqn:appendix:polynomial}
    \ns_\phi(\c, t) =
    \nsmin + (\nsmax - \nsmin) 
    \frac{f_\phi(\c, t)}{f_\phi(\c, t=1)}
\end{equation}

Notice that $\ns_\phi(\c, t)$ has these interesting properties:
\begin{itemize}
    \item Is increasing in $t \in [0, 1]$ which is crucial as mentioned in \sec{subsec:path_integral}.
    \item $\ns_\phi(\c, t)$ has end points at $t=0$ and $t=1$ which the user can specify via $\gamma_\text{min}$ and $\gamma_\text{max}$. Specificaly, $\ns_\phi(\c, t=0) = \gamma_\text{min}\one$ and $\ns_\phi(\c, t=1) = \gamma_\text{max}\one$.
    \item Its time-derivative i.e. $\nabla_t\ns_\phi(\c, t)$ \textbf{can} be zero twice in $t \in [0, 1]$. This isn't a necessary condition but it's nice to have a flexible noise schedule whose time-derivative can be $0$ at the beginning and the end of the diffusion process.
\end{itemize}

\subsection{Variational Lower Bound}\label{appendix:auxiliary_latent_vlb}

In this section we derive the VLB. For ease of reading we use the notation $\x_t$ to denote $\x_{\t}$ and $\x_{t-1}$ to denote $\x_{t(i-1)} \equiv \x_\s$ in the following derivation.
{\footnotesize
\begin{align}
    & -\log \p (\x_0) \nonumber \\
    & \leq \mathbb{E}_\q \left [ - \log \frac{\p (\z, \x_{0:T})}{\q(\z,\x_{1:T} | \x_0)} \right] \nonumber \\
    & = \mathbb{E}_\q \left [
        - \log \frac{\p (\x_{0:T-1} | \z, \x_T)}{\q(\z, \x_{1:T} | \x_0)}
        -\log \p(\x_T) -\log \p(\z) \right] \nonumber \\
    & = \mathbb{E}_\q \left [
        - \log \frac{\p (\x_{0:T-1} | \z, \x_T)}{\q(\x_{1:T} | \z, \x_0)}
        - \log \frac{1}{\q(\z| \x_0)}
        -\log \p(\x_T) -\log \p(\z) \right] \nonumber \\
    & = \mathbb{E}_\q \left [
        - \log \frac{\p (\x_{0:T-1} |\z, \x_T)}{\q(\x_{1:T} | \z, \x_0)}
        -\log \p(\x_T)
        -\log \frac{\p(\z)}{\q(\z| \x_0)} \right] \nonumber \\
    & = \mathbb{E}_\q \left [
        - \sum_{t=1}^{T}\log \frac{\p (\x_{t-1} | \z, \x_{t})}{\q(\x_t | \x_{t-1}, \z, \x_0)}
        -\log \p(\x_T)
        -\log \frac{\p(z)}{\q(z| \x_0)}
        \right] \nonumber \\
    & = \mathbb{E}_\q \left [
        - \log \frac{\p (\x_{0} | \z, \x_{1})}{\q(\x_1 | \x_{0}, \z)}
        - \sum_{t=2}^{T}\log \frac{\p (\x_{t-1} | \z, \x_{t})}{\q(\x_t | \x_{t-1}, \z, \x_0)}
        -\log \p(\x_T)
        -\log \frac{\p(z)}{\q(\z| \x_0)}
        \right] \nonumber \\
    & = \mathbb{E}_\q \left [
        - \log \frac{\p (\x_{0} | \z, \x_{1})}{\q(\x_1 | \x_{0}, \z)}
        - \sum_{t=2}^{T}\log
            \frac{\p (\x_{t-1} | \z, \x_{t})\q(\x_{t-1}| \z, \x_0)}{
            \q(\x_{t-1} | \x_t, \z, \x_0)\q(\x_t| \z, \x_0)}
        -\log \p(\x_T)
        -\log \frac{\p(\z)}{\q(\z| \x_0)}
        \right] \nonumber \\
    & = \mathbb{E}_\q \left [
        - \log \frac{\p (\x_{0} | \z, \x_{1})}{\q(\x_1 | \x_{0}, \z)}
        - \sum_{t=2}^{T}\log
            \frac{\p (\x_{t-1} | \z, \x_{t})}{\q(\x_{t-1} | \x_t, \z, \x_0)}
        - \sum_{t=2}^{T}\log
            \frac{\q(\x_{t-1}| \z, \x_0)}{\q(\x_t| \z, \x_0)} 
        -\log \p(\x_T)
        -\log \frac{\p(z)}{\q(\z| \x_0)}
        \right] \nonumber \\
    & = \mathbb{E}_\q \left [
        - \log \frac{\p (\x_{0} | \z, \x_{1})}{\q(\x_1 | \x_{0}, \z)}
        - \sum_{t=2}^{T}\log
            \frac{\p (\x_{t-1} | \z, \x_{t})}{\q(\x_{t-1} | \x_t, \z, \x_0)}
        - \log \frac{q(\x_{1}| \z, \x_0)}{\q(\x_T| \z, \x_0)}
        -\log \p(\x_T)
        -\log \frac{\p(\z)}{\q(\z| \x_0)}
        \right] \nonumber \\
    & = \mathbb{E}_\q \left [
        - \log \p (\x_{0} | \z, \x_{1})
        - \sum_{t=2}^{T}\log
            \frac{\p (\x_{t-1} | \z, \x_{t})}{\q(\x_{t-1} | \x_t, \z, \x_0)} 
        - \log \frac{1}{\q(\x_T| \z, \x_0)}
        -\log \p(\x_T)
        -\log \frac{\p(\z)}{\q(\z| \x_0)}
        \right] \nonumber \\
    & = \mathbb{E}_\q \left [
        - \log \p (\x_{0} | \z, \x_{1})
        - \sum_{t=2}^{T}\log
            \frac{\p (\x_{t-1} | \z, \x_{t})}{\q(\x_{t-1} | \x_t, \z, \x_0)}
        - \log \frac{\p(\x_T)}{\q(\x_T| \z, \x_0)}
        -\log \frac{\p(\z)}{\q(\z| \x_0)}
        \right] \nonumber \\
    & = \mathbb{E}_\q \left[
        \underbrace{- \log \p (\x_{0} | \z, \x_{1})}_{\lossrecons}
        + \underbrace{
        \sum_{t=2}^{T}\kl [ \p (\x_{t-1} | \z, \x_{t}) \|  \q(\x_{t-1} | \x_t, \z, \x_0)]}_{\lossdiff} \right] \nonumber \\
    &  \hspace{4mm} 
        + \mathbb{E}_\q \left[\underbrace{\kl [\p(\x_T) \| \q(\x_T| \z, \x_0)]}_{\lossprior}
        + \underbrace{\kl [\p(\z) \| q(\z| \x_0)]}_{\lossz} \right]
\end{align}
}
Switching back to the notation used throughout the paper, the VLB is given as:
\begin{align}
    & -\log \p (\x_0) \nonumber \\
    & \leq \mathbb{E}_\q \left[
        \underbrace{- \log \p (\x_{0} | \z, \x_{1})}_{\lossrecons}
        + \underbrace{
        \sum_{i=2}^{T}\kl [ \p (\x_\s | \z, \x_{\t}) \|  \q(\x_{\s} | \x_\t, \z, \x_0)]}_{\lossdiff} \right] \nonumber \\
    &  \hspace{4mm} 
        + \mathbb{E}_\q \left[\underbrace{\kl [\p(\x_1) \| \q(\x_1| \z, \x_0)]}_{\lossprior}
        + \underbrace{\kl [\p(\z) \| \q(\z| \x_0)]}_{\lossz} \right]
\end{align}

\subsection{Diffusion Loss}

To derive the diffusion loss, $\lossdiff$ in
\Eqn{eqn:elbo}, we first derive an expression for
$\kl (\q(\x_{s} | \z, \x_t, \x_0) \| \p(\x_{s} | \z, \x_t))$ using \Eqn{eqn:q_context_conditioned} and 
\Eqn{eqn:p_latent} in the following manner (details in \supp{appendix:mulan}):
\begin{align}\label{eqn:latent_kl}
    & \kl (\q(\x_{s} | \z, \x_t, \x_0) \| \p(\x_{s} | \z, \x_t)) \nonumber \\
    & =
    \frac{1}{2}
    \left(
        (\bmu_\q - \bmu_p)^\top 
        \bSigma^{- 1}_\theta
        (\bmu_\q - \bmu_p)
        + \text{tr} \left(\bSigma_\q \bSigma_p^{-1} - \bfI_n \right)
        - \log \frac{|\bSigma_\q|}{|\bSigma_p|}
    \right) \nonumber \\
    & =
    \frac{1}{2}
    \left(
        (\x_0 - \x_\theta)^\top 
        \diag (\snr(\z, s) - \snr(\z, t))
        (\x_0 - \x_\theta)
    \right) 
\end{align}

Let $\lim_{T \to \infty} T (\snr_s(z) - \snr_t(z)) = -\nabla_t\snr(\z, t)$ be the partial derivative of the vector $\snr(\z, t)$ w.r.t scalar $t$. Then we derive the diffusion loss, $\lossdiff$, for the continuous case in the following manner (for brevity we use the notation $s$ for $s(i) = (i - 1) / T$ and $t$ for $t(i) = i / T$):
\begin{align}\label{eqn:diffusion_loss_continuous2}
    & \lossdiff \nonumber \\
    & = \lim_{T \to \infty} \frac{1}{2} \sum_{i=2}^T \mathbb{E}_{
        \epsilon \sim \mathcal{N}(0, \bfI_n)}
        \kl (q(\x_s | \x_t, \x_0, \z) \| p_\theta(\x_s | \x_t, \z)) \nonumber \\
    &  \hspace{5mm} \text{Using \Eqn{eqn:latent_kl} we get,} \nonumber \\
    & = \lim_{T \to \infty} \frac{1}{2} \sum_{i=2}^T \mathbb{E}_{
        \epsilon \sim \mathcal{N}(0, \bfI_n)}
            (\x_0 - \x_\theta(\x_t, \t))^\top 
            \diag \left(\snr(\s, \z) - \snr(\t, \z) \right)
            (\x_0 - \x_\theta(\x_t, \t)) \nonumber \\
    & = \frac{1}{2} \mathbb{E}_{\epsilon \sim \mathcal{N}(0, \bfI_n)} \left [
        \lim_{T \to \infty} \sum_{i=2}^T 
            T (\x_0 - \x_\theta(\x_t, \t))^\top 
            \diag \left(\snr(\s, \z) - \snr(\t, \z) \right)
            (\x_0 - \x_\theta(\x_t, \t)) \frac{1}{T}\right] \nonumber \\
    & \hspace{5mm} \text{Using the fact that \footnotesize $\lim_{T \to \infty} T \left(\snr(s, \z) - \snr(\z, t) \right) = - \nabla_t\snr(t, \z)$ we get,} \nonumber \\
    & = - \frac{1}{2} \mathbb{E}_{t \sim \{0, \dots, 1\}} 
    \left[
        (\x_0 - \x_\theta(\x_t, t))^\top 
        \left( \nabla_t \snr_t(z) \right)
        (\x_0 - \x_\theta(\x_t, t))
    \right] \nonumber \\
    & \hspace{5mm} \text{\footnotesize Substituting $\x_0 = \balpha^{-1}_t(\z) (\x_t - \bsigma_t(\z) \noise_t)$ from \Eqn{eqn:input_to_noise} and} \nonumber \\
    & \hspace{5mm} \text{\footnotesize Substituting $\x_\theta(\x_t, \z, t) = \balpha^{-1}_t(\z) (\x_t - \bsigma_t(\z) \unet(\x_t, t))$ from \Eqn{eqn:noise_param_p} we get,} \nonumber \\
    & = - \frac{1}{2} \mathbb{E}_{t \sim [0, 1]}
    \left[
        (\noise_t - \unet(\x_t, t))^\top 
        \left(\frac{\bsigma_t^2(\z)}{\balpha_t^2(\z)} \times \nabla_t\snr_t(\z)\right)
        (\noise_t - \unet(\x_t, t))
    \right] \nonumber \\
    & \hspace{5mm} \text{\footnotesize Let $\snr^{-1}(\z, t)$ denote the reciprocal of the values in the vector $\snr(\z, t)$.} \nonumber \\
    & = - \frac{1}{2} \mathbb{E}_{t \sim [0, 1]}
    \left[
        (\noise_t - \unet(\x_t, t))^\top 
        \diag \left(\snr^{-1}(t)(\z) \nabla_t\snr_t(\z)\right)
        (\noise_t - \unet(\x_t, t))
    \right] \nonumber \\
    & \hspace{5mm} \text{\footnotesize Substituting $\snr(\z, t) = \exp(-\ns(\z, t))$ from \sec{appendix:subsec:noise_param}} \nonumber \\
    & = - \frac{1}{2} \mathbb{E}_{t \sim [0, 1]}
    \left[
        (\noise_t - \unet(\x_t, t))^\top 
        \diag \left(\exp{(\ns (\z, t))} \nabla_t\exp{(- \ns (\z, t))}\right)
        (\noise_t - \unet(\x_t, t))
    \right] \nonumber \\
    & = \frac{1}{2} \mathbb{E}_{t \sim [0, 1]}
    \left[
        (\noise_t - \unet(\x_t, t))^\top 
        \diag \left(\exp{(\ns (\z, t))} \exp{(-\ns (\z, t))}\nabla_t\ns (\z, t)\right)
        (\noise_t - \unet(\x_t, t))
    \right] \nonumber \\
    & = \frac{1}{2} \mathbb{E}_{t \sim [0, 1]}
    \left[
        (\noise_t - \unet(\x_t, t))^\top 
        \diag \left( \nabla_t\ns(\z, t) \right)
        (\noise_t - \unet(\x_t, t))
    \right] 
\end{align}

\subsection{Recovering VDM from the Vectorized Representation of the diffusion loss}\label{appendix:recover_vdm_loss_from_path_integral}

Notice that we recover the loss function in VDM when $\snr(\z, t) = \nu(t) \one$ where $\nu_t \in \mathbb{R}^+$ and $\one$ represents a vector of 1s of size $d$ and the noising schedule isn't conditioned on $\z$.

\begin{align}
    \int_{0}^1 \langle \forcefield, \frac{\text{d}}{\text{d}t}\vsnr(t) \rangle \text{d}t & = \int_{0}^1 \langle \forcefieldnoz, \frac{\text{d}}{\text{d}t}(\nu(t)\mathbf{1_n}) \rangle  \text{d}t \nonumber \\
    & = \int_{0}^1 \langle \forcefieldnoz, \one \rangle \nu'(t)\text{d}t \nonumber \\
    & = \int_{0}^1 \nu'(t) \|\forcefieldnoz\|_1^1 \text{d}t \nonumber \\
    & = \int_{0}^1 \nu'(t) \|(\x_0 - \Tilde{\x}_\theta(\x_{\snr(t)}, \snr(t)))\|_2^2 \text{d}t    
\end{align}
$\int_{0}^1 \frac{d}{d t}\nu(t) \|(\x_0 - \Tilde{\x}_\theta(\x_{\snr(t)}, \snr(t)))\|_2^2 dt$ denotes the diffusion loss, $\lossdiff$, as used in VDM; see \citet{kingma2021variational}.

\begin{figure}
    \centering
    \includegraphics[width=\linewidth]{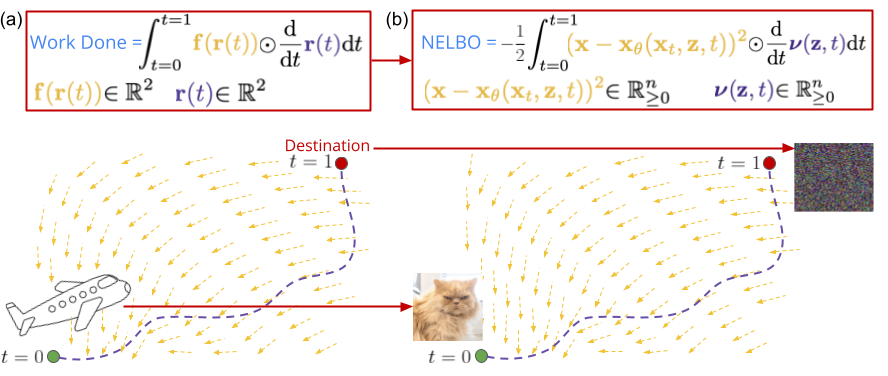}
    \caption{\textbf{(a)} Imagine piloting a plane across a region with cyclones and strong winds, as shown in~\fig{fig:intuitive-explanation}. Plotting a direct, straight-line course through these adverse weather conditions requires more fuel and effort due to increased resistance. By navigating around the cyclones and winds, however, the plane reaches its destination with less energy, even if the route is longer.This intuition translates into mathematical and physical terms. The plane’s trajectory is denoted by $\mathbf{r}(t) \in \mathbb{R}^n_{+}$, while the forces acting on it are represented by $\mathbf{f}(\mathbf{r}(t)) \in \mathbb{R}^n$. The work required to navigate is given by $\int_{0}^{1} \mathbf{f}(\mathbf{r}(t)) \cdot \frac{d}{dt}\mathbf{r}(t) , dt$. Here, the work depends on the trajectory because $\mathbf{f}(\mathbf{r}(t))$ is not a conservative field. \\
    \textbf{(b)} This concept also applies to the diffusion NELBO. From~\Eqn{eqn:dot_product}, it’s clear that the trajectory $\mathbf{r}(t)$ is parameterized by the noise schedule $\snr(\z, t)$, which is influenced by complex forces, $\f$ (analogous to weather patterns), represented by the dimension-wise reconstruction error of the denoising model, $(\x_0 - \x_\theta(\x_t, \z, t))^2$. Thus, the diffusion loss, $\lossdiff$, can be interpreted as the work done along the trajectory $\snr(\z, t)$ in the presence of these vector field forces $\f$. By learning the noise schedule, we can avoid “high-resistance” paths (those where the loss accumulates rapidly), thereby minimizing the overall “energy” expended, as measured by the NELBO. }
    \label{fig:intuitive-explanation}
\end{figure}

\section{Subset Sampling}
Sampling a subset of $k$ items from a collection of collection of $n$ items, $x_1, x_2, \dots, x_3$ belongs to a category of algorithms called reservoir algorithms. In weighted reservoir sampling, every $x_i$ is associated with a weight $w_i \geq 0$. The probability associated with choosing the sequence $\swrs = [i_1, i_2, \dots, i_k]$ be a tuple of indices. Then the probability associated with sampling this sequence is
\begin{equation}
    p(S_{\text{wrs}} | \w) = \frac{w_{i_1}}{Z} \frac{w_{i_2}}{Z - w_{i_1}} \dots \frac{w_{i_k}}{Z - \sum_{j=1}^{k-1}w_{i_j}}    
\end{equation}

\citet{EFRAIMIDIS2006181} give an algorithm for weighted reservoir sampling where each item is assigned a random key $r_i = u_i^{\frac{1}{w_i}}$ where $u_i$ is drawn from a uniform distribution [0, 1] and $w_i$ is the weight of item $x_i$. Let TopK($\mathbf{r}, k$) which takes keys $\mathbf{r}=[r_1, r_2, \dots, r_n]$ and returns a sequence $[i_1, i_2, \dots, i_k]$. \citet{EFRAIMIDIS2006181} proved that TopK($\mathbf{r}, k$) is distributed according to $ p(\swrs | \w)$.

Let's represent a subset $S \in \{0, 1\}^n$ with exactly $k$ non-zero elements that are equal to 1. Then the probability associated with sampling $S$ is given as,
\begin{equation}
    p(S | \w) = \sum_{\swrs \in \Pi(S)}p(\swrs | \w)
\end{equation}
where $\Pi(S)$ denotes all possible permutations of the sequence $S$. By ignoring the ordering of the elements in $\swrs$ we can sample using the same algorithm.
\citet{xie2019reparameterizable} show that this sampling algorithm is equivalent to TopK($\mathbf{\hat r}, k$) where $\mathbf{\hat r}=[\hat r_1, \hat r_2, \dots, \hat r_n]$ where $\hat r_i = - \log (- \log (r_i)) = \log w_i + $ Gumbel(0, 1). 
This holds true because the monotonic transformation $- \log (- \log (x))$ preserves the ordering of the keys and thus TopK($\mathbf{r}, k$) $\equiv$ TopK($\mathbf{\hat r}, k$).

\paragraph{Sum of Gamma Distribution.}\label{appendix:eqn:sog}
\citet{niepert2021implicit} show that adding SOG noise instead of Gumbel noise leads to better performance.

\citet{niepert2021implicit} show that $\z \sim \p(\z; \theta)$ is equivalent to $\z = \arg\max_{y \in Y}\langle \theta + \epsilon_g, y \rangle$ where $\epsilon_g$ is a sample from Sum-of-Gamma distribution given by
\begin{equation}\label{eqn:app:sog}
    \operatorname{SoG}(k, \tau, s)
        = \frac{\tau}{k} \biggl(\sum_{i=1}^{s} \operatorname{Gamma}\Bigl(\frac 1k, \frac ki\Bigr) - \log s\biggr),
\end{equation}
where $s$ is a positive integer and $\operatorname{Gamma}(\alpha, \beta)$ is the Gamma distribution with $(\alpha, \beta)$ as the shape and scale parameters.

And hence, given logits $\log \w$, we sample a $k$-hot vector using TopK($\log \w + \epsilon$). We choose a categorical prior with uniform distribution across $n$ classes. Thus the KL loss term is given by:
\begin{align}
    - \sum_{i=1}^n \frac{w_i}{Z} \log \left (n \frac{w_i}{Z}\right)
\end{align}
    
\section{Experiment Details}\label{section:experiment_details}
\subsection{Model Architecture}
\paragraph{Denoising network.}
Our model architecture is extremely similar to VDM. The UNet of our pixel-space diffusion has an unchanged architecture from~\citet{kingma2021variational}.This structure is specifically designed for optimal performance in maximum likelihood training. We employ features from VDM such as the elimination of internal downsampling/upsampling processes and the integration of Fourier features to enhance fine-scale prediction accuracy. In alignment with the configurations suggested by Kingma et al. (2021), our approach varies depending on the dataset: For CIFAR-10, we employ a U-Net with a depth of 32 and 128 channels; for ImageNet-32, the U-Net also has a depth of 32, but the channel count is increased to 256. Additionally, all these models incorporate a dropout rate of 0.1 in their intermediate layers. 

\paragraph{Encoder network.}
$\q(\z | \x)$ is modeled using a sequence of 4 Resnet blocks with a channel count of 128 for CIFAR-10 and 256 for ImageNet-32 with a drop out of 0.1 in their intermediate layers.

\paragraph{Noise schedule.}
For polynomial noise schedule, we use an MLP that maps the latent vector $\z$ to $\a(\z), \b(\z), \c(\z)$; see ~\Eqn{appendix:subsec:polynomial_ns} for details. The MLP has 2 hidden layers of size $3072$ with \texttt{swish} activation function. The final layer is a linear mapping to $3 \times 3072$ values corresponding to $\a(\z), \b(\z), \c(\z)$. Note that $\a(\z), \b(\z), \c(\z)$ have the same dimensionality of $3072$.

\subsection{Hardware.}
For the ImageNet experiments, we used a single GPU node with 8-A100s. For the cifar-10 experiments, the models were trained on 4 GPUs spanning several GPUs types like V100, A5000s, A6000s, and 3090s with \texttt{float32} precision.

\subsection{Hyperparameters}
We follow the same default training settings as ~\citet{kingma2021variational}. For all our experiments, we use the Adam~\citep{kingma2014adam} optimizer with learning rate
$2 \times 10^{-4}$, exponential decay rates of $\beta_1 = 0.9$, $\beta_2 = 0.99$ and decoupled weight decay~\citep{loshchilov2017decoupled} coefficient of $0.01$. We also maintain an exponential moving average (EMA) of model parameters with an EMA rate of $0.9999$ for evaluation.
For other hyperparameters, we use fixed start and end times which satisfy
$\nsmin = -13.3$, $\nsmax = 5.0$,
which is used in ~\citet{kingma2021variational, zheng2023improved}. 

\section{Datasets and Visualizations}\label{appendix:datasets}
In this section we provide a brief description of the datasets used in the paper and visualize the generated samples and the noise schedules.

\subsection{CIFAR-10}
The CIFAR-10 dataset~\citep{krizhevsky2009learning} is a collection of images consisting of 60,000 $32\times32$ color images in 10 different classes, with each class representing a distinct object or scene. The dataset is divided into 50,000 training images and 10,000 test images, with each class having an equal representation in both sets. The classes in CIFAR-10 include:
Airplane, Automobile, Bird, Cat, Deer, Dog, Frog, Horse, Ship, Truck.

\begin{figure}[H]
    \centering
    \begin{subfigure}[t]{.47\linewidth}
    \includegraphics[width=\linewidth]{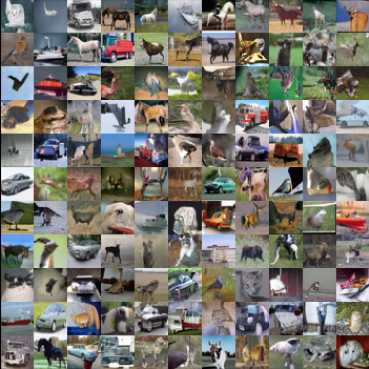}
    \caption{\method{} with velocity reparameterization after 8M training iterations.}
    \label{fig:cifar10_mulan_samples}
    \end{subfigure}\hfill
    \begin{subfigure}[t]{.47\linewidth}
    \includegraphics[width=\linewidth]{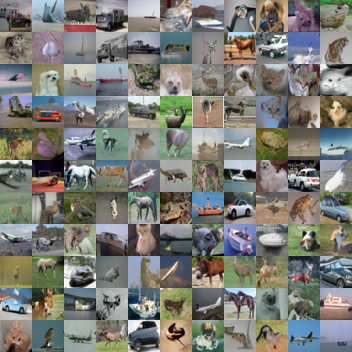}
    \caption{VDM after 10M training iterations.}
    \label{fig:cifar10_vdm_samples}
    \end{subfigure}
    \caption{CIFAR-10 samples generated by different methods.}
    \label{fig:samples}
\end{figure}

Randomly generated samples for the CIFAR-10 datasaet are provided in~\fig{fig:cifar10_mulan_samples} for \method{} and ~\fig{fig:cifar10_vdm_samples} for VDM. We visualize the noise schedule in \fig{fig:noise_schedules_all}.

\subsection{ImageNet-32}

ImageNet-32 is a dataset derived from ImageNet~\citep{5206848}, where the original images have been resized to a resolution of $32\times32$. This dataset comprises 1,281,167 training samples and 50,000 test samples, distributed across 1,000 labels. 

\begin{figure}[H]
    \centering
    \includegraphics[width=\linewidth]{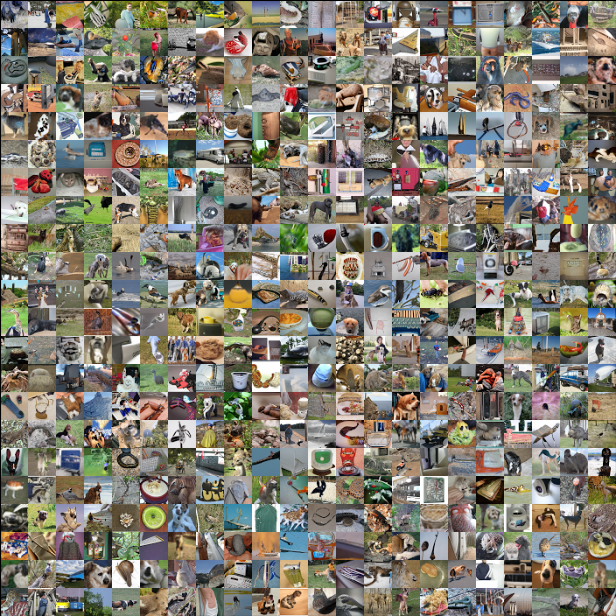}
    \caption{\method{} with noise parameterization after 2M training iterations.}
    \label{fig:imagenet_mulan_samples}
\end{figure}

\begin{figure}[H]
    \centering
    \includegraphics[width=\linewidth]{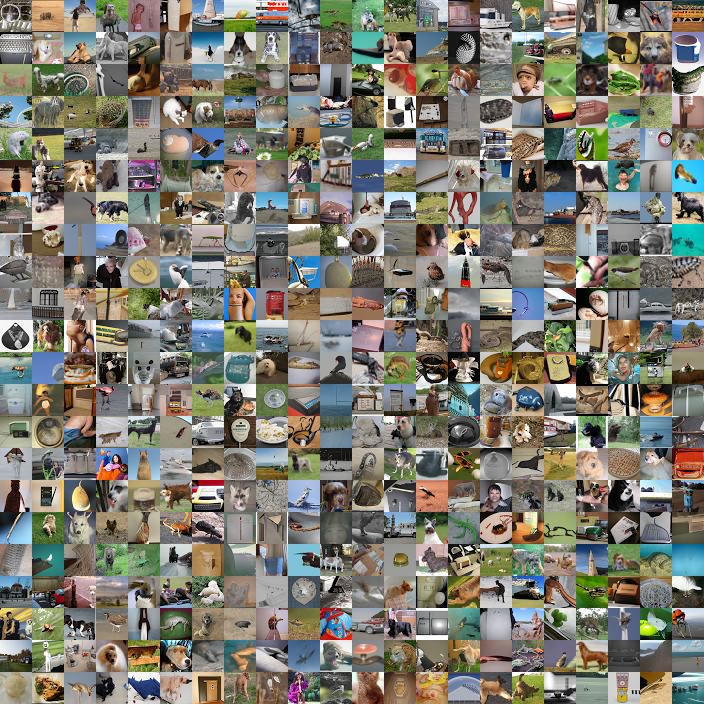}
    \caption{VDM after 2M training iterations.}
    \label{fig:imagenet_vdm_samples}
\end{figure}

Randomly generated samples for the ImageNet datasaet are provided in~\fig{fig:imagenet_mulan_samples} for \method{} and ~\fig{fig:imagenet_vdm_samples} for VDM. We visualize the noise schedule in \fig{fig:noise_schedules_all}.


\subsection{Frequency}
\begin{figure}[H]
    \centering
    \begin{subfigure}[t]{.47\linewidth}
    \includegraphics[width=\linewidth]{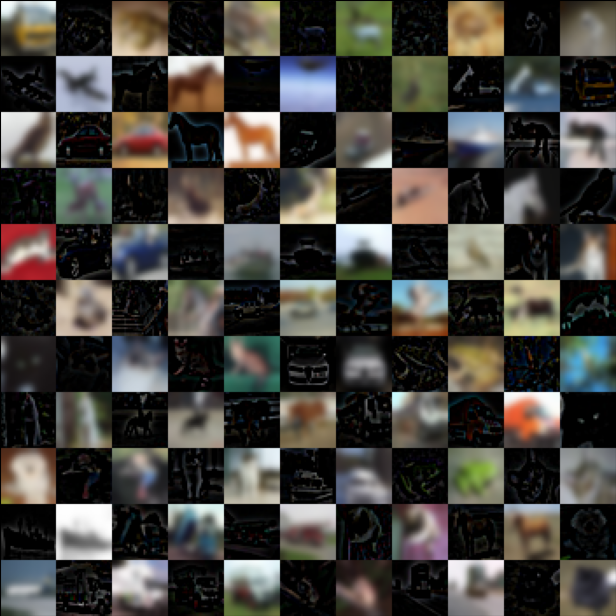}
    \caption{Training samples.}
    \label{fig:freq_train}
    \end{subfigure}\hfill
    \begin{subfigure}[t]{.47\linewidth}
    \includegraphics[width=\linewidth]{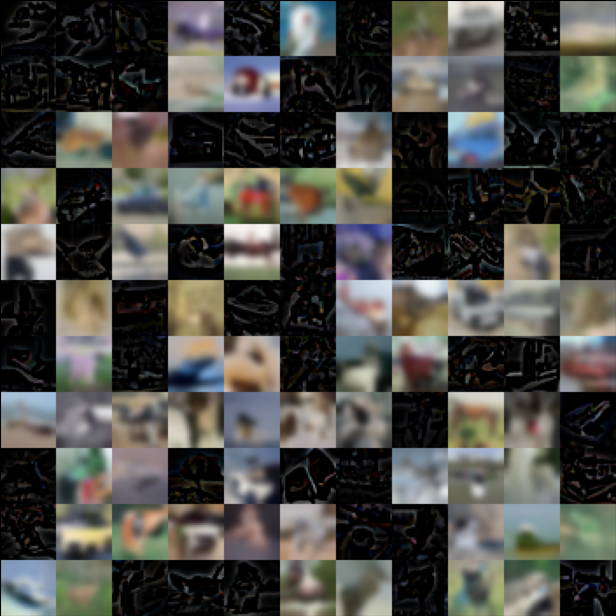}
    \caption{Samples generated by \method{} with noise parameterization after 500K training iterations.}
    \label{fig:freq_samples}
    \end{subfigure}
    \caption{Frequency Split CIFAR-10 dataset.}
\end{figure}

To see if \method{} learns different noise schedules for parts of the images with different frequencies, we modify the images in the CIFAR-10 dataset where we modify an image where we randomly remove the low frequency component an image or remove the high frequency with equal probability. 
\fig{fig:freq_train} shows the training samples. \method{} was trained for 500K steps. The samples generated by \method{} is shown in \fig{fig:freq_samples}. The corresponding noise schedules is shown in \fig{fig:noise_schedules_all}. As compared to CIFAR-10, we notice that the spatial variation in the noise schedule increases (SNRs for all the pixels form a wider band) while the variance of the noise schedule across instances decreases slightly.

\subsection{Frequency-2}
To see if \method{} learns different noise schedules for images with different frequencies, we modify the images in the CIFAR-10 dataset where we modify an image where we randomly remove the low frequency component an image or remove the high frequency with equal probability. 
\fig{fig:freq_train} shows the training samples. \method{} was trained for 500K steps. The samples generated by \method{} is shown in \fig{fig:freq_samples}. The corresponding noise schedules is shown in \fig{fig:noise_schedules_all}. As compared to CIFAR-10, we notice that the spatial variation in the noise schedule increases (SNRs for all the pixels form a wider band) and the variance of the noise schedule across instances increases as well. 

\begin{figure}[H]
    \centering
    \begin{subfigure}[t]{.47\linewidth}
    \includegraphics[width=\linewidth]{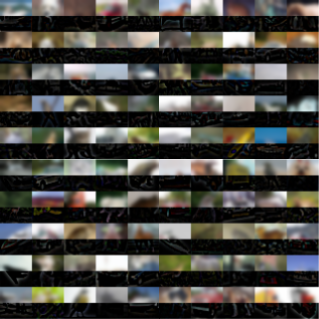}
    \caption{Training samples.}
    \label{fig:freq2_train}
    \end{subfigure}\hfill
    \begin{subfigure}[t]{.47\linewidth}
    \includegraphics[width=\linewidth]{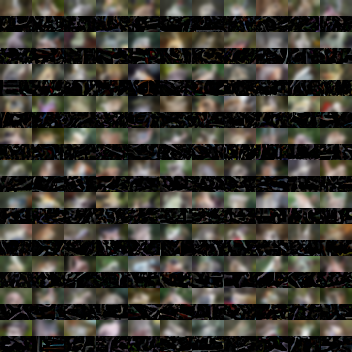}
    \caption{Samples generated by \method{} with noise parameterization after 500K training iterations.}
    \label{fig:freq2_samples}
    \end{subfigure}
    \caption{Frequency Split-2 CIFAR-10 dataset.}
\end{figure}

\subsection{CIFAR-10: Intensity}
\begin{figure}[H]
    \centering
    \begin{subfigure}[t]{.47\linewidth}
    \includegraphics[width=\linewidth]{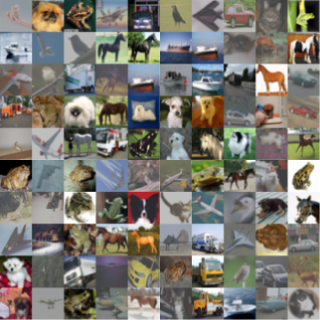}
    \caption{Training samples.}
    \label{fig:intensity_train}
    \end{subfigure}\hfill
    \begin{subfigure}[t]{.47\linewidth}
    \includegraphics[width=\linewidth]{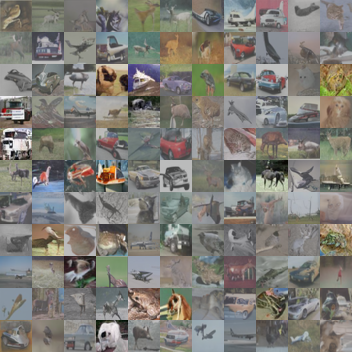}
    \caption{Samples generated by \method{} with noise parameterization after 500K training iterations.}
    \label{fig:intensity_samples}
    \end{subfigure}
    \caption{Intensity CIFAR-10 dataset.}
\end{figure}

To see if \method{} learns different noise schedules for images with different intensities, we modify the images in the CIFAR-10 dataset where we randomly convert an image into a low intensity or a high intensity image with equal probability. Originally, the CIFAR10 images are in the range [0, 255]. To convert an image into a low intensity image we multiply all pixel values by 0.5. To convert an image into a high intensity image we multiply all the pixel values by 0.5 and add 127.5 to them. \fig{fig:intensity_train} shows the training samples. \method{} was trained for 500K steps. The samples generated by \method{} is shown in \fig{fig:intensity_samples}. The corresponding noise schedules is shown in \fig{fig:noise_schedules_all}. As compared to CIFAR-10, we notice that the spatial variation in the noise schedule slightly increases (SNRs for all the pixels form a wider band) while the variance of the noise schedule across instances slightly decreases.

\subsection{Mask}
\begin{figure}[H]
    \centering
    \begin{subfigure}[t]{.47\linewidth}
    \includegraphics[width=\linewidth]{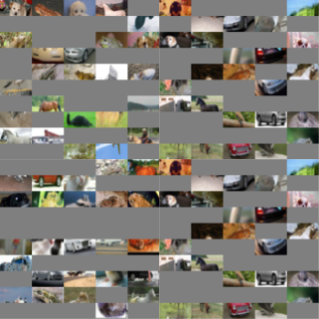}
    \caption{Training samples.}
    \label{fig:mask_train}
    \end{subfigure}\hfill
    \begin{subfigure}[t]{.47\linewidth}
    \includegraphics[width=\linewidth]{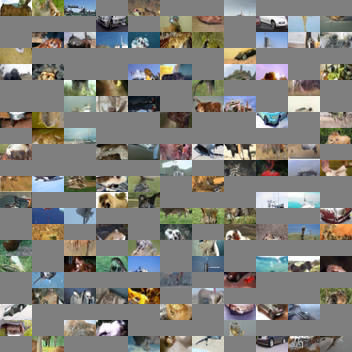}
    \caption{Samples generated by \method{} with noise parameterization after 500K training iterations.}
    \label{fig:mask_samples}
    \end{subfigure}
    \caption{Intensity CIFAR-10 dataset.}
\end{figure}

\begin{figure}[H]
    \centering\includegraphics[height=0.9\textheight, width=0.85\linewidth]{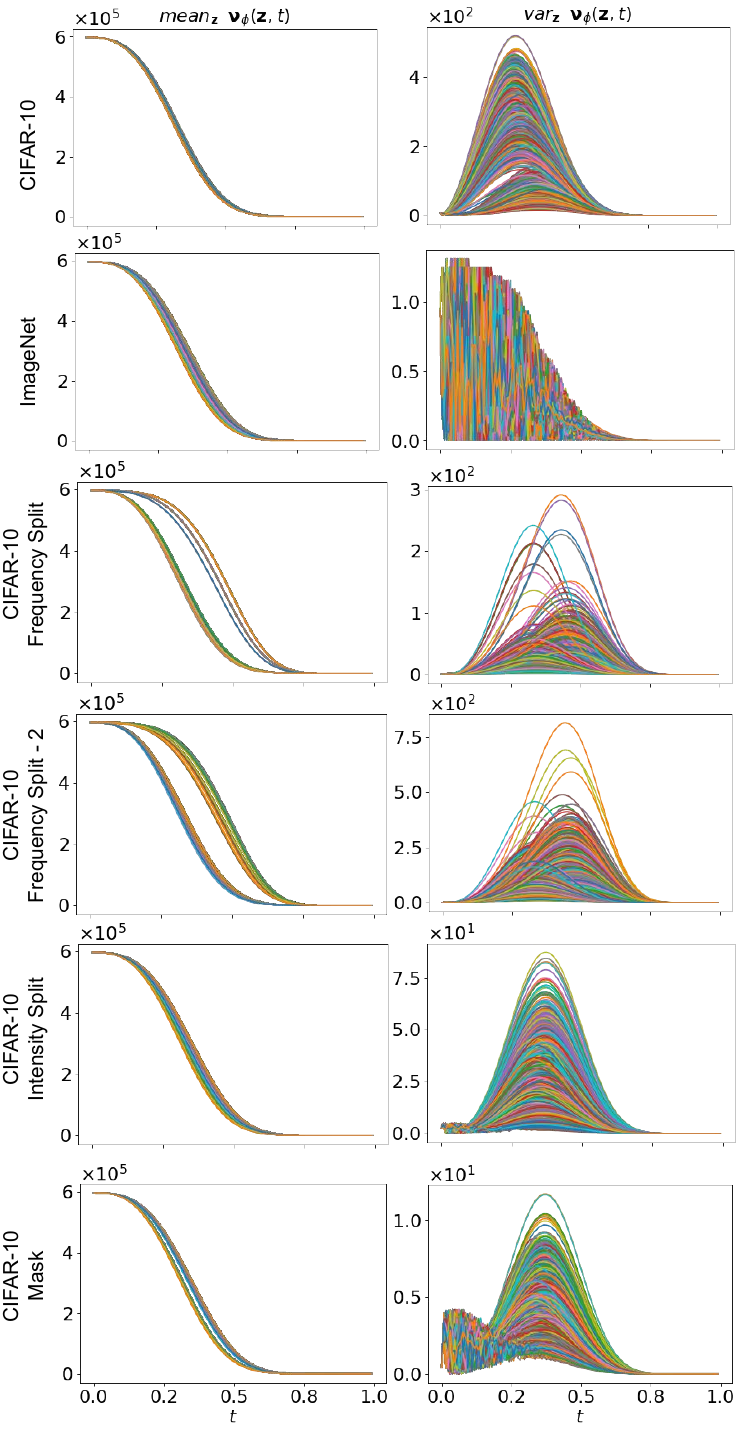}
    \caption{signal-to-noise ratio for different datasets.}
    \label{fig:noise_schedules_all}
\end{figure}

We modify the CIFAR-10 dataset where we randomly mask (i.e. replace with $\mathbf{0}$s) the top of an image or the bottom half of an image with equal probability. \fig{fig:mask_train} shows the training samples. \method{} was trained for 500K steps. The samples generated by \method{} is shown in \fig{fig:mask_samples}. The corresponding noise schedules is shown in \fig{fig:noise_schedules_all}. As compared to CIFAR-10, we notice that the spatial variation in the noise schedule slightly increases (SNRs for all the pixels form a wider band) while the variance of the noise schedule across instances decreases.

\section{Likelihood Estimation}
We used both Variance Lower Bound (VLB) and ODE-based methods to compute BPD. 

\subsection{VLB Estimate}\label{appendix:subsec:vlb}
In the VLB-based approach, we employ \Eqn{eqn:elbo}. To compute $\lossdiff$, we use $T = 128$ in \Eqn{eqn:diffusion_loss_discrete}, discretizing the timesteps, $t \in [0, 1]$ into $128$ bins.

\subsection{Exact likelihood computation using Probability Flow ODE}\label{appendix:subsec:mulan_ode}
A diffusion process whose marginal is given by (the same as in \Eqn{eqn:multivariate_forward}), 
\begin{equation}
    q(\x_t | \x_0) = \mathcal{N}(\x_{t}; \balpha_{t} \x_0, \diag(\bsigma^2_{t})), \;\x_0 \sim q_0(\x_0),
\end{equation}
can be modeled as the solution to an It$\hat{\text{o}}$  Stochastic Differential Equation (SDE):
\begin{equation}\label{eqn:sde_forward_simple}
    \dif\x_t = \f (t) \x_t\dif t + \g(t) \dif \w_t, \;\x_0 \sim q_0(\x_0),
\end{equation}
where $\f (t) \in \mathbb{R}^d, \g(t) \in \mathbb{R}^d$ take the following expressions~\citep{song2020score}:
\begin{align}
    \f(t) &= \frac{\dif}{\dif t} \log \balpha_t, \nonumber \\
    \g^2(t) &= \frac{\dif}{\dif t} \bsigma^2_t - 2 \bsigma^2_t\frac{\dif}{\dif t} \log \balpha_t \nonumber
\end{align}
The corresponding reverse process, \Eqn{eqn:q_context_conditioned}, can also be modelled by an equivalent reverse-time SDE:
\begin{align}\label{eqn:sde_reverse}
    \dif\x_t = [\f (t) - \g(t)^2 \gradx \log q(\x_t | \x_0)] \dif t + \g(t) \dif \bar{\w}_t, \;\x_1 \sim \p(\x_1),
\end{align}
where $\bar{\w}$ is a standard Wiener process when time flows backwards from $1 \to 0$, and $\dif t$ is an infinitesimal negative timestep. \citet{song2020score} show that the marginals of \Eqn{eqn:sde_reverse} can be described by the following Ordinary Differential Equation (ODE) in the reverse process:
\begin{equation}\label{eqn:ode_reverse_simple}
    \dif\x_t = \left[ \f (t)\x_t - \frac{1}{2}\g^2(t) \gradx \log q(\x_t) \right] \dif t.
\end{equation}
This ODE, also called the probablity flow ODE, allows us to compute the exact likelihood on any input data via the instantaneous change of variables formula as proposed in \citet{chen2018neural}. Note that during the reverse process, 
the term $ q(\x_t)$ is unknown and is approximated by parameterized by $ \p(\x_t)$.
For the probability flow defined in \Eqn{eqn:ode_reverse_simple},
\citet{chen2018neural} show that the log-
likelihood of $\p(\x_0)$ can be computed using the following equation:
\begin{align}\label{eqn:ode_likelihood_simple}
    \log \p(\x_0) & = \log \p(\x_1) - \int_{t=0}^{t=1} \text{tr}\left(\gradx \mathbf{h_\theta}(\x_t, t) \right) \dif t, \\
    & \text{\footnotesize where}\; \mathbf{h_\theta}(\x_t, t) \equiv \f (t)\x_t - \frac{1}{2} \g^2(t) \gradx \log \p(\x_t) \nonumber 
\end{align}

\subsubsection{Probability Flow ODE for \method{}.}
Similarly for the forward process conditioned on the auxiliary latent variable, $\z$, 
\begin{equation}
    \q(\x_t | \x_0, \z) = \mathcal{N}(\x_{t}; \balpha_{t}(\z) \x_0, \diag(\bsigma^2_{t}(\z))), \;\x_0 \sim q_0(\x_0),\;\z \sim \q(\z | \x_0),
\end{equation}
we can extend \Eqn{eqn:sde_forward_simple} in the following manner,
\begin{equation}
    \dif\x_t = \f (\z, t) \x_t\dif t + \g(\z, t) \dif \w_t, \;\x_0 \sim q_0(\x_0),\;\z \sim \q(\z | \x_0),
\end{equation}
to obtain the corresponding SDE formulation. Notice that the random variable $\z$ in the above equation doesn't have a subscript $t$, and hence, $\z$ is drawn from $\q(\z | \x_0)$ once and the same $\z$ is used as $\x_0$ diffuses to $\x_1$.  The expressions for
$\f (\z, t): \mathbb{R}^m \times [0, 1] \to \mathbb{R}^d$,
$\g(\z, t): \mathbb{R}^m \times [0, 1] \to \mathbb{R}^d$ is given as follows:
\begin{align}
    \f(\z, t) &= \frac{\dif}{\dif t} \log \balpha_t(\z), \nonumber \\
    \g^2(\z, t) &= \frac{\dif}{\dif t} \bsigma^2_t(\z) - 2 \bsigma^2_t(\z)\frac{\dif}{\dif t} \log \balpha_t(\z) \nonumber
\end{align}
Recall that $\balpha^2_t(\z) = \text{sigmoid}(- \ns_\phi(\z,t))$,
$\bsigma^2_t(\z) = \text{sigmoid}(\ns_\phi(\z, t))$. Substituting these in the above equations, the expressions for $\f(\z, t)$ and $\g^2(\z, t)$ simplify to the following:
\begin{align}
    \f(\z, t) &= - \frac{1}{2} \text{sigmoid}(\ns_\phi(\z, t))\frac{\dif}{\dif t}\ns_\phi(\z, t), \nonumber \\
    \g^2(\z, t) &= \text{sigmoid}(\ns_\phi(\z, t)) \frac{\dif}{\dif t}\ns_\phi(\z, t) \nonumber
\end{align}

The corresponding reverse-time SDE is given as:
\begin{align}\label{eqn:sde_reverse_mulan}
    \dif\x_t = [\f (t) - \g(t)^2 \gradx \log \q(\x_t | \x_0, \z)] \dif t + \g(t) \dif \bar{\w}_t, \;\x_1 \sim \p(\x_1), \;\z \sim \p(\z),
\end{align}
where $\bar{\w}$ is a standard Wiener process when time flows backwards from $1 \to 0$, and $\dif t$ is an infinitesimal negative timestep. Given, $\score(\x_t, \z)$, an approximation to the true score function, $\gradx \log \q(\x_t | \x_0, \z)$, 
\citet{song2020score} show that the marginals of \Eqn{eqn:sde_reverse_mulan} can be described by the following Ordinary Differential Equation (ODE):
\begin{equation}\label{eqn:change_of_variable_mulan}
    \dif\x_t = \left[ \f (\z, t) - \frac{1}{2}\g^2(\z, t) \score(\x_t, \z) \right] \dif t,
\end{equation}
\citet{zheng2023improved} show that the score function, $\score(\x_t, \z)$, for the noise and the v-parameterization is given as follows:
\begin{subnumcases} {\label{weqn} \score(\x_t, \z) =}
   - \frac{\unet(\x_t, t)}{\bsigma_t(\z)}  & \text{Noise parameterization; see~\sec{appendix:subsec:noise_param}}\\
  -\x_t - \exp\left(-\frac{1}{2} \ns_\phi(\z, t)\right)\mathbf{v}_\theta(\x_t, \z, t)
  & \text{v-parameterization; see~\sec{appendix:subsec:velocity_param}}
\end{subnumcases}
Applying the change of variables formula~\citep{chen2018neural} on \Eqn{eqn:change_of_variable_mulan},  
$\log p_\theta(\x_0 | \z)$ can be computed in the following manner:
\begin{align}\label{eqn:ode_likelihood_mulan}
    \log \p(\x_0 | \z) &= \log \p(\x_1) - \int_{t=0}^{t=1} \text{tr}\left(\gradx \mathbf{h_\theta}(\x_t, \z, t) \right) \dif t, \\
    & \text{\footnotesize where ${\mathbf h}_\theta(\x_t, \z, t) \equiv \f (\z, t) - \frac{1}{2}\g^2(\z, t) \score(\x_t, \z)$} \nonumber
\end{align}

The expression for log-likelihood~(\Eqn{eqn:elbo2x}) is as follows, 
\begin{align}\label{eqn:exact_likelihood_mulan}
    \log p_\theta(\x_0) 
    & \geq \mathbb{E}_{\q(\z|\x_0)} [ \log p_\theta(\x_0 | \z)] - \kl(\q(\z|\x_0) \| p_\theta(\z)) \nonumber\\
    & \;\text{\footnotesize Using~\Eqn{eqn:ode_likelihood_mulan},} \nonumber \\
    & = \mathbb{E}_{\q(\z|\x_0)} \left[ \log \p(\x_1) - \int_{t=0}^{t=1} \text{tr}\left(\gradx \mathbf{h_\theta}(\x_t, t, \z) \right) \dif t \right] - \kl(\q(\z|\x_0) \| p_\theta(\z))
\end{align}
Computing $\text{tr}\left(\gradx \mathbf{h_\theta}(\x_t, t, \z) \right)$ is expensive and we follow \citet{chen2018neural, zheng2023improved, grathwohl2018ffjord} to estimate it with Skilling-Hutchinson trace estimator \citep{Skilling1989TheEO, hutchinson1989stochastic}. In particular, we have 
\begin{align}\label{eqn:hutchinson}
    \text{tr}\left(\gradx \mathbf{h_\theta}(\x_t, t, \z)\right) = \mathbb{E}_{p(\epsilon)}\left[\epsilon^\top
    \gradx \mathbf{h_\theta}(\x_t, t, \z) \epsilon
    \right],
\end{align}
where the random variable $\epsilon$ satisfies $\mathbb{E}_{p(\epsilon)}[\epsilon] = \mathbf{0}$ and $\text{Cov}_{p(\epsilon)}[\epsilon] = \mathbf{I}$. Common choices for $p(\epsilon)$ include Rademacher or Gaussian distributions. Notably, the term $\gradx \mathbf{h_\theta}(\x_t, t, \z) \epsilon$ can be computed efficiently using ``Jacobian-vector-product'' computation in JAX.
In our experiments, we follow the exact evaluation procedure for computing likelihood as outlined in \citet{song2020score, grathwohl2018ffjord}. Specifically, for the computation of \Eqn{eqn:hutchinson}, we employ a Rademacher distribution for $p(\epsilon)$.
To calculate the integral in \Eqn{eqn:exact_likelihood_mulan}, we utilize the RK45 ODE solver \citep{DORMAND198019} provided by \texttt{scipy.integrate.solve\_ivp} with \texttt{atol=1e{-5}} and \texttt{rtol=1e{-5}}.

\subsubsection{Dequantization.}
Real-world datasets for images or texts often consist of discrete data. Attempting to learn a continuous density model directly on these discrete data points can lead to degenerate outcomes 
\citep{uria2013rnade} and fail to provide meaningful density estimations. Dequantization \citep{salimans2017pixelcnn++, ho2020denoising, zheng2023improved} is a common solution in such cases.
To elaborate, let $x_0$ represent
8-bit discrete data scaled to [-1, 1].
Dequantization methods assume that we have
trained a continuous model distribution $\p$  for $x_0$, and
define the discrete model distribution by
$$P_\theta(\x_0) = \int_{[-\frac{1}{256}, \frac{1}{256})^d} \p(\x_0 + u)\dif u.$$
To train $P_\theta(\x_0)$ by maximum likelihood estimation, variational dequantization \citep{ho2020denoising, zheng2023improved} introduces a dequantization distribution $q(u | \x_0)$ and jointly train $p_\text{model}$ and
$q(u|\x_0)$ by a variational lower bound:
\begin{align}\label{eqn:dequantization}
    \log P_\theta(\x_0) \geq \mathbb{E}_{q(u | \x_0)}[\p(\x_0 + u) - \log q(u | \x_0)].
\end{align}

\paragraph{Truncated Normal Dequantization.} \citet{zheng2023improved} show that truncated Normal distribution,
$$q(u|\x_0) = \mathcal{TN}\left(\mathbf{0}, \mathbf{I}, -\frac{1}{256}, \frac{1}{256}\right)$$
with mean $\mathbf{0}$, covariance $\mathbf{I}$, and bounds $\left[-\frac{1}{256}, \frac{1}{256}\right]$ along each dimension, leads to a better likelihood estimate. Thus, \Eqn{eqn:dequantization} simplifies to the following (for details please refer to section A. in \citet{zheng2023improved}):
\begin{align}\label{eqn:tn}
    \log P_\theta(\x_0) \geq & \mathbb{E}_{
        \tneps \sim \mathcal{TN}\left(\mathbf{0}, \mathbf{I}, -\tau, \tau\right)} 
        \left[\log \p \left(\x_0 + \frac{\sigma_\epsilon}{\alpha_\epsilon}\tneps \right)\right]
        + \frac{d}{2} (1 + \log(2 \pi \sigma^2_\epsilon)) - 0.01522 \times d \\
    & \text{\footnotesize with  $\frac{\sigma_\epsilon}{\alpha_\epsilon} = \text{exp}(- \frac{1}{2} \times 13.3)$,} \nonumber \\ 
    & \text{\footnotesize $\sigma_\epsilon = \texttt{sqrt}(\text{sigmoid}(-13.3))$, and $\tau = 3$.} \nonumber 
\end{align}
$\log \p \left(\x_0 + \frac{\sigma_\epsilon}{\alpha_\epsilon}\tneps\right)$ is evaluated using \Eqn{eqn:exact_likelihood_mulan}.

\paragraph{Importance Weighted Estimator.} \Eqn{eqn:tn} can also be extended to obtain an importance weighted likelihood estimator to get a tighter bound on the likelihood. The variational bound is given by (for details please refer to section A. in \citet{zheng2023improved}):
\begin{align}\label{eqn:tn_iwbo}
    \log P_\theta(\x_0) \geq & \mathbb{E}_{
        \tneps^{(1)}, \dots, \tneps^{(K)} \sim \mathcal{TN}\left(\mathbf{0}, \mathbf{I}, -\tau, \tau\right)} 
        \left[\log \left(\frac{1}{K} \sum_{i=1}^K  \frac{\p \left(\x_0 + \frac{\sigma_\epsilon}{\alpha_\epsilon}\tneps^{(k)} \right)}{q(\tneps^{(i)})}\right)\right]
        + d \log\sigma_\epsilon \\
    & \text{\footnotesize with  $\frac{\sigma_\epsilon}{\alpha_\epsilon} = \text{exp}(- \frac{1}{2} \times 13.3)$,
    $\log \sigma_\epsilon = \frac{1}{2}(-13.3 + \text{softplus}(-13.3))$,} \nonumber \\
    & \text{\footnotesize $q(\tneps) = \frac{1}{(2\pi Z)^2}\text{exp}\left(-\frac{1}{2} \|\tneps\|_2^2\right)$, $Z = 0.9974613$, and $\tau = 3$.} \nonumber
\end{align}
Note that for $K=1$, \Eqn{eqn:tn_iwbo} is equivalent to \Eqn{eqn:tn}; see~\citet{zheng2023improved}.
$\log \p \left(\x_0 + \frac{\sigma_\epsilon}{\alpha_\epsilon}\tneps\right)$ is evaluated using \Eqn{eqn:exact_likelihood_mulan}.
In~\tab{tab:dequantization}, we report BPD values for \method{} on CIFAR10 (8M training steps, v-parameterization) and ImageNet (2M training steps, noise parameterization) using both the VLB-based approach, and the ODE-based approach with $K=1$ and $K=20$ importance samples.

\begin{table}
    \centering
    \caption{NLL (mean and 95\% Confidence Interval for \method{}) on CIFAR10 (8M training steps, v-parameterization) and ImageNet (2M training steps, noise parameterization) using both the VLB-based approach, and the ODE-based approach. $K=1$ means that we do not use importance weighted estimator since \Eqn{eqn:tn_iwbo} is equivalent to \Eqn{eqn:tn} for this case; see~\citet{zheng2023improved}.}
    \label{tab:dequantization}
    \begin{tabular}{l|c|c}
         Likelihood Estimation type& CIFAR-10 ($\downarrow$) & Imagenet ($\downarrow$)  \\
         \hline
         VLB-based & 2.59 $\pm 10^{-3}$ & 3.71 $\pm 10^{-3}$ \\
         ODE-based ($K=1$; \Eqn{eqn:tn}) & 2.59 $\pm 3 \times 10^{-4}$ & 3.71 $\pm 10^{-3}$ \\
         ODE-based ($K=20$; \Eqn{eqn:tn_iwbo}) & 2.55 $\pm 3 \times 10^{-4}$ & 3.67 $\pm 10^{-3}$ \\
    \end{tabular}
\end{table}

\newpage
\section*{NeurIPS Paper Checklist}

\end{appendices}
\begin{enumerate}

\item {\bf Claims}
    \item[] Question: Do the main claims made in the abstract and introduction accurately reflect the paper's contributions and scope?
    \item[] Answer: \answerYes{} %
    \item[] Justification: See our introduction for a list of claims including getting SOTA results on density estimation. %

\item {\bf Limitations}
    \item[] Question: Does the paper discuss the limitations of the work performed by the authors?
    \item[] Answer: \answerYes{} %
    \item[] Justification: Yes, our model does not get state of the art FID due to it not having a lower frequency bias. See the paper for more details.
    \item[] Guidelines:
    \begin{itemize}
        \item The answer NA means that the paper has no limitation while the answer No means that the paper has limitations, but those are not discussed in the paper. 
        \item The authors are encouraged to create a separate "Limitations" section in their paper.
        \item The paper should point out any strong assumptions and how robust the results are to violations of these assumptions (e.g., independence assumptions, noiseless settings, model well-specification, asymptotic approximations only holding locally). The authors should reflect on how these assumptions might be violated in practice and what the implications would be.
        \item The authors should reflect on the scope of the claims made, e.g., if the approach was only tested on a few datasets or with a few runs. In general, empirical results often depend on implicit assumptions, which should be articulated.
        \item The authors should reflect on the factors that influence the performance of the approach. For example, a facial recognition algorithm may perform poorly when image resolution is low or images are taken in low lighting. Or a speech-to-text system might not be used reliably to provide closed captions for online lectures because it fails to handle technical jargon.
        \item The authors should discuss the computational efficiency of the proposed algorithms and how they scale with dataset size.
        \item If applicable, the authors should discuss possible limitations of their approach to address problems of privacy and fairness.
        \item While the authors might fear that complete honesty about limitations might be used by reviewers as grounds for rejection, a worse outcome might be that reviewers discover limitations that aren't acknowledged in the paper. The authors should use their best judgment and recognize that individual actions in favor of transparency play an important role in developing norms that preserve the integrity of the community. Reviewers will be specifically instructed to not penalize honesty concerning limitations.
    \end{itemize}

\item {\bf Theory Assumptions and Proofs}
    \item[] Question: For each theoretical result, does the paper provide the full set of assumptions and a complete (and correct) proof?
    \item[] Answer: \answerYes{} %
    \item[] Justification: Please see our detailed proofs.
    \item[] Guidelines:
    \begin{itemize}
        \item The answer NA means that the paper does not include theoretical results. 
        \item All the theorems, formulas, and proofs in the paper should be numbered and cross-referenced.
        \item All assumptions should be clearly stated or referenced in the statement of any theorems.
        \item The proofs can either appear in the main paper or the supplemental material, but if they appear in the supplemental material, the authors are encouraged to provide a short proof sketch to provide intuition. 
        \item Inversely, any informal proof provided in the core of the paper should be complemented by formal proofs provided in appendix or supplemental material.
        \item Theorems and Lemmas that the proof relies upon should be properly referenced. 
    \end{itemize}

    \item {\bf Experimental Result Reproducibility}
    \item[] Question: Does the paper fully disclose all the information needed to reproduce the main experimental results of the paper to the extent that it affects the main claims and/or conclusions of the paper (regardless of whether the code and data are provided or not)?
    \item[] Answer: \answerYes{} %
    \item[] Justification: Not only do we show all equations and train on standard datasets, we will open source the code.%
    \item[] Guidelines:
    \begin{itemize}
        \item The answer NA means that the paper does not include experiments.
        \item If the paper includes experiments, a No answer to this question will not be perceived well by the reviewers: Making the paper reproducible is important, regardless of whether the code and data are provided or not.
        \item If the contribution is a dataset and/or model, the authors should describe the steps taken to make their results reproducible or verifiable. 
        \item Depending on the contribution, reproducibility can be accomplished in various ways. For example, if the contribution is a novel architecture, describing the architecture fully might suffice, or if the contribution is a specific model and empirical evaluation, it may be necessary to either make it possible for others to replicate the model with the same dataset, or provide access to the model. In general. releasing code and data is often one good way to accomplish this, but reproducibility can also be provided via detailed instructions for how to replicate the results, access to a hosted model (e.g., in the case of a large language model), releasing of a model checkpoint, or other means that are appropriate to the research performed.
        \item While NeurIPS does not require releasing code, the conference does require all submissions to provide some reasonable avenue for reproducibility, which may depend on the nature of the contribution. For example
        \begin{enumerate}
            \item If the contribution is primarily a new algorithm, the paper should make it clear how to reproduce that algorithm.
            \item If the contribution is primarily a new model architecture, the paper should describe the architecture clearly and fully.
            \item If the contribution is a new model (e.g., a large language model), then there should either be a way to access this model for reproducing the results or a way to reproduce the model (e.g., with an open-source dataset or instructions for how to construct the dataset).
            \item We recognize that reproducibility may be tricky in some cases, in which case authors are welcome to describe the particular way they provide for reproducibility. In the case of closed-source models, it may be that access to the model is limited in some way (e.g., to registered users), but it should be possible for other researchers to have some path to reproducing or verifying the results.
        \end{enumerate}
    \end{itemize}

\item {\bf Open access to data and code}
    \item[] Question: Does the paper provide open access to the data and code, with sufficient instructions to faithfully reproduce the main experimental results, as described in supplemental material?
    \item[] Answer: \answerNo{} %
    \item[] Justification: We will open source after paper acceptance.
    \item[] Guidelines:
    \begin{itemize}
        \item The answer NA means that paper does not include experiments requiring code.
        \item Please see the NeurIPS code and data submission guidelines (\url{https://nips.cc/public/guides/CodeSubmissionPolicy}) for more details.
        \item While we encourage the release of code and data, we understand that this might not be possible, so “No” is an acceptable answer. Papers cannot be rejected simply for not including code, unless this is central to the contribution (e.g., for a new open-source benchmark).
        \item The instructions should contain the exact command and environment needed to run to reproduce the results. See the NeurIPS code and data submission guidelines (\url{https://nips.cc/public/guides/CodeSubmissionPolicy}) for more details.
        \item The authors should provide instructions on data access and preparation, including how to access the raw data, preprocessed data, intermediate data, and generated data, etc.
        \item The authors should provide scripts to reproduce all experimental results for the new proposed method and baselines. If only a subset of experiments are reproducible, they should state which ones are omitted from the script and why.
        \item At submission time, to preserve anonymity, the authors should release anonymized versions (if applicable).
        \item Providing as much information as possible in supplemental material (appended to the paper) is recommended, but including URLs to data and code is permitted.
    \end{itemize}

\item {\bf Experimental Setting/Details}
    \item[] Question: Does the paper specify all the training and test details (e.g., data splits, hyperparameters, how they were chosen, type of optimizer, etc.) necessary to understand the results?
    \item[] Answer: \answerYes{} %
    \item[] Justification: Yes, we include all hyperparameters in the paper and will open source code.
    \item[] Guidelines:
    \begin{itemize}
        \item The answer NA means that the paper does not include experiments.
        \item The experimental setting should be presented in the core of the paper to a level of detail that is necessary to appreciate the results and make sense of them.
        \item The full details can be provided either with the code, in appendix, or as supplemental material.
    \end{itemize}

\item {\bf Experiment Statistical Significance}
    \item[] Question: Does the paper report error bars suitably and correctly defined or other appropriate information about the statistical significance of the experiments?
    \item[] Answer: \answerYes{}%
    \item[] Justification: we report the deviatations for BPD in Table 2.
    \item[] Guidelines:
    \begin{itemize}
        \item The answer NA means that the paper does not include experiments.
        \item The authors should answer "Yes" if the results are accompanied by error bars, confidence intervals, or statistical significance tests, at least for the experiments that support the main claims of the paper.
        \item The factors of variability that the error bars are capturing should be clearly stated (for example, train/test split, initialization, random drawing of some parameter, or overall run with given experimental conditions).
        \item The method for calculating the error bars should be explained (closed form formula, call to a library function, bootstrap, etc.)
        \item The assumptions made should be given (e.g., Normally distributed errors).
        \item It should be clear whether the error bar is the standard deviation or the standard error of the mean.
        \item It is OK to report 1-sigma error bars, but one should state it. The authors should preferably report a 2-sigma error bar than state that they have a 96\% CI, if the hypothesis of Normality of errors is not verified.
        \item For asymmetric distributions, the authors should be careful not to show in tables or figures symmetric error bars that would yield results that are out of range (e.g. negative error rates).
        \item If error bars are reported in tables or plots, The authors should explain in the text how they were calculated and reference the corresponding figures or tables in the text.
    \end{itemize}

\item {\bf Experiments Compute Resources}
    \item[] Question: For each experiment, does the paper provide sufficient information on the computer resources (type of compute workers, memory, time of execution) needed to reproduce the experiments?
    \item[] Answer: \answerYes{} %
    \item[] Justification: We provide this in the paper.
    \item[] Guidelines:
    \begin{itemize}
        \item The answer NA means that the paper does not include experiments.
        \item The paper should indicate the type of compute workers CPU or GPU, internal cluster, or cloud provider, including relevant memory and storage.
        \item The paper should provide the amount of compute required for each of the individual experimental runs as well as estimate the total compute. 
        \item The paper should disclose whether the full research project required more compute than the experiments reported in the paper (e.g., preliminary or failed experiments that didn't make it into the paper). 
    \end{itemize}
    
\item {\bf Code Of Ethics}
    \item[] Question: Does the research conducted in the paper conform, in every respect, with the NeurIPS Code of Ethics \url{https://neurips.cc/public/EthicsGuidelines}?
    \item[] Answer: \answerYes{} %
    \item[] Justification: Our paper is just a diffusion model useful for compression.
    \item[] Guidelines:
    \begin{itemize}
        \item The answer NA means that the authors have not reviewed the NeurIPS Code of Ethics.
        \item If the authors answer No, they should explain the special circumstances that require a deviation from the Code of Ethics.
        \item The authors should make sure to preserve anonymity (e.g., if there is a special consideration due to laws or regulations in their jurisdiction).
    \end{itemize}

\item {\bf Broader Impacts}
    \item[] Question: Does the paper discuss both potential positive societal impacts and negative societal impacts of the work performed?
    \item[] Answer: \answerNA{} %
    \item[] Justification: 
    \item[] Guidelines:
    \begin{itemize}
        \item The answer NA means that there is no societal impact of the work performed.
        \item If the authors answer NA or No, they should explain why their work has no societal impact or why the paper does not address societal impact.
        \item Examples of negative societal impacts include potential malicious or unintended uses (e.g., disinformation, generating fake profiles, surveillance), fairness considerations (e.g., deployment of technologies that could make decisions that unfairly impact specific groups), privacy considerations, and security considerations.
        \item The conference expects that many papers will be foundational research and not tied to particular applications, let alone deployments. However, if there is a direct path to any negative applications, the authors should point it out. For example, it is legitimate to point out that an improvement in the quality of generative models could be used to generate deepfakes for disinformation. On the other hand, it is not needed to point out that a generic algorithm for optimizing neural networks could enable people to train models that generate Deepfakes faster.
        \item The authors should consider possible harms that could arise when the technology is being used as intended and functioning correctly, harms that could arise when the technology is being used as intended but gives incorrect results, and harms following from (intentional or unintentional) misuse of the technology.
        \item If there are negative societal impacts, the authors could also discuss possible mitigation strategies (e.g., gated release of models, providing defenses in addition to attacks, mechanisms for monitoring misuse, mechanisms to monitor how a system learns from feedback over time, improving the efficiency and accessibility of ML).
    \end{itemize}
    
\item {\bf Safeguards}
    \item[] Question: Does the paper describe safeguards that have been put in place for responsible release of data or models that have a high risk for misuse (e.g., pretrained language models, image generators, or scraped datasets)?
    \item[] Answer: \answerNA{} %
    \item[] Justification: We do not believe our method will have a high risk of abuse as our models are not perceptually SOTA, they only provide for state of the art logliklihood.
    \item[] Guidelines:
    \begin{itemize}
        \item The answer NA means that the paper poses no such risks.
        \item Released models that have a high risk for misuse or dual-use should be released with necessary safeguards to allow for controlled use of the model, for example by requiring that users adhere to usage guidelines or restrictions to access the model or implementing safety filters. 
        \item Datasets that have been scraped from the Internet could pose safety risks. The authors should describe how they avoided releasing unsafe images.
        \item We recognize that providing effective safeguards is challenging, and many papers do not require this, but we encourage authors to take this into account and make a best faith effort.
    \end{itemize}

\item {\bf Licenses for existing assets}
    \item[] Question: Are the creators or original owners of assets (e.g., code, data, models), used in the paper, properly credited and are the license and terms of use explicitly mentioned and properly respected?
    \item[] Answer: \answerNA{} %
    \item[] Justification: We are using standard benchmark datasets.
    \item[] Guidelines:
    \begin{itemize}
        \item The answer NA means that the paper does not use existing assets.
        \item The authors should cite the original paper that produced the code package or dataset.
        \item The authors should state which version of the asset is used and, if possible, include a URL.
        \item The name of the license (e.g., CC-BY 4.0) should be included for each asset.
        \item For scraped data from a particular source (e.g., website), the copyright and terms of service of that source should be provided.
        \item If assets are released, the license, copyright information, and terms of use in the package should be provided. For popular datasets, \url{paperswithcode.com/datasets} has curated licenses for some datasets. Their licensing guide can help determine the license of a dataset.
        \item For existing datasets that are re-packaged, both the original license and the license of the derived asset (if it has changed) should be provided.
        \item If this information is not available online, the authors are encouraged to reach out to the asset's creators.
    \end{itemize}

\item {\bf New Assets}
    \item[] Question: Are new assets introduced in the paper well documented and is the documentation provided alongside the assets?
    \item[] Answer: \answerNA{} %
    \item[] Justification: %
    \item[] Guidelines:
    \begin{itemize}
        \item The answer NA means that the paper does not release new assets.
        \item Researchers should communicate the details of the dataset/code/model as part of their submissions via structured templates. This includes details about training, license, limitations, etc. 
        \item The paper should discuss whether and how consent was obtained from people whose asset is used.
        \item At submission time, remember to anonymize your assets (if applicable). You can either create an anonymized URL or include an anonymized zip file.
    \end{itemize}

\item {\bf Crowdsourcing and Research with Human Subjects}
    \item[] Question: For crowdsourcing experiments and research with human subjects, does the paper include the full text of instructions given to participants and screenshots, if applicable, as well as details about compensation (if any)? 
    \item[] Answer: \answerNA{} %
    \item[] Justification: No crowdsource or research with human subjets %
    \item[] Guidelines:
    \begin{itemize}
        \item The answer NA means that the paper does not involve crowdsourcing nor research with human subjects.
        \item Including this information in the supplemental material is fine, but if the main contribution of the paper involves human subjects, then as much detail as possible should be included in the main paper. 
        \item According to the NeurIPS Code of Ethics, workers involved in data collection, curation, or other labor should be paid at least the minimum wage in the country of the data collector. 
    \end{itemize}

\item {\bf Institutional Review Board (IRB) Approvals or Equivalent for Research with Human Subjects}
    \item[] Question: Does the paper describe potential risks incurred by study participants, whether such risks were disclosed to the subjects, and whether Institutional Review Board (IRB) approvals (or an equivalent approval/review based on the requirements of your country or institution) were obtained?
    \item[] Answer: \answerNA{} %
    \item[] Guidelines:
    \begin{itemize}
        \item The answer NA means that the paper does not involve crowdsourcing nor research with human subjects.
        \item Depending on the country in which research is conducted, IRB approval (or equivalent) may be required for any human subjects research. If you obtained IRB approval, you should clearly state this in the paper. 
        \item We recognize that the procedures for this may vary significantly between institutions and locations, and we expect authors to adhere to the NeurIPS Code of Ethics and the guidelines for their institution. 
        \item For initial submissions, do not include any information that would break anonymity (if applicable), such as the institution conducting the review.
    \end{itemize}

\end{enumerate}

\end{document}